\documentclass[10pt,twocolumn,letterpaper]{article}

\usepackage[pagenumbers]{cvpr} 
\usepackage{tablefootnote}

\usepackage[dvipsnames]{xcolor}

\usepackage[pagebackref,breaklinks,colorlinks]{hyperref}

\def\paperID{89} 
\def\confName{3DV\xspace}
\def\confYear{2024\xspace}

\title{FastHuman: Reconstructing High-Quality Clothed Human in Minutes}

\author{Lixiang Lin$^{1}$ \qquad Songyou Peng$^{2}$ \qquad Qijun Gan$^{1}$ \qquad Jianke Zhu$^{1}$\\
$^{1}$Zhejiang University\qquad $^{2}$ETH Zurich
}

\begin{document}
\maketitle

\begin{abstract}
We propose an approach for optimizing high-quality clothed human body shapes in minutes, using multi-view posed images. While traditional neural rendering methods struggle to disentangle geometry and appearance using only rendering loss, and are computationally intensive, our method uses a mesh-based patch warping technique to ensure multi-view photometric consistency, and sphere harmonics (SH) illumination to refine geometric details efficiently. We employ oriented point clouds' shape representation and SH shading, which significantly reduces optimization and rendering times compared to implicit methods. Our approach has demonstrated promising results on both synthetic and real-world datasets, making it an effective solution for rapidly generating high-quality human body shapes. Project page \href{https://l1346792580123.github.io/nccsfs/}{https://l1346792580123.github.io/nccsfs/}
\end{abstract}

\section{Introduction}

\begin{figure}
	\centering
    \includegraphics[width=0.4\textwidth]{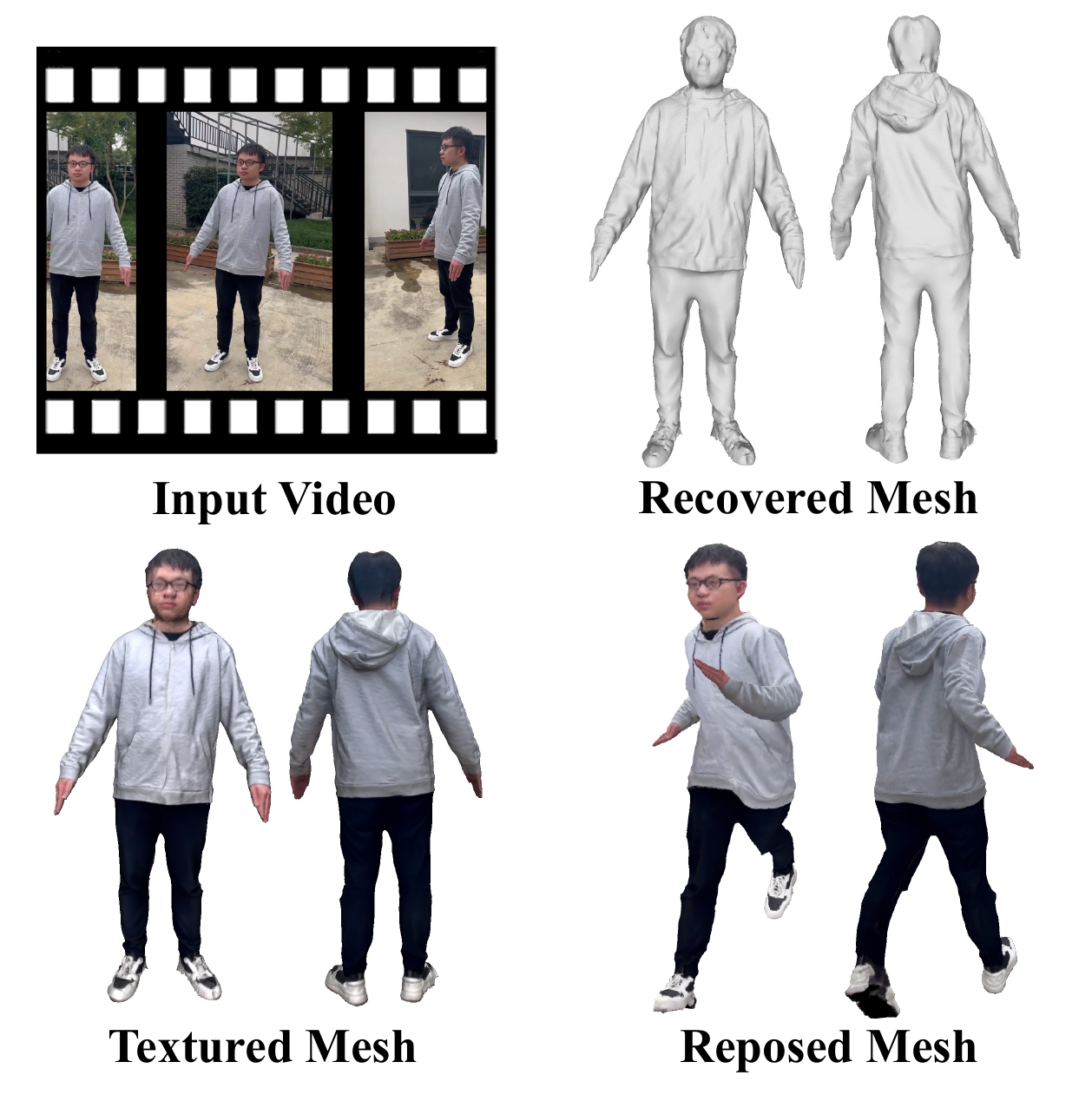}
	\caption{\textbf{FastHuman.} Our reconstruction results of an in-the-wild video captured by a phone. Camera parameters are estimated by Colmap~\cite{DBLP:conf/eccv/colmap16}, and RVM~\cite{DBLP:conf/wacv/rvm22} is used for human segmentation. We select an image every 20 frames from the video to construct a set of multi-view images. The whole process can be completed within 10 minutes.}
	\label{fig:teaser}
	\vspace{-0.2in}
\end{figure}

Human reconstruction is a challenging task due to its high complexity of extreme body poses and sophisticated clothing styles. In general, high-precision laser radar or photometric stereo~\cite{woodham1980photometric} is required to reconstruct the human body, which greatly increases the cost and only can be done in a controlled environment. Fig.~\ref{fig:teaser} shows our reconstruction results of an in-the-wild video captured by a phone. Our proposed method can reconstruct high-quality human meshes under general lighting environment in a few minutes. By leveraging the power of parametric model~\cite{DBLP:journals/tog/SMPL15} and deformation transfer~\cite{DBLP:journals/tog/deformtrans04} techniques, we are able to generate highly realistic reposed meshes.

With the rapid advancement of neural fields~\cite{Xie2022EUROGRPHICS}, there has been a surge of research devoted to representing 3D geometry and radiance fields using deep neural networks~\cite{DBLP:conf/eccv/nerf20, DBLP:conf/nips/idr20, DBLP:conf/nips/neus21,DBLP:conf/cvpr/neuralwarp22,Oechsle2021ICCV,Mescheder2019CVPR,Peng2020ECCV,Park2019CVPR,Zhu2022CVPR,Yu2022NEURIPS,Lionar2021WACV}. In these works, 3D geometries are commonly represented using volume density, occupancy, or signed distance functions (SDF). In order to model human avatars, some approaches~\cite{peng2021neural, DBLP:conf/iccv/animatenerf21, humannerf22,Wang2022ECCV,Chen2021ICCV} incorporate the estimated human skeleton and neural rendering to model animatable human avatars in an implicit manner. Neural rendering-based methods, while promising, often struggle with reconstructing accurate geometry. This is due to the inherent ambiguity between geometry and appearance, making it challenging to obtain accurate shapes through rendering loss alone. An image can be described by either a plane with complex appearance or a complex geometry with simple appearance. Deep networks can produce smooth surfaces, as the neural network may overfit the color differences between different views. However, shallow networks may lead to local optima due to their poor representation capability. Therefore, it's important to explicitly add multi-view consistency constraints to ensure accurate shape representation.

The implicit Multi Layer Perceptron (MLP) representation is not straightforward, as it requires forward inference to derive geometric information, such as volume density, SDF, etc. It usually takes long time to train these neural rendering-based methods as the whole MLP is updated during each iteration, resulting in slow convergence. Moreover, the rendering process is computational demanding, since the color of each pixel requires a forward network inference. Although some approaches~\cite{Reiser2021ICCV,yu2022plenoxels,mueller2022instant} are proposed to enable real-time rendering, large storage and GPU memory are required as trade off. 

To overcome the limitations mentioned above, we introduce a novel coarse-to-fine approach to reconstruct high-quality human meshes from multi-view images. Our approach utilizes an oriented point cloud as shape representation, which allows us to leverage the differentiable Poisson solver~\cite{Peng2021SAP} for efficient optimization. This ensures that our resulting surfaces are topology-agnostic and watertight, thereby improving the overall quality of the reconstructed mesh. Based on the traditional multi-view stereo approach~\cite{DBLP:journals/pami/mvs10, DBLP:conf/eccv/colmap16}, we warp small patches from the reference frame to source images. We then optimize the shape to ensure local photometric consistency. In addition, we incorporate shape-from-shading (SFS) techniques~\cite{Zhang1999PAMI} and estimate the 3rd sphere harmonic~(SH) coefficients to represent illumination and jointly refine the shapes and albedos with the shading formulation. As we adopt the simple shading model, the rendering process is sped up substantially compared to the conventional neural rendering methods. In summary, the main contributions of this paper are in the following.
\begin{itemize}
    \item We present \emph{FastHuman}, a coarse-to-fine pipeline to reconstruct high-quality clothed human bodies from multi-view images in just a few minutes.
    \item We propose a mesh-based patch-warping strategy to regularize surface optimization in the first stage. In the second stage, we fix the mesh topology and suggest an effective shading-based objective to refine the geometric details further and recover albedos based on shape from shading framework.
    \item We show the state-of-the-art 3D reconstruction results with significantly reduced computational time compared to existing methods on both synthetic and real-world datasets. 
    \item By taking advantage of parametric model, deformation transfer and SH illumination, we produce realistic reposing and relighting images.
\end{itemize}

\section{Related Works}

\subsection{Human Reconstruction}

Recovering 3D human body shapes from 2D images or videos has been extensively studied for decades~\cite{Liu2009TVCG,Starck2007surface,Vlasic2009SIGGRAPH,Wu2011ICCV,Alldieck2018CVPR}.
Existing approaches can be roughly divided into two categories: parametric model-based methods and model-free approaches.

\noindent\textbf{Parametric Model-based Human Reconstruction}
Many research efforts are devoted to building the statistical human body models from 3D scans~\cite{DBLP:journals/tog/SCAPE05, DBLP:journals/tog/SMPL15, DBLP:conf/cvpr/TotalCapture18, DBLP:conf/cvpr/SMPLX19,Xu2020CVPR}. In this way, human reconstruction is reduced to the parameter estimation problem, where the shape coefficients and joints transformation are predicted from images~\cite{DBLP:conf/eccv/Simplify16, DBLP:conf/cvpr/SMPLX19, DBLP:conf/cvpr/HMR18, DBLP:conf/iccv/SPIN19}. Most of these parametric model-based approaches only produce a naked human body, so the geometries of clothing, hair, and other accessories are typically ignored. 

\noindent\textbf{Model-free Human Reconstruction} Some approaches reconstruct human body without a predefined model. ~\cite{DBLP:conf/iccv/PIFu19, DBLP:conf/cvpr/pifuhd20} propose to represent the detailed human by a pixel-aligned implicit function, which predicts the occupancy for any locations. The occupancy for the sampled 3D point can be computed on the fly. With the rapid development of neural rendering, human reconstruction can be also viewed as the by-product of image synthesis. ~\cite{peng2021neural, DBLP:conf/iccv/animatenerf21, humannerf22} dynamically synthesize the human image. These methods use the SMPL~\cite{DBLP:journals/tog/SMPL15} parameters as inputs, and the volume rendering is employed to render images. Coarse human mesh can be extracted from the volume density through Marching cube algorithm~\cite{DBLP:conf/siggraph/Marchingcubes87}.

\begin{figure*}
    \centering
    \includegraphics[width=0.9 \linewidth]{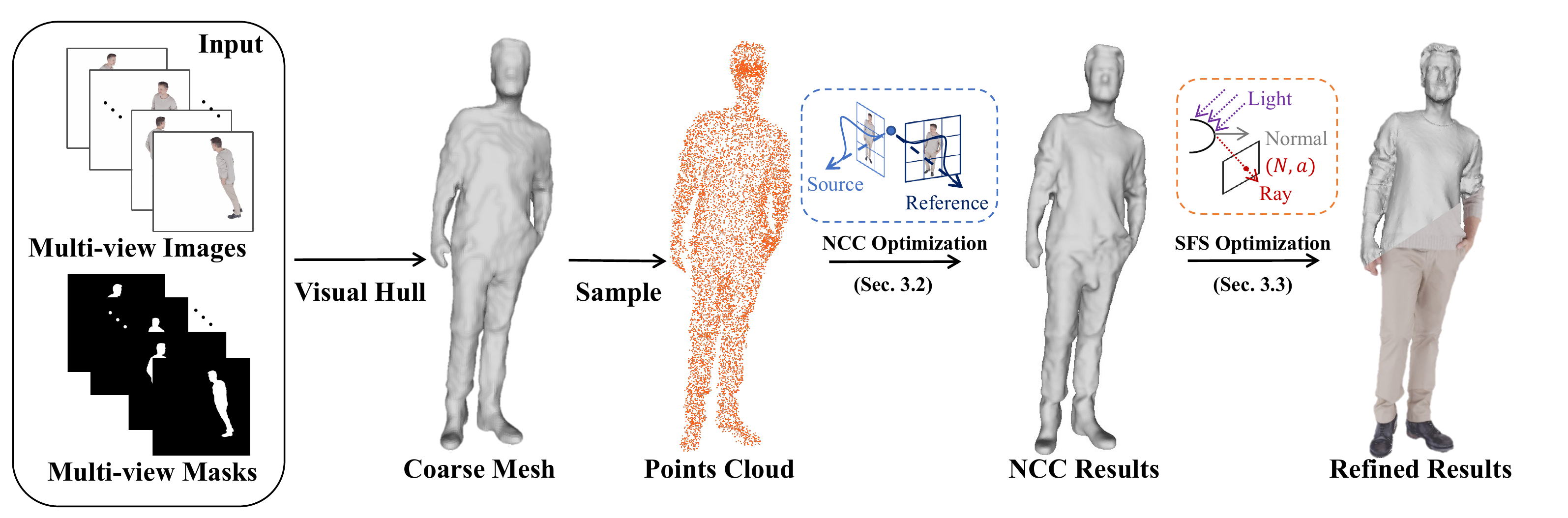}
    \caption{\textbf{Method Overview}. The initialized mesh is obtained by visual hull over the provided multi-view human masks. The oriented point clouds are sampled and optimized by multi-view patch-based photometric constraints. Moreover, we fix the mesh topology and employ a shape from shading refinement to refine the coarse mesh and recover the albedo.}
    \label{fig:overview}
    \vspace{-0.2in}
\end{figure*}

\subsection{Multi-View 3D Reconstruction}

Traditional Multi-View Stereo (MVS) methods estimate the depth maps by matching feature points across different views. Most of them assume that the appearance of a surface point is consistent in all visible views~\cite{DBLP:journals/pami/mvs10, DBLP:conf/eccv/colmap16}. Depth fusion and Poisson surface reconstruction~\cite{DBLP:conf/sgp/psr06} are required to extract a watertight mesh, whereas the surface details may be smoothed due to depth fusion. Recently, the learning-based MVS methods have received a lot attentions~\cite{DBLP:conf/cvpr/deepmvs18, DBLP:conf/eccv/mvsnet18}. DeepMVS~\cite{DBLP:conf/cvpr/deepmvs18} follow the traditional pipeline while replace the hand-crafted features to deep features. MVSNet~\cite{DBLP:conf/eccv/mvsnet18} warps deep features into the reference camera frustum to build a cost volume via differentiable homographies.

The recent trend in neural implcit representation and neural rendering also makes an impact in multi-view 3D reconstruction. IDR~\cite{DBLP:conf/nips/idr20} employs an MLP to represent scenes through SDF and light field implicitly, where color is only calculated at the surface intersection with a ray. ~\cite{DBLP:conf/nips/neus21,Oechsle2021ICCV,yariv2021volume,DBLP:conf/cvpr/neuralwarp22} incorporates volume rendering and implicit surface representation to learn the geometry from multi-view images. Although they can produce decent object-level 3D reconstruction, they suffer from reconstructing fine geometric details due to the use of a simple color loss, which cannot resolve the shape-appearance ambiguity. Moreover, using an MLP as shape representation leads to very slow optimization speed. In contrast, we propose in this paper two additional constraints for optimization, and use a lightweight point representation to speed up the optimziation.

\subsection{Shape from Shading}

Shape-from-shading (SFS) deals with the recovery of shape from a gradual variation of shading in image, which was first proposed by Horn~\cite{DBLP:phd/us/sfs70}. SFS is an ill-posed problem due to the ambiguity among lighting, reflectance and shape. There are some numerical solutions for SFS such like variational approach~\cite{DBLP:journals/cvgip/HornB86, DBLP:conf/emmcvpr/QueauMCCD17} and PDE methods~\cite{DBLP:journals/siamnum/CristianiF07}.

With the prevalence of consumer-level depth cameras, SFS approaches use the rough depth map as initialization for shape refinement~\cite{DBLP:conf/cvpr/sfs13,10.1145/sfs14,DBLP:conf/cvpr/sfs15,DBLP:journals/tog/ZollhoferDIWSTN15}. A series of work~\cite{Haefner2019PAMI,Peng2017ICCVW,Haefner2018CVPR,Sang2020WACV} also incorporate shape from shading and uncalibrated photometric stereo to upsample the low-resolution depth map from a RGB-D sensor in order to match the corresponding RGB image. We take the inspiration from SFS literatures and apply the shading refinement for the multi-view reconstruction task to further disambiguate shape from appearances, so we can recover fine geometric details.

\section{Methods}

We firstly introduce the oriented point clouds representation in Section~\ref{sec:shape_representation}. Next, we introduce our overall pipeline and the patch-based photometric consistency loss in Section~\ref{sec:photometric_cons}, and the shape refinement from shading in Section~\ref{sec:sfs}.
Details for our implementation are given in Section~\ref{sec:implement}.

\subsection{Oriented Point Clouds Shape Representation}\label{sec:shape_representation}
A recent work~\cite{Peng2021SAP} introduces a hybrid shape representation called Shape As Points (SAP), where they introduce an efficient differentiable Poisson solver (DPSR) to bridge oriented point clouds, implicit indicator functions, and meshes altogether. Compared to works using neural implicit-based shape representations~\cite{DBLP:conf/nips/neus21,Yu2022NEURIPS,DBLP:conf/nips/idr20,Liu2020CVPR,Oechsle2021ICCV,Niemeyer2020CVPR}, SAP allows representing any shapes as light-weight oriented point clouds, and yields the high-quality watertight meshes much more efficiently. Therefore, we leverage the power of SAP's optimization-based pipeline as the geometric representation for human reconstruction.

\subsection{Multi-view Photometric Consistency}\label{sec:photometric_cons}
Fig.~\ref{fig:overview} shows the overview of our proposed coarse-to-fine framework. Given masks of multi-view images, we firstly estimate an initial mesh via visual hull~\cite{DBLP:journals/pami/visualhull94}. Next, we sample an oriented point cloud $S = \{ \boldsymbol{x} \in \mathbb{R}^3, \boldsymbol{n} \in \mathbb{R}^3 \}$ from the inital mesh as the shape representation. During optimization, we generate a watertight mesh via DPSR and differentiable marching cubes (DMC):
\begin{equation}
    \chi = \text{DPSR}(S)
\end{equation}
\begin{equation}
  \mathcal{M} (V, F) = \text{DMC}(\text{tanh}(\chi)).
\end{equation}
$\chi$ represents an indicator function, where 1 indicates inside the object and 0 outside. $V$ and $F$ denote the vertices and faces of the mesh $\mathcal{M}$, respectively. The forward inference of DMC is the generic marching cube algorithm, and the gradients can be effectively approximated by the inverse surface normal~\cite{DBLP:conf/nips/meshsdf20}. The whole process is fully differentiable, so the loss can be backpropagated to update the oriented point clouds $S$. 

Given the input mesh $\mathcal{M}(V,F)$ with vertices $V$ and faces $F$, a differentiable renderer~\cite{Laine2020diffrast} denoted as $\zeta$ renders the attributes on vertices to pixels given the camera parameter $\pi$, which contains intrinsic matrix $K$ and extrinsic matrix $T$.
The rendered silhouette $\hat{M}$ can be obtained by interpolating the constant value of $\boldsymbol{1}$

\begin{equation}
    \hat{M} = \zeta(V,F, \boldsymbol{1} ;\pi).
\end{equation}

We impose the silhouette loss to limit the boundary of the generated mesh within the mask annotations,
\begin{equation}
    \mathcal{L}_{sil} = \sum_{i=1}^{N} ||M_{i} - \hat{M_{i}}||^2_2,
    \label{equ:mask}
\end{equation}
where $|| \cdot ||^2_2$ represents $L_2$ norm. $i=1, \cdots, N$ represents all views.

\begin{figure}
	\centering
    \includegraphics[width=0.6\textwidth]{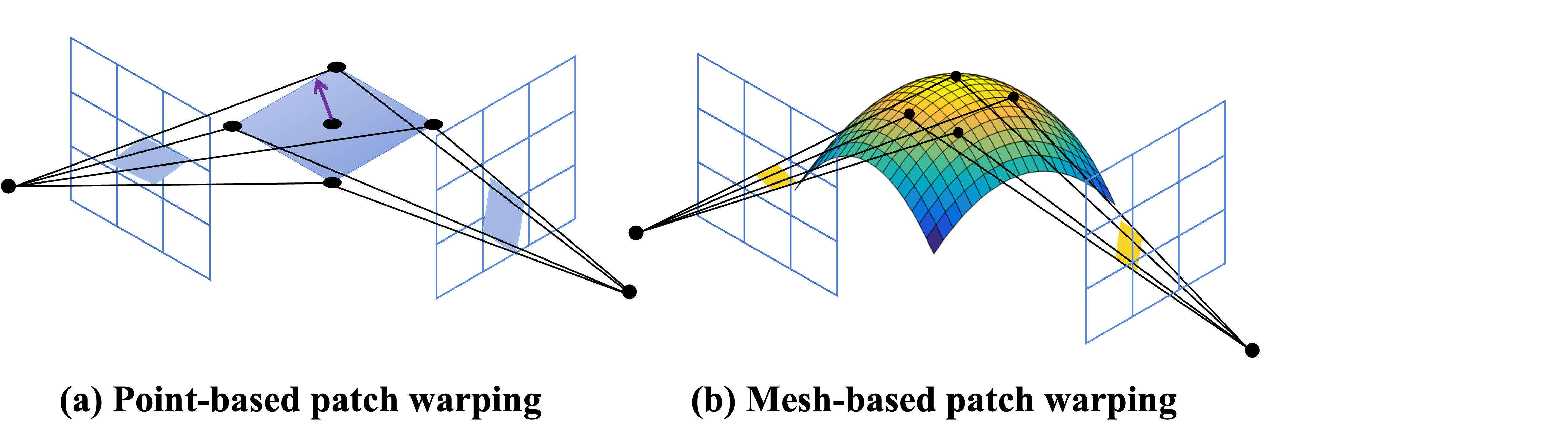}
	\caption{\textbf{Point-based v.s. Mesh-based Patch Warping}. For point-based patch warping (left), the patch is created by the position of the center pixel and the corresponding normal, which can not represent real geometric information. While for mesh-based patch warping (right), each patch pixel maps to the accurate 3D position through rasterization.}
	\label{fig:patch}
	\vspace{-0.2in}
\end{figure}

In order to enforce multi-view photometric consistency, we introduce a mesh-based patch warping strategy. Traditional point-based patch warping relies on the center pixel and normal vector to define a plane, which is then transformed into the source image through homography transformation. However, it is important to note that the resulting plane may not accurately capture the real geometry of the object. By utilizing differentiable Poisson reconstruction and mesh representation, we can determine the accurate 3D position of each pixel within the patch and then transform it into the source image via inverse projection. Fig.~\ref{fig:patch} illustrates the contrast between the two methods.

We define a patch on an image as $\boldsymbol{p}$, and the color information in the patch is supposed to be consistent among different views. We obtain the exact 3D position corresponding to every pixel through the renderer $\zeta$
\begin{equation}
    \hat{P} = \zeta(V,F,TV;\pi)
\end{equation}
where $\hat{P}$ is the rendered position map. Each valid pixel of $\hat{P}$ represents the corresponding 3D location in the camera coordinate. The third dimension of $\hat{P}$ represents the rendered depth map $\hat{D}$. We warp the patch $\boldsymbol{p}$ on the reference frame to the source frame
\begin{equation}
    H_{r \rightarrow s}(\boldsymbol{p}) = \pi_{s}({\pi}_{r}^{-1}(\boldsymbol{p}))
\end{equation}
\begin{equation}
    \hat{P}_{s}(\boldsymbol{p}) = \mathcal{I}(\hat{P}_{s}, H_{r \rightarrow s}(\boldsymbol{p}))
    \label{equ:warp},
\end{equation}
where $\hat{P}_r(\boldsymbol{p})$ and $\hat{P}_s(\boldsymbol{p})$ indicate the 3D position of reference patch and source patch, respectively. we use $s$ and $r$ as subscript to represent source and reference images. $H_{r \rightarrow s}(\boldsymbol{p})$ represents the source patch reprojected from reference patch. $\mathcal{I}$ is the bilinear interpolation operation. Grayscale and depth of source patch also can be obtained by the formula similar to Eq.~\ref{equ:warp}.
\begin{equation}
    G_s(\boldsymbol{p}) = \mathcal{I}(G_{s}, H_{r \rightarrow s}(\boldsymbol{p}))
\end{equation}
\begin{equation}
    D_s(\boldsymbol{p}) = \mathcal{I}(D_{s}, H_{r \rightarrow s}(\boldsymbol{p}))
\end{equation}
We convert color images $\{I_i\}$ into grayscale images $\{G_i\}$, and maximize the normalized cross-correlation (NCC) to ensure multi-view photometric consistency
\begin{equation}
    \text{NCC}(G_r(\boldsymbol{p}), G_s(\boldsymbol{p})) = \frac{\text{Cov}(G_r(\boldsymbol{p}), G_s(\boldsymbol{p}))}{\sqrt{\text{Var}(G_r(\boldsymbol{p}) \text{Var}(G_s(\boldsymbol{p}))}}
\end{equation}
where $\text{Cov}$ is covariance and $\text{Var}$ is variance. $G_r(\boldsymbol{p})$ and $G_s(\boldsymbol{p})$ represent the gray value of reference patch and source patch, respectively. NCC scores are computed between the sampled reference and source patches on all source images. 

To avoid corresponding patches from occlusion, we compare the the rendered patch depth and the reprojected patch depth, and discard the patches that differ widely. To further guarantee the patches are visible on all source views, we only consider those patches whose NCC scores are above a certain threshold. We impose the multi-view photometric consistency loss on the mesh vertices, and backprop to update the oriented point cloud
\begin{equation}
    \mathcal{L}_\text{ncc} = \left\{
    \begin{aligned}
    & 1 - \text{NCC}(G_r(\boldsymbol{p}), G_s(\boldsymbol{p})) \quad \delta > 0 \\
    & 0 \quad \text{else}
    \end{aligned}
    \right
    .
    \label{equ:ncc}
\end{equation}
\begin{equation}
    \begin{aligned}
    \delta = & (|T_s T_r^{-1} K_r^{-1} \boldsymbol{p} - \hat{D}_{s}(\boldsymbol{p})| < \delta_d) \\& \times (\text{NCC}(G_r(\boldsymbol{p}), G_s(\boldsymbol{p})) > \delta_\text{ncc})
    \end{aligned}
\end{equation}
where $\delta_d$ is depth threshold and $\delta_{\text{ncc}}$ is NCC threshold. $\hat{D}_{s}(\boldsymbol{p})$ and $T_s T_r^{-1} K_r^{-1} \boldsymbol{p}$ represent the interpolated patch depth and the reprojected patch depth, respectively. $K$ and $T$ represent intrinsic matrix and extrinsic matrix.

\subsection{Shape from Shading Refinement}\label{sec:sfs} 
Compared to the initial mesh from visual hull, Multi-view photometric consistency helps to obtain decent geometric details through patch warping. In order to further obtain more fine details on a per-pixel level and recover albedos, we employ SFS refinement discussed as following. In general, the color of human skin and clothes mainly have diffuse reflection, which fits the assumption of shape from shading (SFS) algorithm. Once the mesh is obtained from previous subsection, an SFS refinement is employed to improve the mesh and extract albedos from multi-view images. We firstly review the image formation model using the SH illumination, and then propose the geometry refinement and albedo extraction method in detail.

\noindent\textbf{Image Formation Model}
When objects in a scene are non-emitters and the light source are infinitely distant, the image irradiance equation can be defined as in~\cite{DBLP:conf/siggraph/req86},
\begin{equation}
    B(x,w_o) = \int_{\Omega} L(w_{i}) \rho(w_{i}, w_{o})\max(w_i \cdot \boldsymbol{n}_{x}, 0) dw_{i},
    \label{equ:req}
\end{equation}
where $B(x,w_o)$ is the reflected radiance. $x,\boldsymbol{n},w_i,w_o$ are the spatial location, surface normal, incident light direction and viewing direction, respectively. The domain of integral $\Omega$ is a semi-sphere centered at $x$. $L(w_{i})$ is the light intensity from direction $w_i$. $\rho(w_{i}, w_{o})$ is the bidirectional reflectance distribution function (BRDF) of the surface. With the assumption of Lambertian Surface, the reflection is constant from all directions of views.

We make use of 3 rd sphere harmonic (SH) coefficients to represent the general lighting. Due to the orthogonality of the SH basis, Eq.~\ref{equ:req} can be derived as below
\begin{equation}
    B(x) = \rho_{x} \sum_{i=1}^{3^2} l_i Y_i(\boldsymbol{n}_x)
\end{equation}
where $\rho_{x}$ is the albedo at point $x$. $l_i$ is SH coefficients. $Y_i$ is the SH function determined by the surface normal $\boldsymbol{n}$.

We estimate the $l_i$ according to the mesh from subsection by minimizing the difference between the image density and the computed image irradiances
\begin{equation}
    \hat{\boldsymbol{l}} = \mathop{\arg\min}\limits_{\boldsymbol{l}} \sum\limits_{x} || \sum_{i=1}^{n^2} l_i Y_i(\hat{\boldsymbol{N}}(\pi(x))) - G(\pi(x)) ||_2^2
\end{equation}
\begin{equation}
    \hat{\boldsymbol{N}} = \zeta(V,F,V_n;\pi)
\end{equation}
We interpolate the vertex normal $V_n$ to build the normal map $\hat{\boldsymbol{N}}$. For each valid pixel and its corresponding point $x$, we try to minimize the $L_2$ norm between the grayscale value and the computed irradiance to obtain the SH coefficients. Since this is an overdetermined problem, we use least squares to estimate SH coefficients.

Once the SH coefficients are estimated, we fix them to refine the coarse mesh and extract albedo. We first extract albedo from captured images, and then refine the albedo and geometry jointly.
\begin{equation}
    \hat{A} = \zeta(V,F,V_\text{albedo};\pi)
\end{equation}
\begin{equation}
    \mathcal{L}_\text{sfs} = \sum\limits_{x} | \hat{A}(\pi(x)) \sum_{i=1}^{n^2} l_i Y_i(\hat{\boldsymbol{N}}(\pi(x)) - I(\pi(x)) |
\end{equation}
where $\hat{A}$ is the interpolated albedo map. $| \cdot |$ denotes $L_1$ norm. In order to prevent overfitting or getting stuck at the local optima, we introduce the regularization terms to penalize the surface deformations and texture consistency
\begin{equation}
\begin{aligned}
\mathcal{L}_\text{reg} &= \mathcal{L}_\text{mesh} + \mathcal{L}_\text{albedo} \\
&= |LV| + |LV_\text{albedo}|
\end{aligned}
\end{equation}
where $L$ denotes the Laplacian matrix. $\mathcal{L}_\text{reg}$ forces that the adjacent vertices have similar positions and colors.

\subsection{Implementation Details}\label{sec:implement}
We firstly extract initial mesh via visual hull at a grid resolution of $128^3$. Secondly, we uniformly sample 50k oriented points as our shape representation, and the silhouette and NCC loss in Eq.~\ref{equ:mask},~\ref{equ:ncc} are backpropagated to update the point cloud. For the patch warping loss $\mathcal{L}_{ncc}$, we pre-select 4 adjacent images for each reference image as source images. We perform the optimization at the resolution of $512^3$ and the degree of gaussian smoothing $sig=4$ for 10 epochs. The NCC threshold is $\delta_\text{ncc}=0.5$, and the depth threshold is $\delta_d=0.01$. The patch size is $11\times11$. For ablation studies in the number of oriented points and patch size, please refer to supplementary materials. 
As done in ~\cite{Peng2021SAP}, we resample points and normals every other epoch in order to increase the robustness of the optimization process. The weights for the loss terms are $\lambda_{sil} = 20$, $\lambda_\text{ncc} = 5$. we use Adam optimizer~\cite{DBLP:journals/corr/adam14} for optimization, and the learning rate for updating oriented point clouds is $1e^{-3}$. 
In SFS refinement stage, mesh is exported from point cloud and the topology is fixed. we firstly optimize vertex albedo for 200 epochs at the learning rate of $1e^{-2}$. Then, we refine the mesh and albedo simultaneously for another 100 epochs. The learning rate for vertices and albedo are $1e^{-3}$ and $5e^{-3}$, respectively. The weights for the loss terms are $\lambda_\text{sfs} = 20$, $\lambda_\text{mesh} = 50$ and $\lambda_\text{albedo} = 1$, respectively.

\begin{table*}
	\centering
	\begin{tabular}{l|ccc|c|c}
		\toprule
		Methods & Normal C. & Chamfer-$L_1$ & PSNR & Optimization Time & Rendering Time\\ 
		\midrule
		NeRF~\cite{DBLP:conf/eccv/nerf20} & 0.43 & 2.32 & 26.43 & 126 min & 3.0 s \\
        Colmap~\cite{DBLP:conf/eccv/colmap16} & 0.18 & 0.36 & - & 5 min & - \\
		IDR~\cite{DBLP:conf/nips/idr20} &  0.12 & 0.34 & 24.46 & 63 min &  2.5 s \\
		NeuS~\cite{DBLP:conf/nips/neus21} & 0.11 & 0.34 & \textbf{30.64} & 338 min & 56 s \\
		NeuralWarp~\cite{DBLP:conf/cvpr/neuralwarp22} & 0.14 & 0.40 & 23.39 & 245 min & 63 s \\
        NeuS NGP~\cite{DBLP:conf/nips/neus21,mueller2022instant} & 0.15 & 0.48 & 29.31 & 8 min & 0.05s \\
		DiffuStereo~\cite{shao2022diffustereo} & 0.17 & 0.41 & - & \textbf{1 min} & - \\
		FastHuman (Ours) &  \textbf{0.06} & \textbf{0.18} & 30.58 & 6 min & \textbf{0.01 s} \\
		\bottomrule
        \end{tabular}
        \caption{\textbf{Quantitative Comparison on Our Synthetic Dataset.} We report a quantitative comparison of 3D reconstruction from multi-view images, and the numbers are average from 40 scans. Compared to baselines, our method can attain better reconstruction quality and also high-quality rendering.
        Moreover, our optimization speed is significantly faster than neural implicit-based approaches and on part with method with pretraining~\cite{shao2022diffustereo} or optimized traditional pipeline~\cite{DBLP:conf/eccv/colmap16}. Moreover, We show the possibility of real-time rendering at 100 FPS.}
	\label{tab:rp}
\end{table*}

\section{Experiments}

In this section, we first present the experimental results for reconstructing human body from multi-view images. We compare our proposed method with the current state-of-the-art techniques, and demonstrate that it is also applicable to multi-view videos. We also conduct ablation studies to evaluate the effectiveness of multi-view photometric consistency and shading refinement in our approach.

\subsection{Results on Multi-view Images}

Since the existing multi-view human datasets do not have ground truth 3D meshes, we render meshes to generate multi-view images for the quantitative comparison. We collect 40 high-resolution photogrammetry scans from RenderPeople~\cite{renderpeople}, and render these meshes using the off-the-shelf software Blender~\cite{blender}. For each scan, we render 19 images in a circle around the mesh with the resolution of $1024 \times 1024$.

We compare our proposed method against recent multi-view human reconstruction~\cite{shao2022diffustereo}, multi-view scene reconstruction~\cite{DBLP:conf/eccv/nerf20, DBLP:conf/nips/idr20, DBLP:conf/nips/neus21,DBLP:conf/cvpr/neuralwarp22} and colmap~\cite{DBLP:conf/eccv/colmap16}. We also compare to NeuS NGP, where we run NeuS with multi-res feature grids from this repo\footnote{https://github.com/bennyguo/instant-nsr-pl} with our data. Since PIFu~\cite{DBLP:conf/iccv/PIFu19} and PIFuHD~\cite{DBLP:conf/cvpr/pifuhd20} are proposed for human reconstruction from a single image, they do not consider the camera information. We do not compare with PIFu and PIFuHD. Similar to PIFu, we adopt three reconstruction performance metrics including normal projection error, Chamfer distance and PSNR.

Table~\ref{tab:rp} shows the quantitative results. Our proposed approach achieves the lowest Chamfer distance and normal projection error. The PSNR results show that our extracted albedo is able to render the photo-realistic images. We employ Marching Cube on the volume density estimated by NeRF to extract mesh and remove the extra surfaces according to masks. The results of Colmap and Diffustereo are in point clouds form, we employ Screened Poisson surface Reconstruction~\cite{DBLP:journals/tog/spr13} and remove outliers by masks. Since DiffuStereo has to use DoubleField~\cite{DBLP:conf/cvpr/doublefieldL22} to generate coarse mesh and normals as input while the implementation of DoubleField is not publicly available. We use the coarse mesh optimized by our proposed multi-view photometric constraints as the input, and fuse the point clouds extracted from all views. 

\begin{figure*}
    \centering
    \begin{tabular}{@{\hskip2pt}c@{\hskip2pt}@{\hskip2pt}c@{\hskip2pt}@{\hskip2pt}c@{\hskip2pt}@{\hskip2pt}c@{\hskip2pt}@{\hskip2pt}c@{\hskip2pt}@{\hskip2pt}c@{\hskip2pt}@{\hskip2pt}c@{\hskip2pt}}
    \includegraphics[width=0.13\textwidth,trim=450 250 250 140,clip]{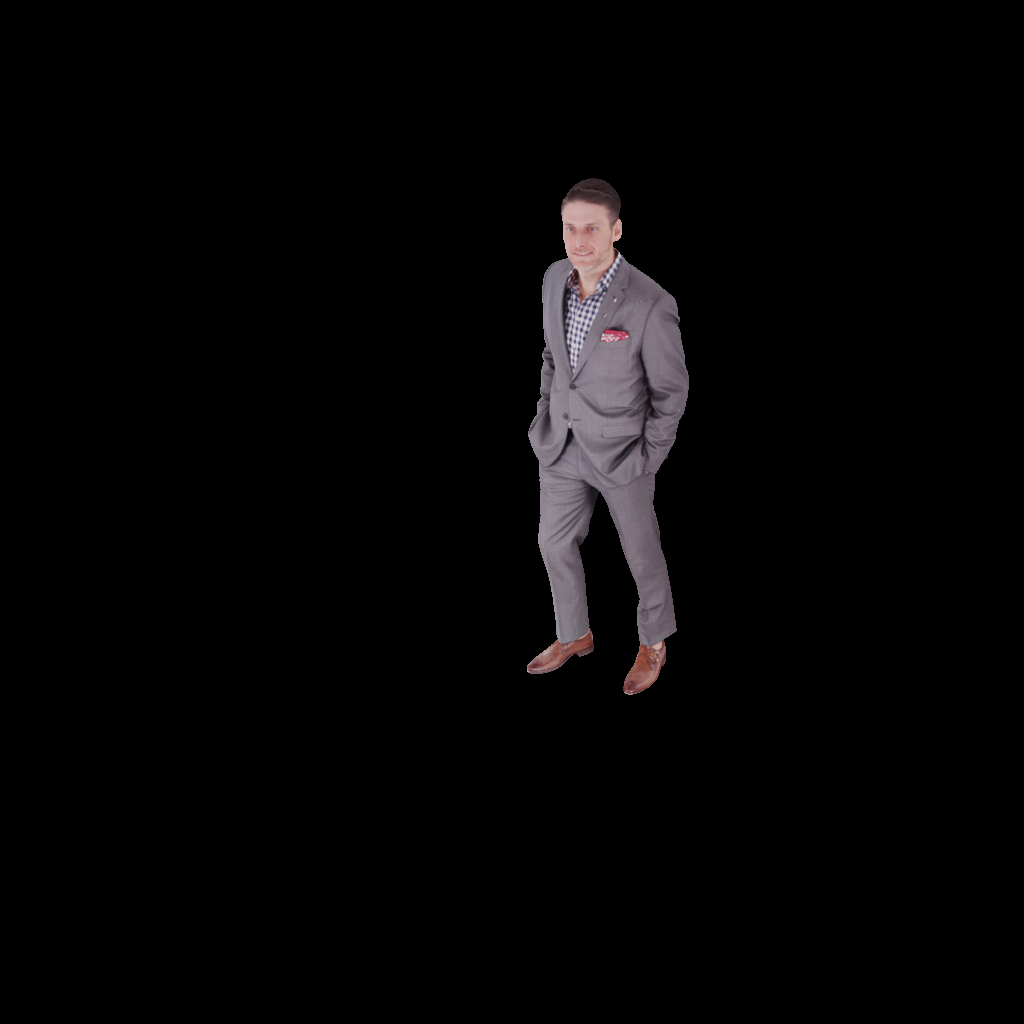} & 
    \includegraphics[width=0.13\textwidth,trim=450 250 250 140,clip]{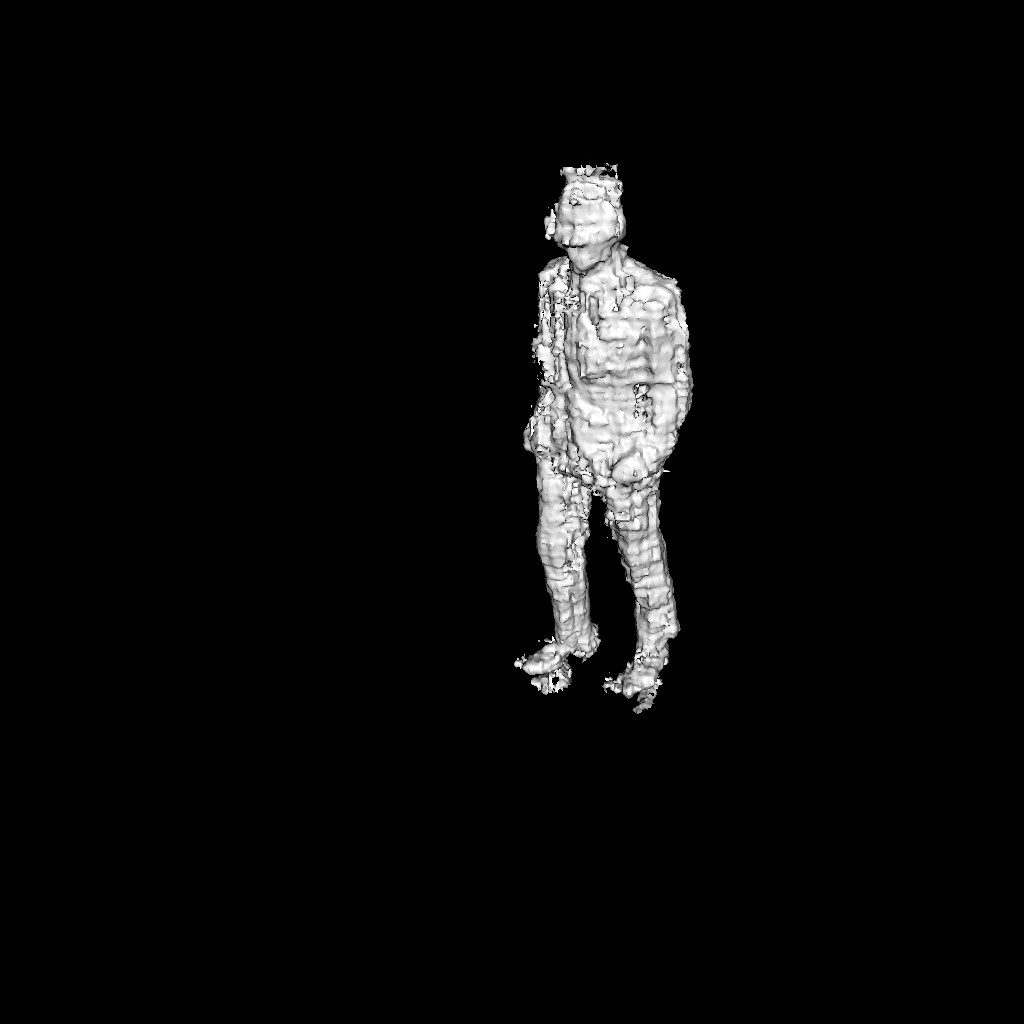} &
    \includegraphics[width=0.13\textwidth,trim=450 250 250 140,clip]{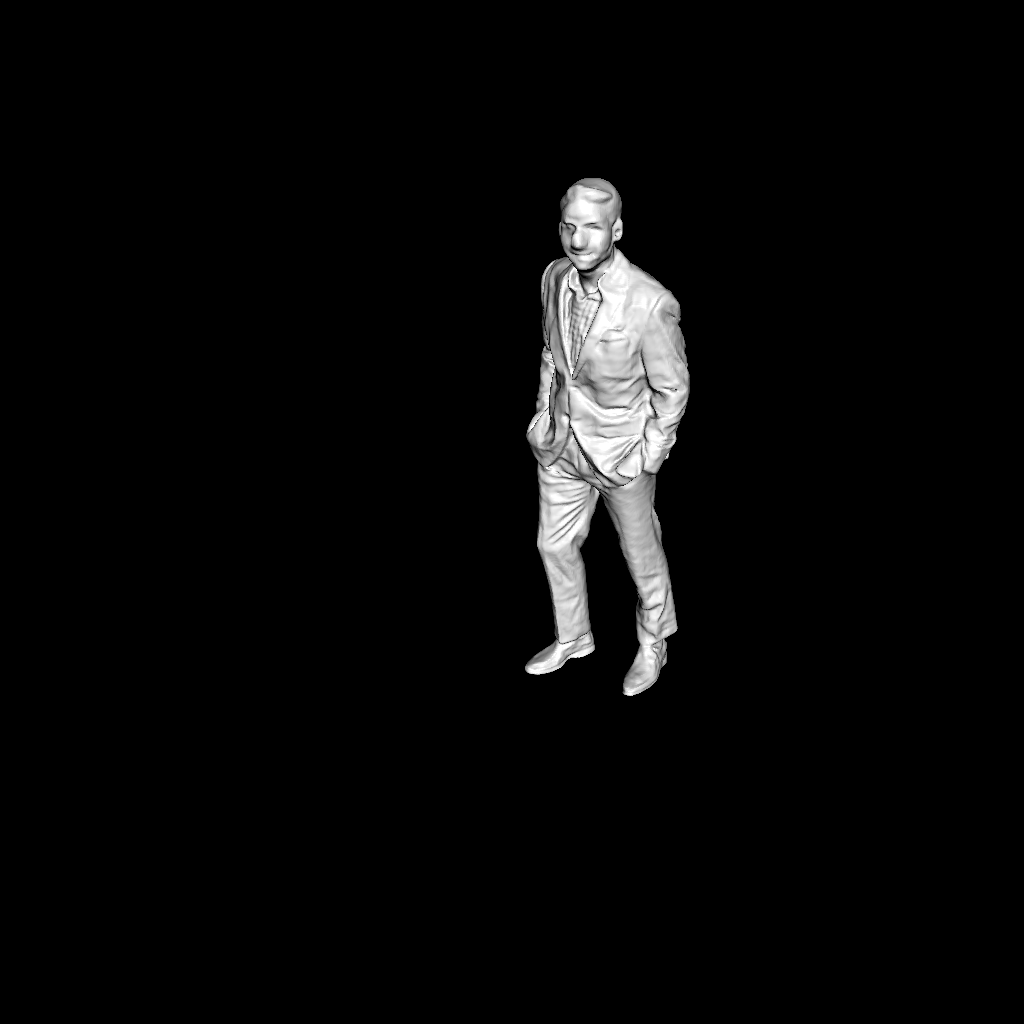} &
    \includegraphics[width=0.13\textwidth,trim=450 250 250 140,clip]{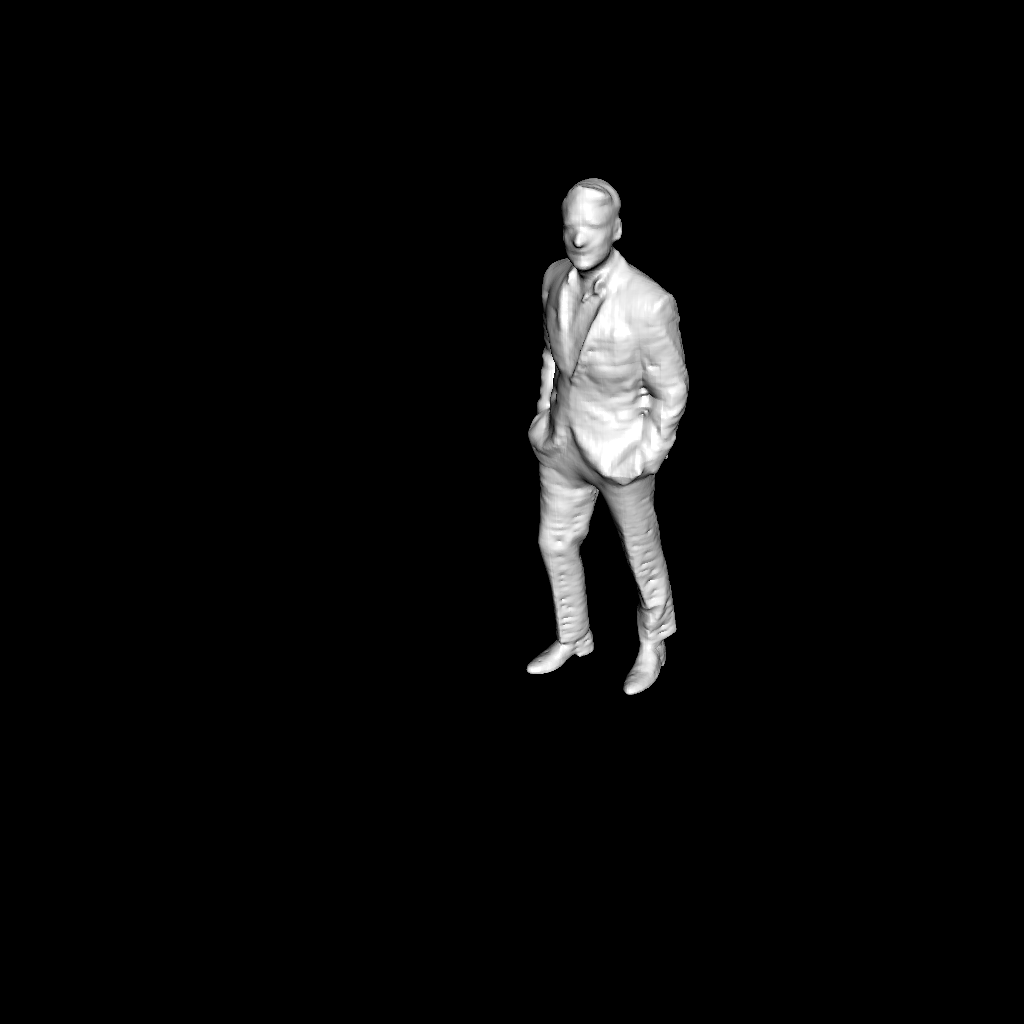} &
    \includegraphics[width=0.13\textwidth,trim=450 250 250 140,clip]{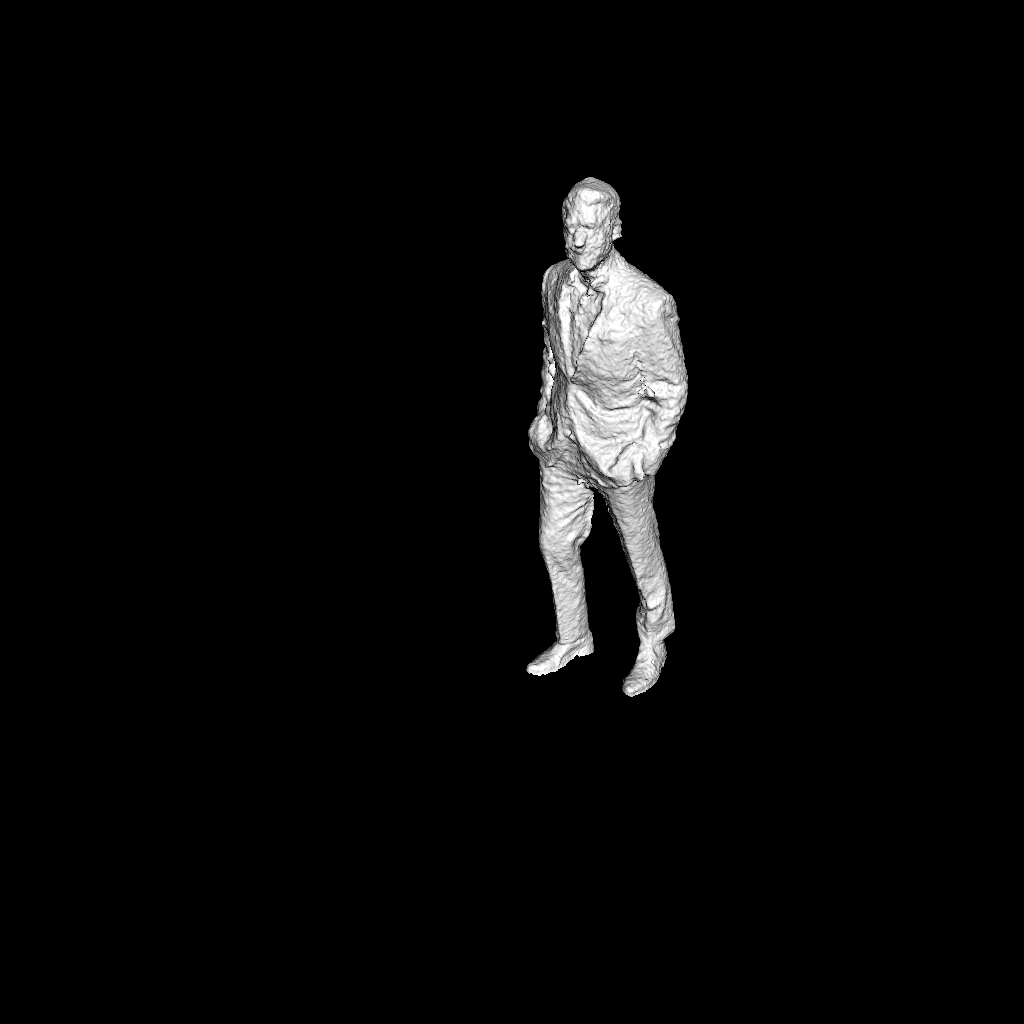} &
    \includegraphics[width=0.13\textwidth,trim=450 250 250 140,clip]{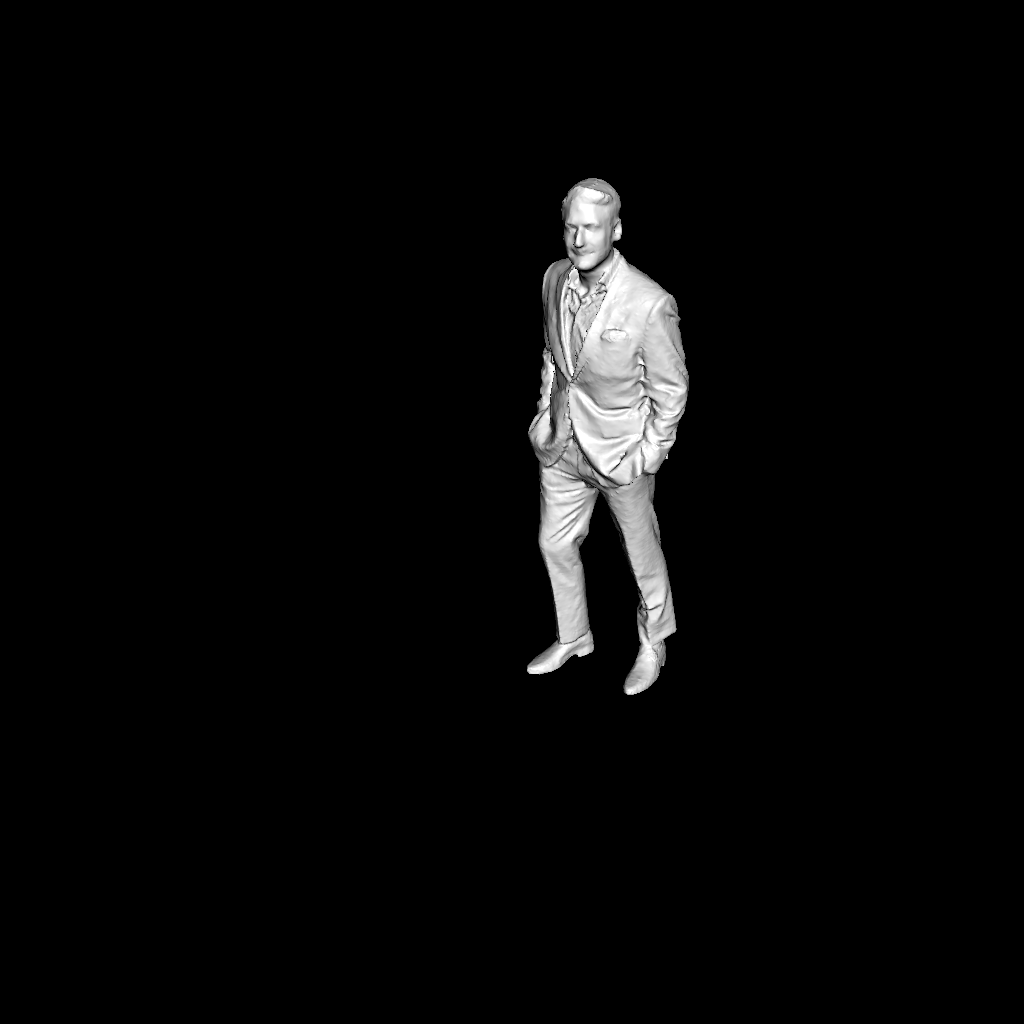} &
    \includegraphics[width=0.13\textwidth,trim=450 250 250 140,clip]{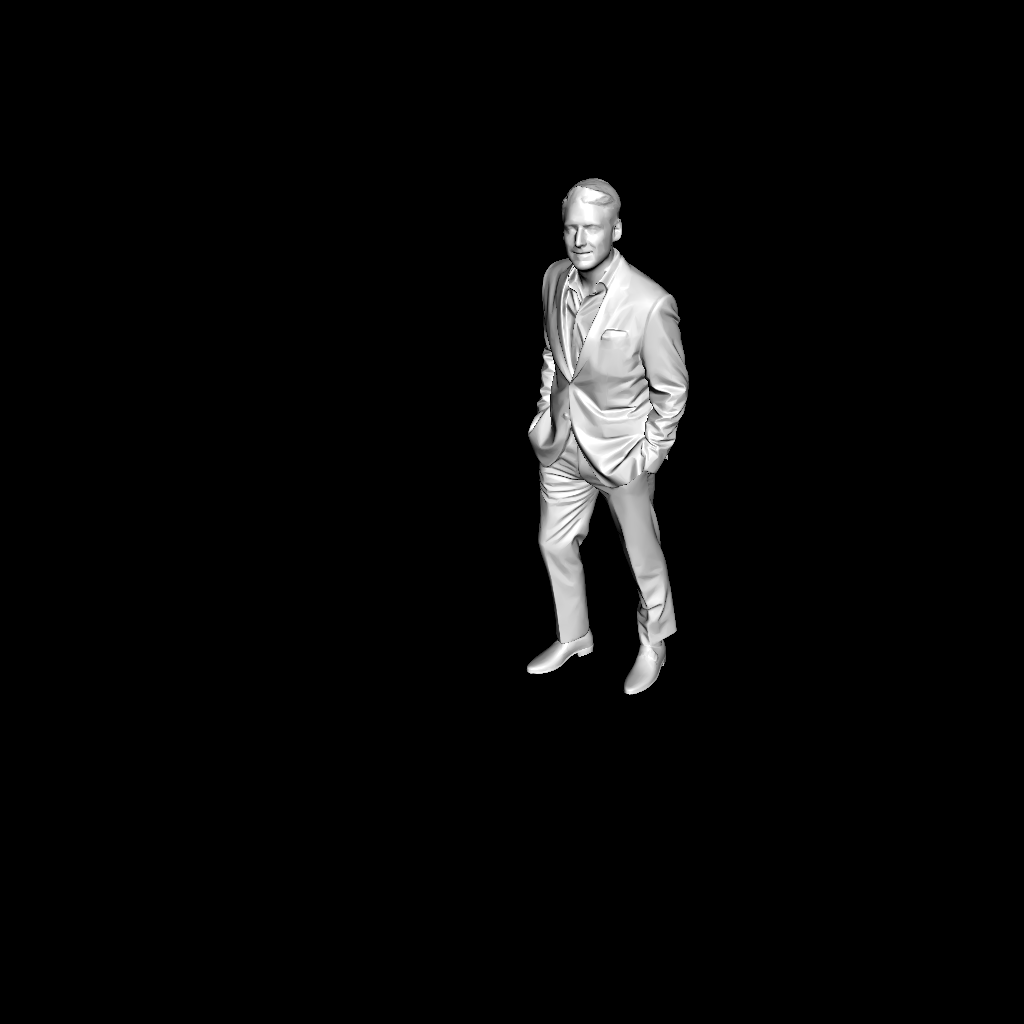} \\ 
    \includegraphics[width=0.13\textwidth,trim=450 250 250 140,clip]{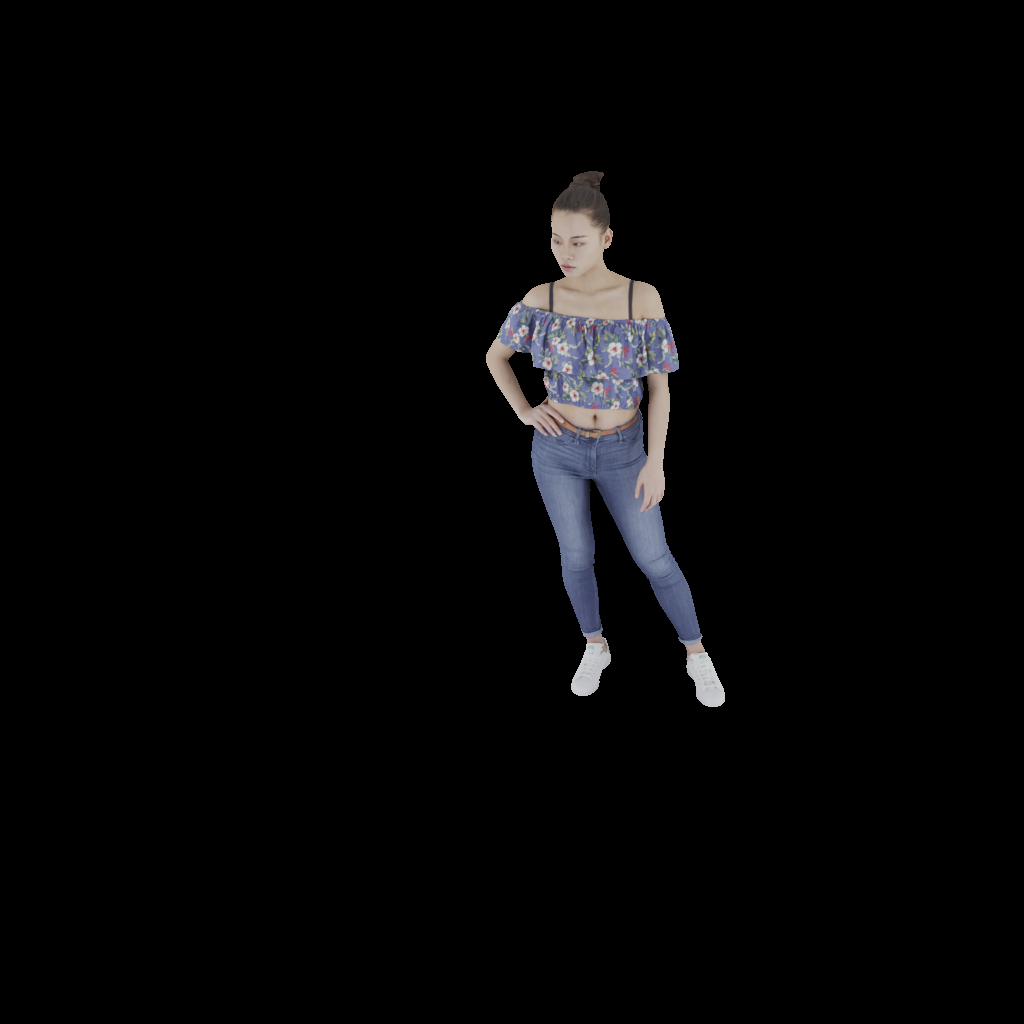} &
    \includegraphics[width=0.13\textwidth,trim=450 250 250 140,clip]{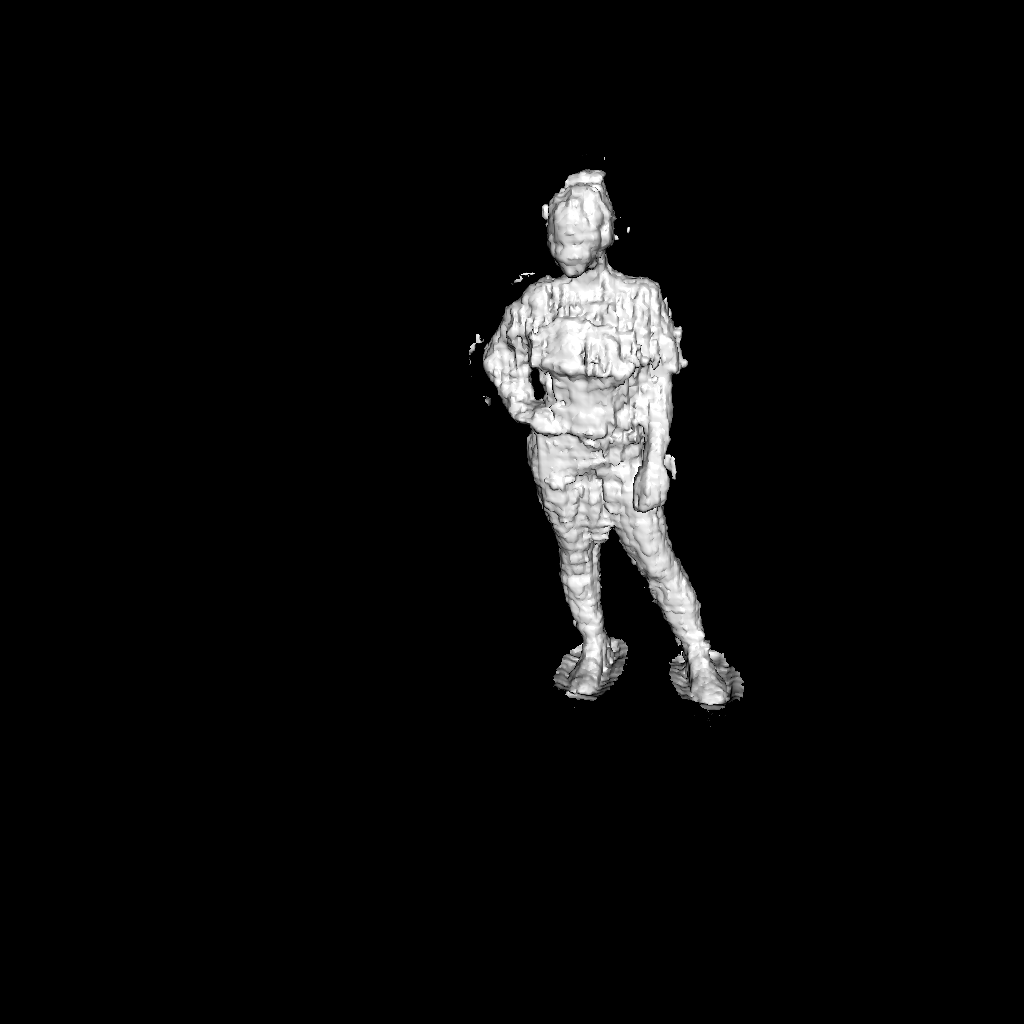} &
    \includegraphics[width=0.13\textwidth,trim=450 250 250 140,clip]{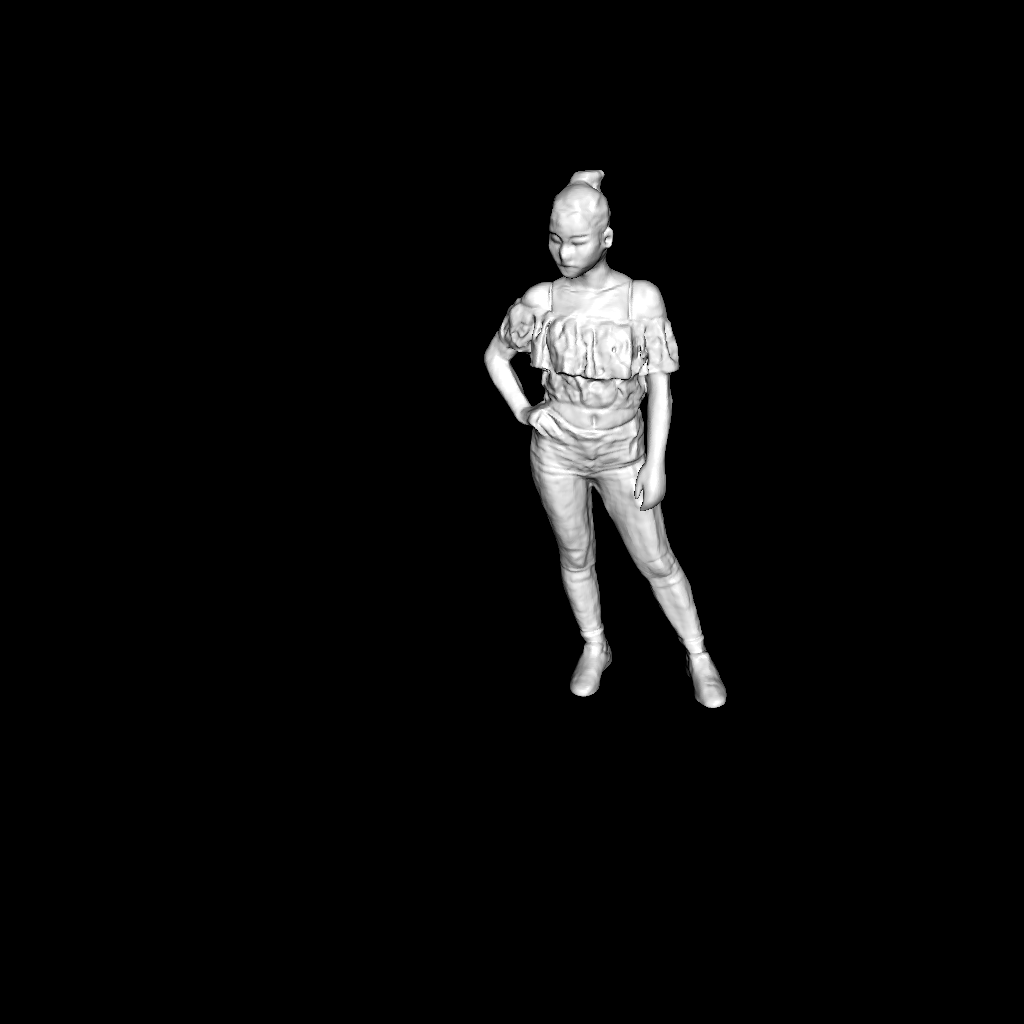} &
    \includegraphics[width=0.13\textwidth,trim=450 250 250 140,clip]{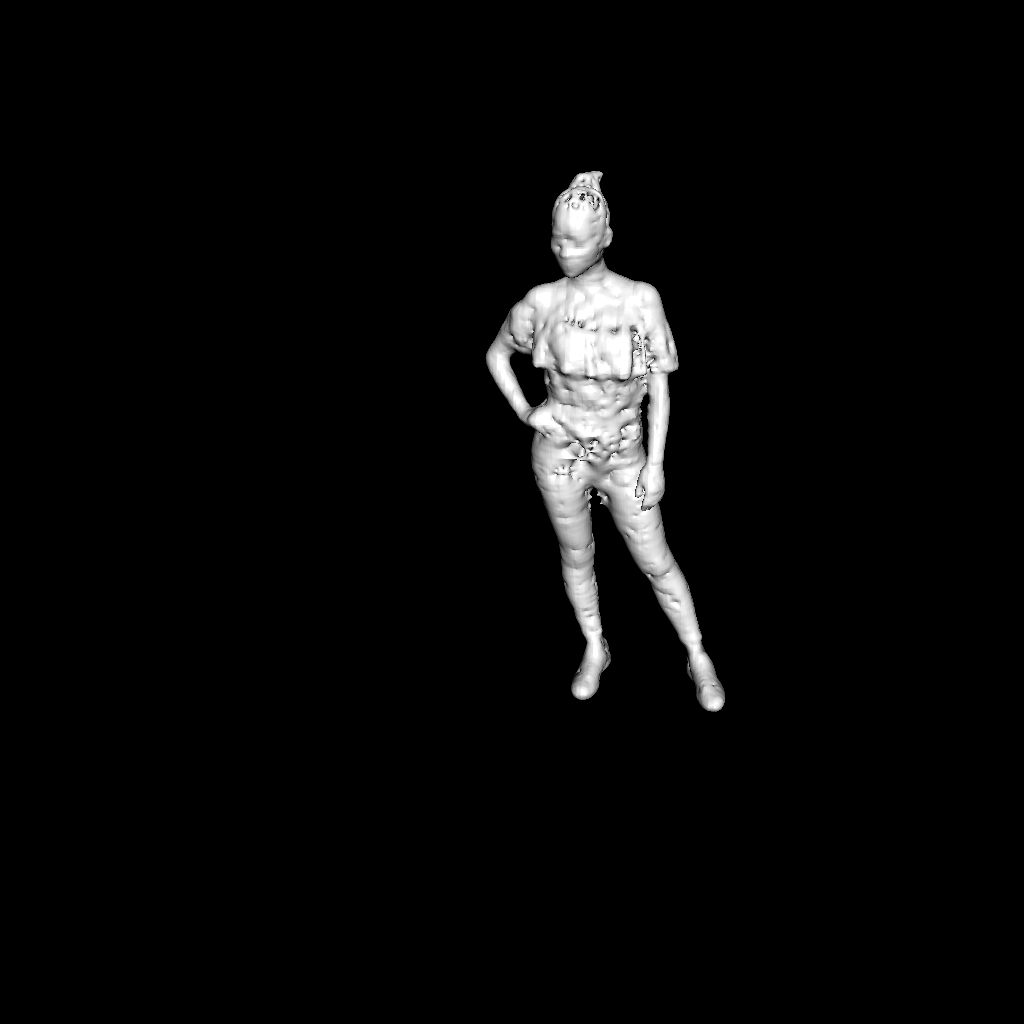} &
    \includegraphics[width=0.13\textwidth,trim=450 250 250 140,clip]{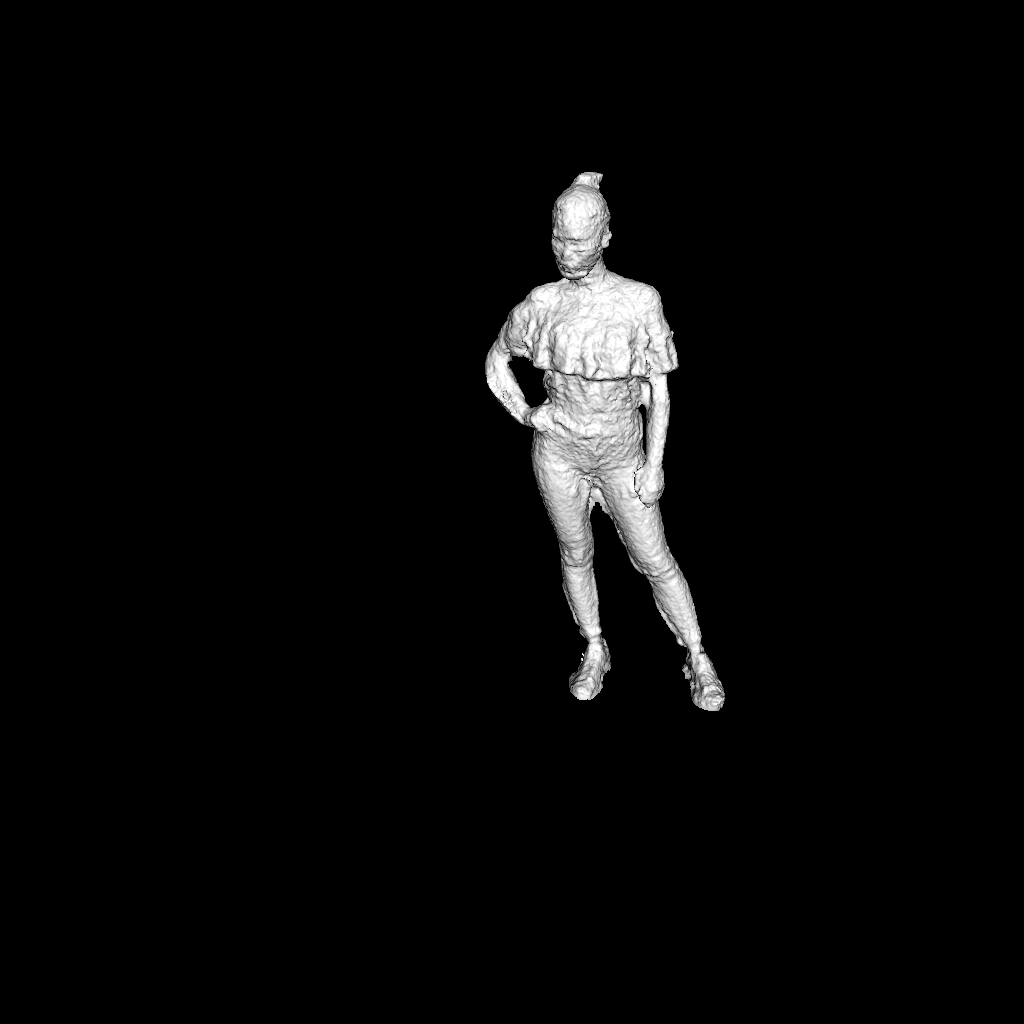} &
    \includegraphics[width=0.13\textwidth,trim=450 250 250 140,clip]{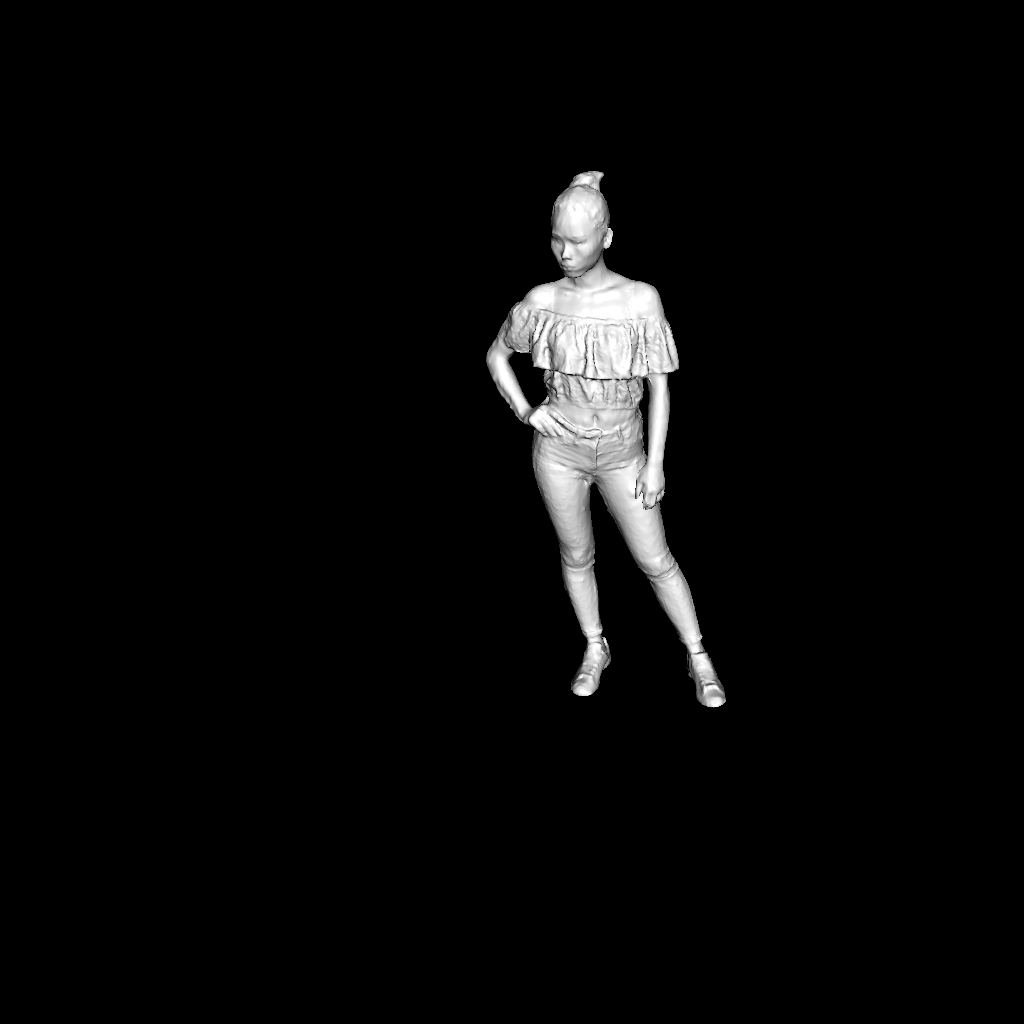} &
    \includegraphics[width=0.13\textwidth,trim=450 250 250 140,clip]{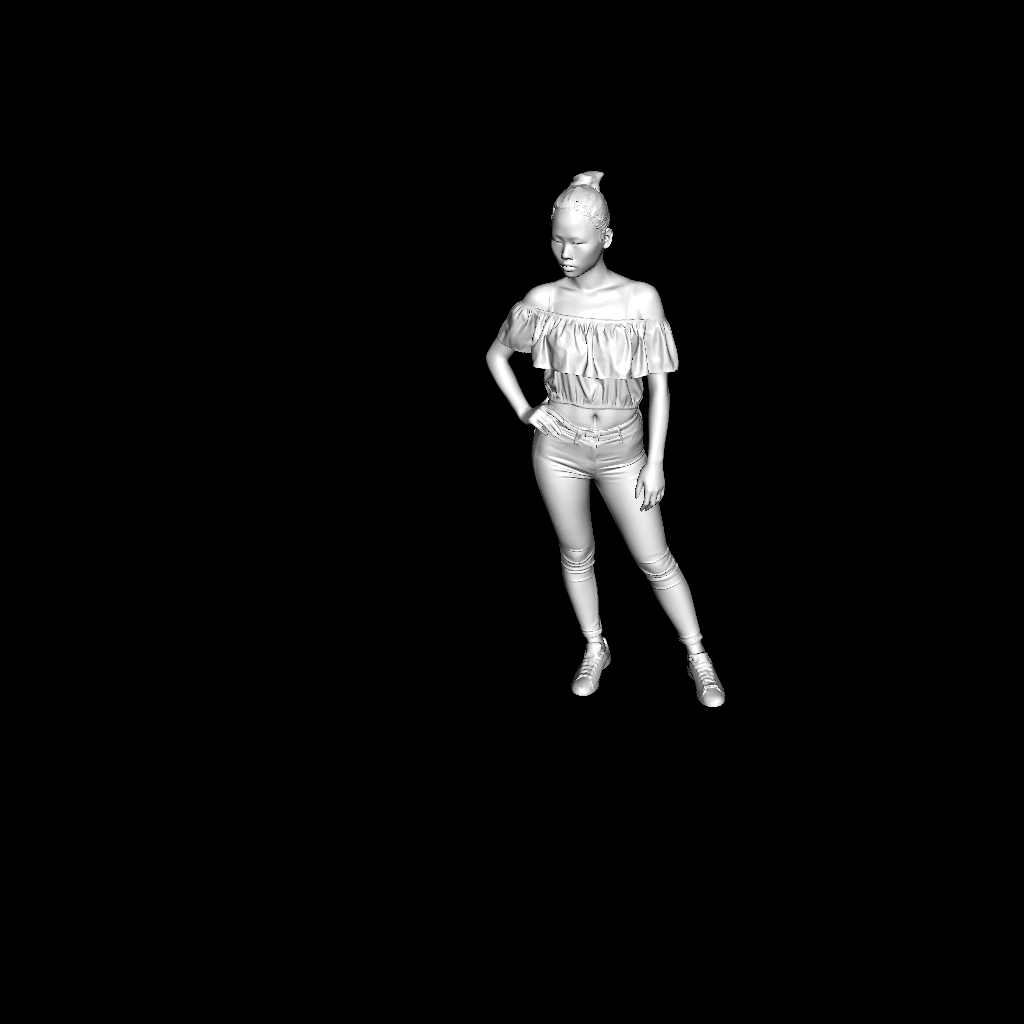} \\
    GT View & NeRF~\cite{DBLP:conf/eccv/nerf20} & NeuS~\cite{DBLP:conf/nips/neus21} & NeuS NGP & Diffustereo~\cite{shao2022diffustereo} & \textbf{Ours} & GT Mesh
    \end{tabular}
    \caption{\textbf{Qualitative Comparison on Our Synthetic Dataset.} We show a qualitative comparison of recosntructed surfaces from multi-view images. Compared to baselines, our reconstructions capture most geometric details.}
    \label{fig:ret}
\end{figure*}

The reconstructed mesh of NeRF is very rough and inaccurate. NeuS renders high-fidelity images. However, it is difficult for rendering loss to handle the ambiguity between appearance and geometry. The reconstructed mesh is different from the ground truth. We use the pre-trained Diffustereo, whose number of cameras and camera positions are different from the training data. Thus, the results of Diffustereo are not satisfied. The reason for poor NeuS NGP performance is the sparsity of input views since feature-grid based methods tend to struggle with this level of ambiguity more than MLP-based methods. Fig.~\ref{fig:ret} shows the reconstruction results of various approaches.

Our proposed method also has great advantages in terms of computational time on both optimization and rendering. The optimization time for IDR, NeRF, NeuralWarp, NeuS, NeuS NGP are 1 hour, 2 hours, 4 hours 5.5 hours, and 8 min respectively. Since these neural rendering methods use deep neural network as implicit shape representation, it takes lots of time to reach convergence. Although Diffustereo only takes 1 minute for reconstruction after training, DoubleField is needed to generate the coarse mesh as input. Note that DoubleField requires to be pretrained on a large-scale dataset and fine-tuned for 20 minutes. Colmap needs 5 minutes for dense reconstruction, whose reconstruction results are coarse. It takes 5 minutes for our proposed NCC optimization and 1 minute for SFS refinement. Our proposed method has far less computational time on optimization time, which does not require pre-training on large-scale datasets and fine-tuning. 

In terms of rendering time, Our approach enables real-time rendering by taking advantage of SH illumination and albedo shading model. Moreover, our reconstructed mesh and texture are compatible with the existing rendering engines such like Blender, Unreal Engine~\cite{unreal}, and so on. For the implicit representation-based methods, it takes seconds to render an image, since a forward network inference is required for each valid pixel. Although NeuS NGP can also achieve real-time rendering, it requires much more computations compared to SH shading model. All the experiments about computational time are conducted on the same machine with a single NVIDIA 3090Ti GPU.

\begin{figure}
    \centering
    \begin{tabular}{@{\hskip2pt}c@{\hskip2pt}@{\hskip2pt}c@{\hskip2pt}@{\hskip2pt}c@{\hskip2pt}@{\hskip2pt}c@{\hskip2pt}@{\hskip2pt}c@{\hskip2pt}}
    \includegraphics[width=0.09\textwidth,trim=300 0 400 200,clip]{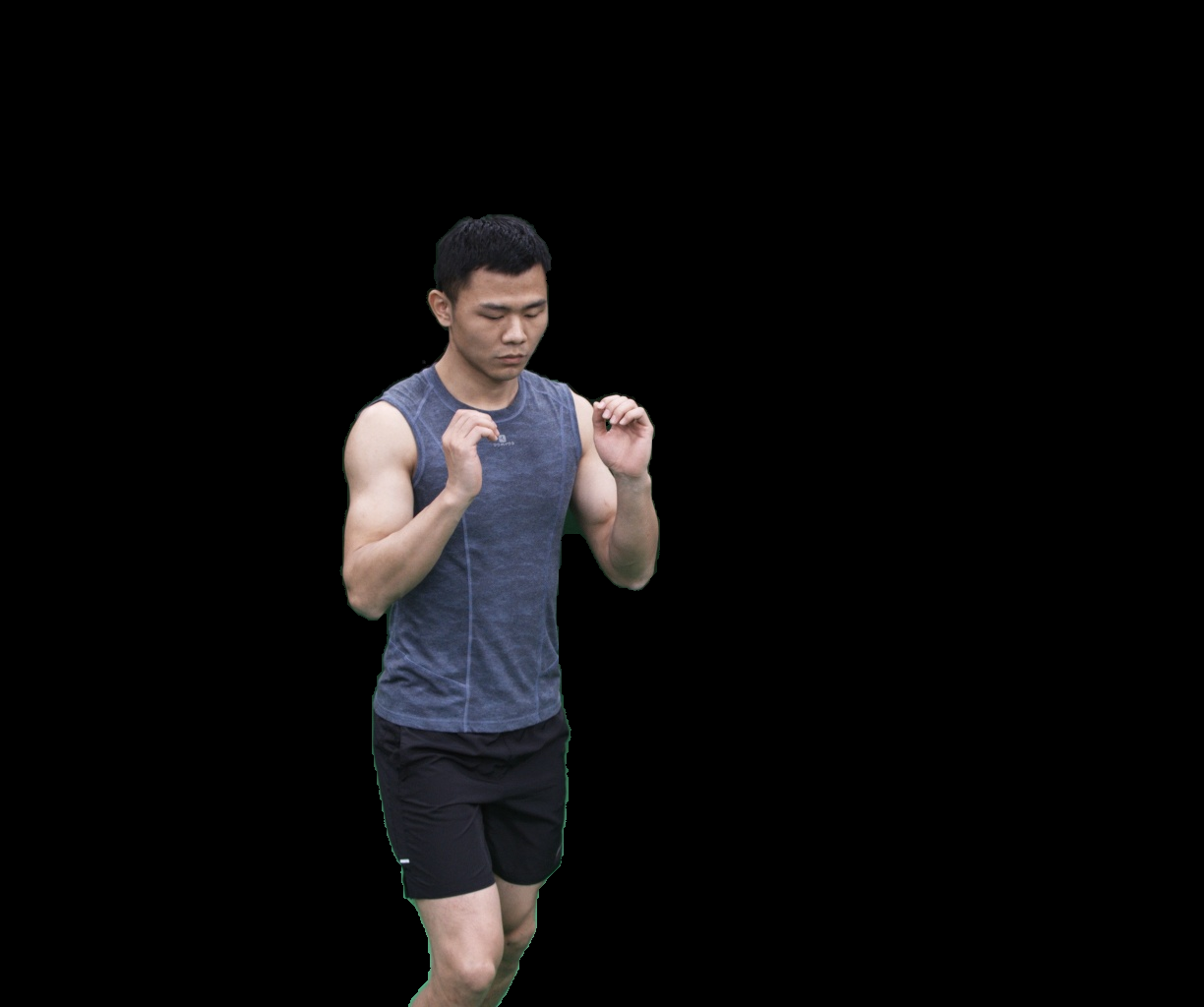} &
    \includegraphics[width=0.09\textwidth,trim=300 0 400 200,clip]{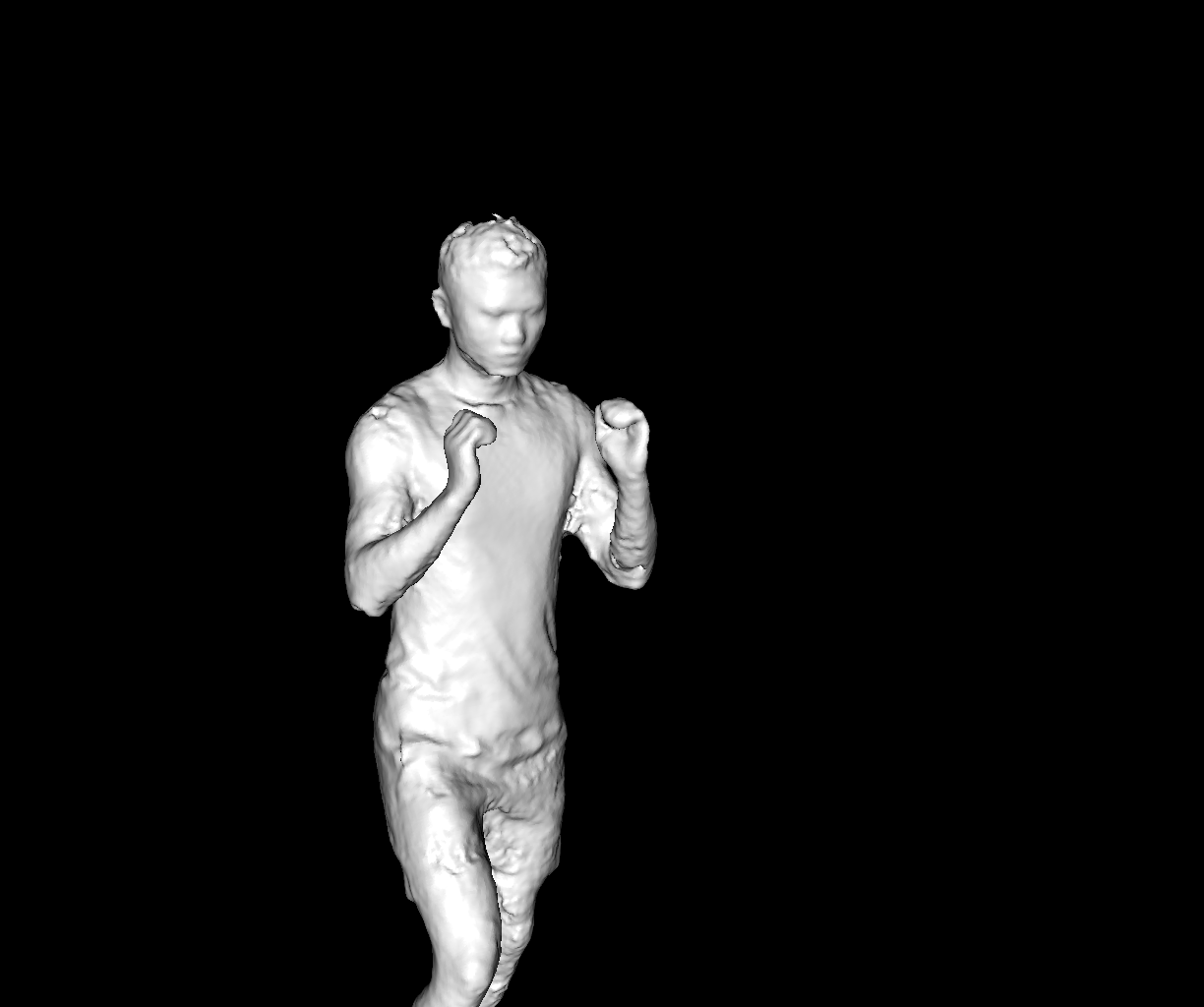} &
    \includegraphics[width=0.09\textwidth,trim=300 0 400 200,clip]{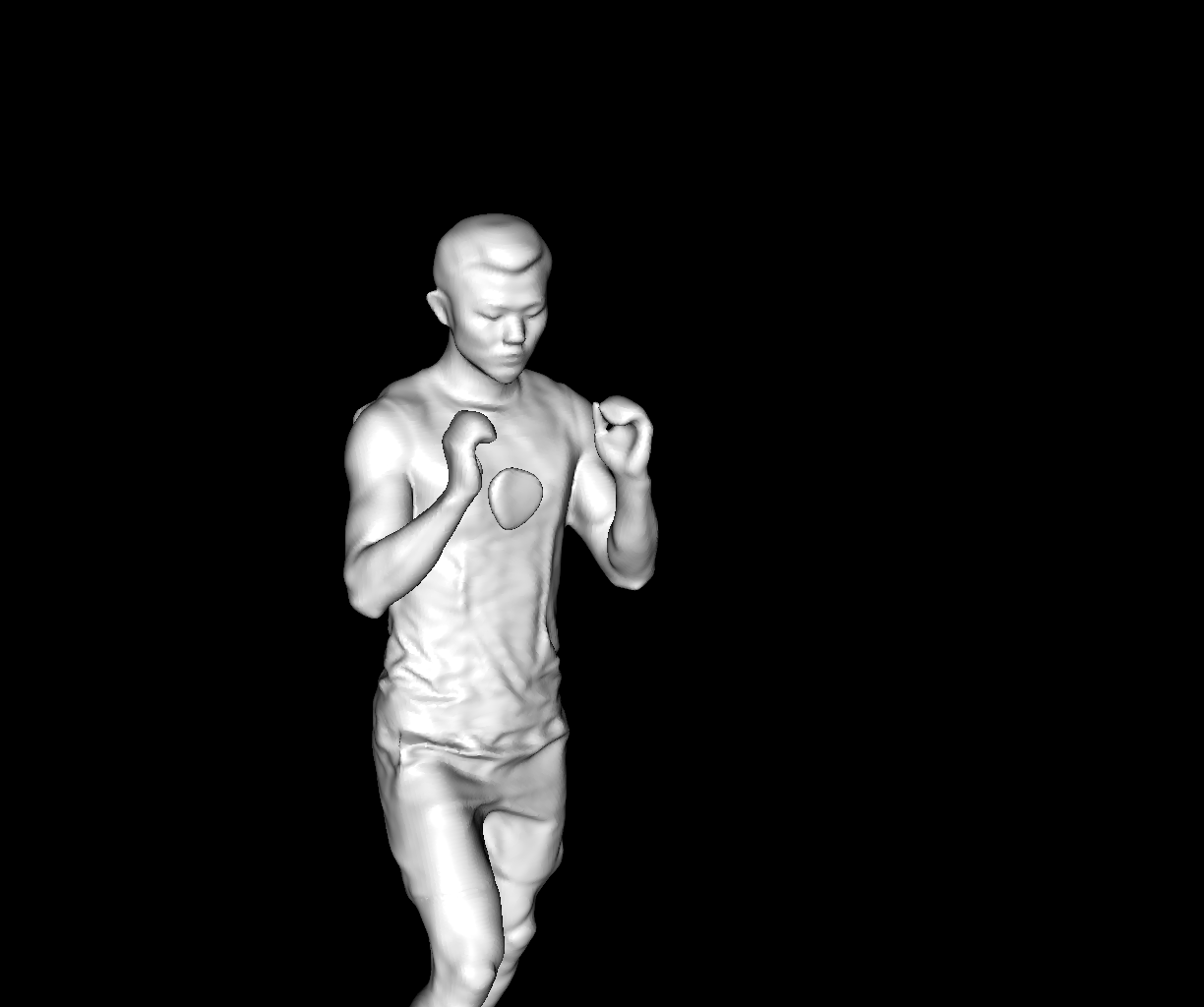} &
    \includegraphics[width=0.09\textwidth,trim=300 0 400 200,clip]{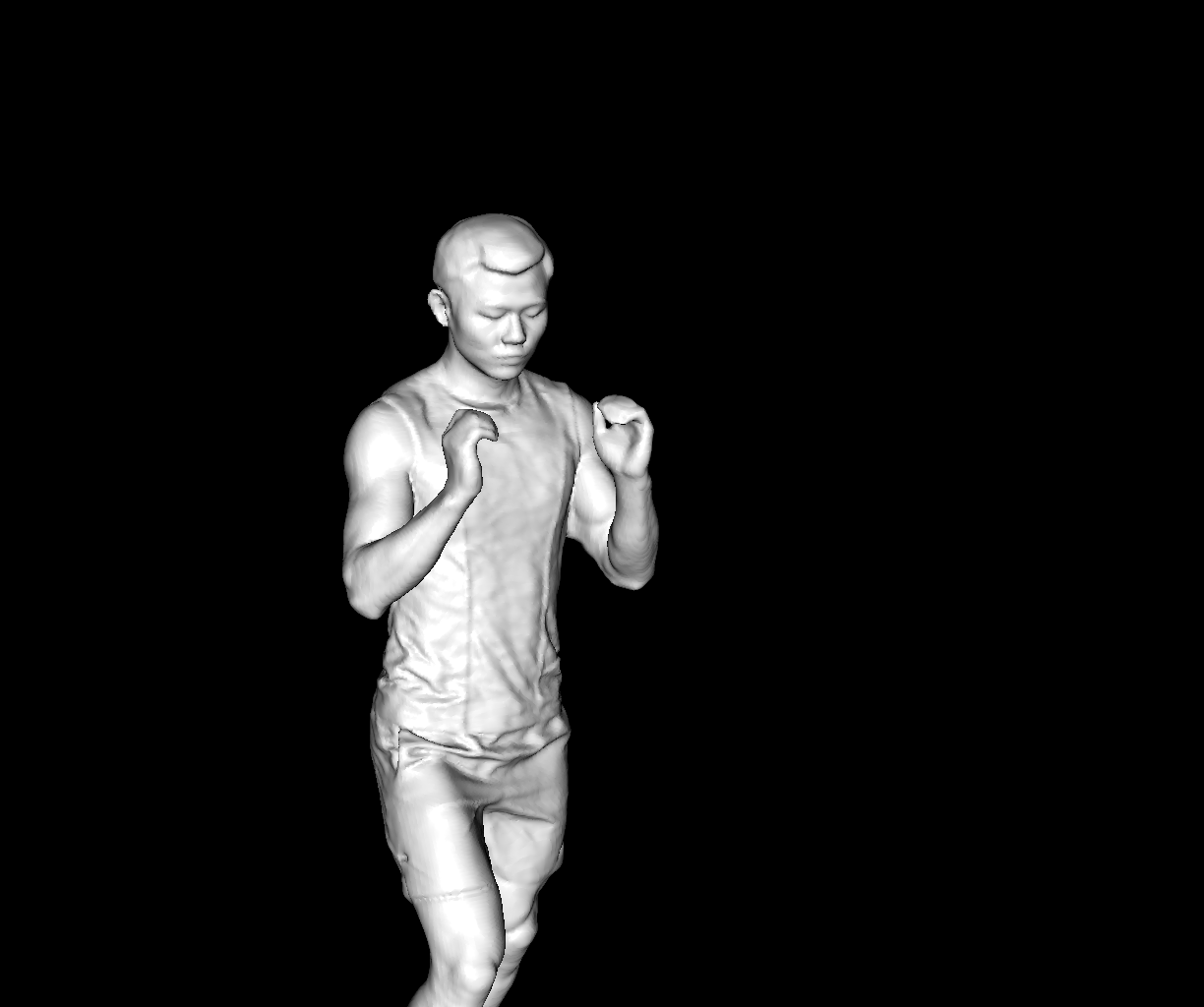} &
    \includegraphics[width=0.09\textwidth,trim=300 0 400 200,clip]{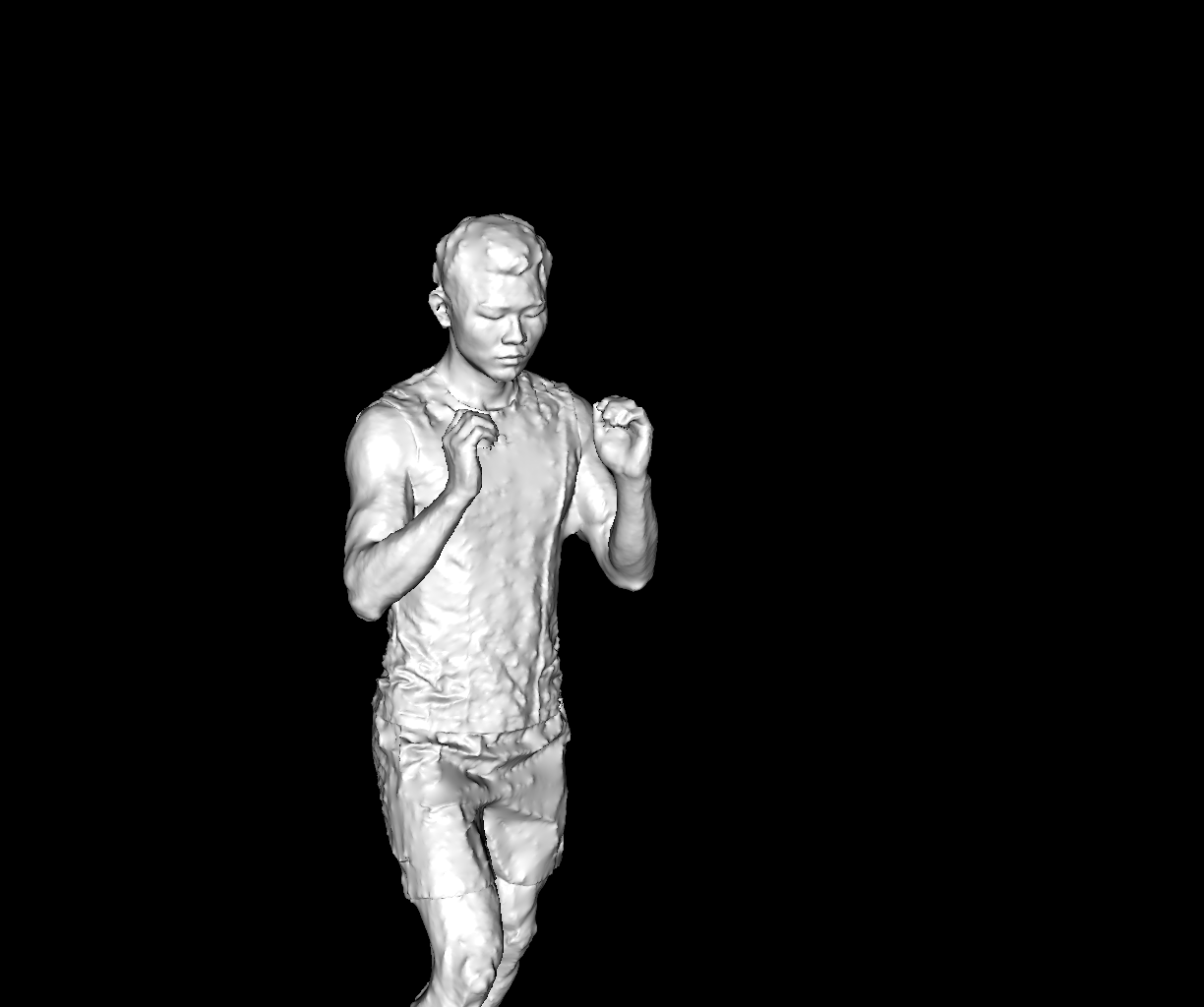} \\
    \includegraphics[width=0.09\textwidth,trim=250 0 400 0,clip]{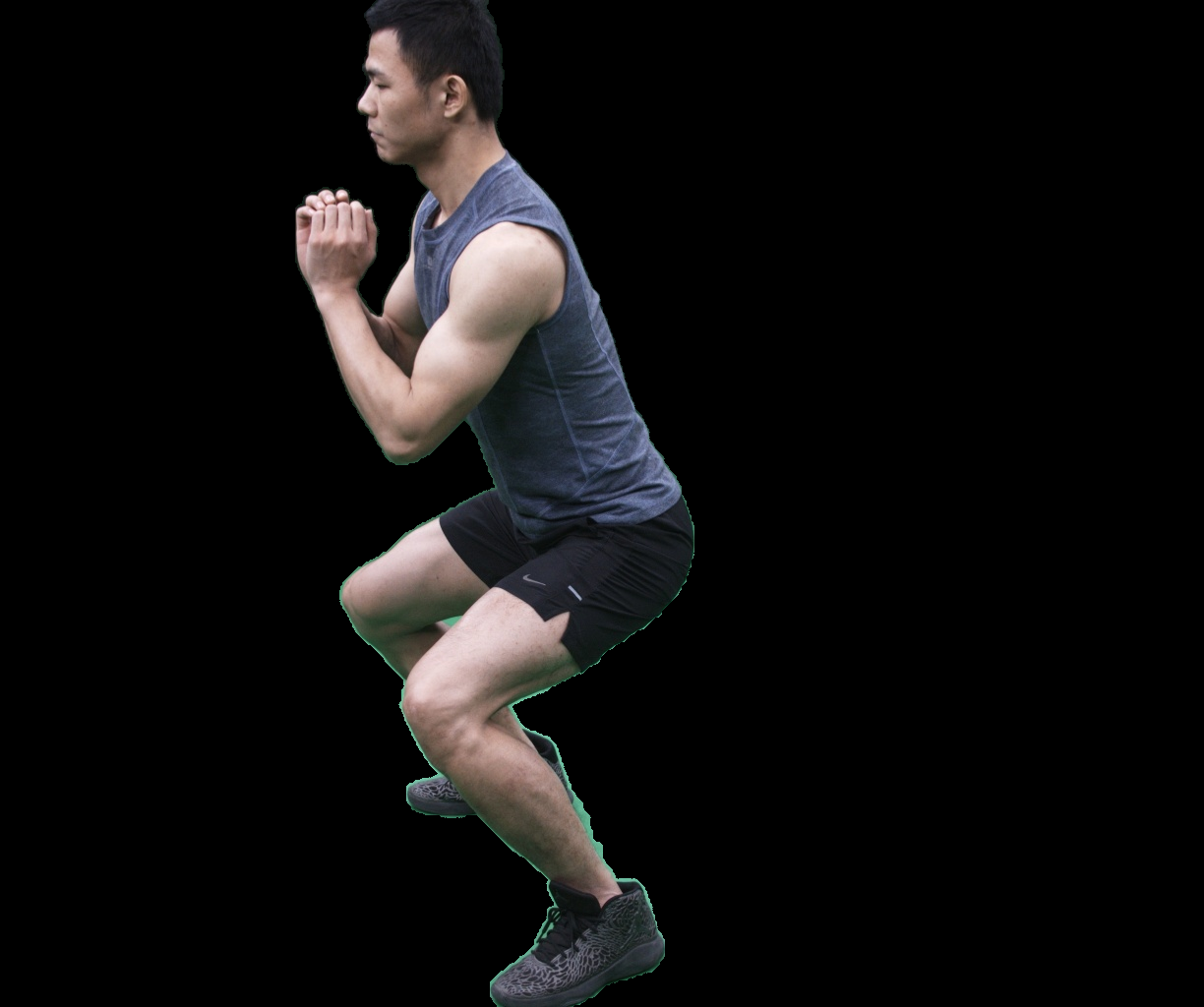} &
    \includegraphics[width=0.09\textwidth,trim=250 0 400 0,clip]{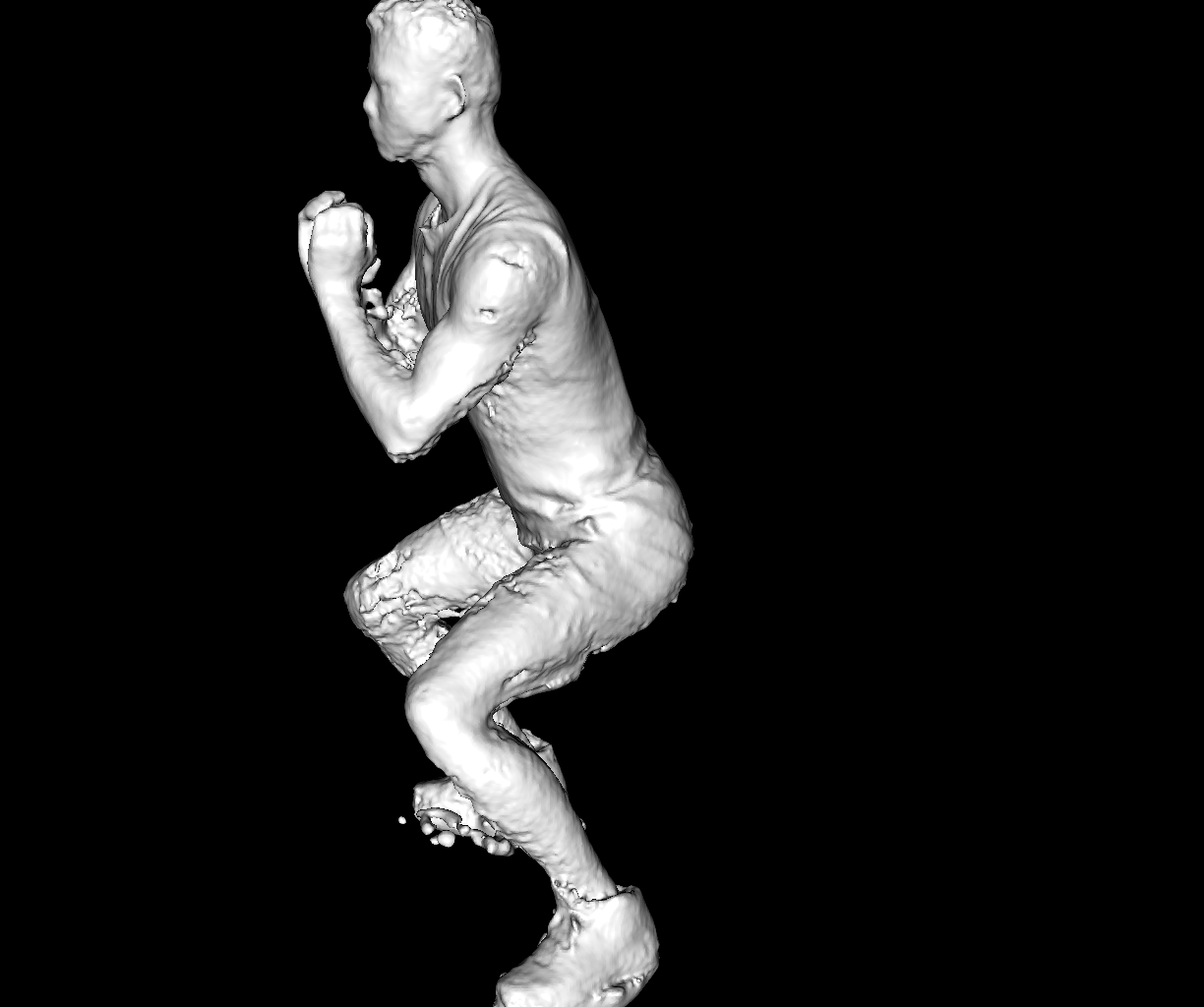} &
    \includegraphics[width=0.09\textwidth,trim=250 0 400 0,clip]{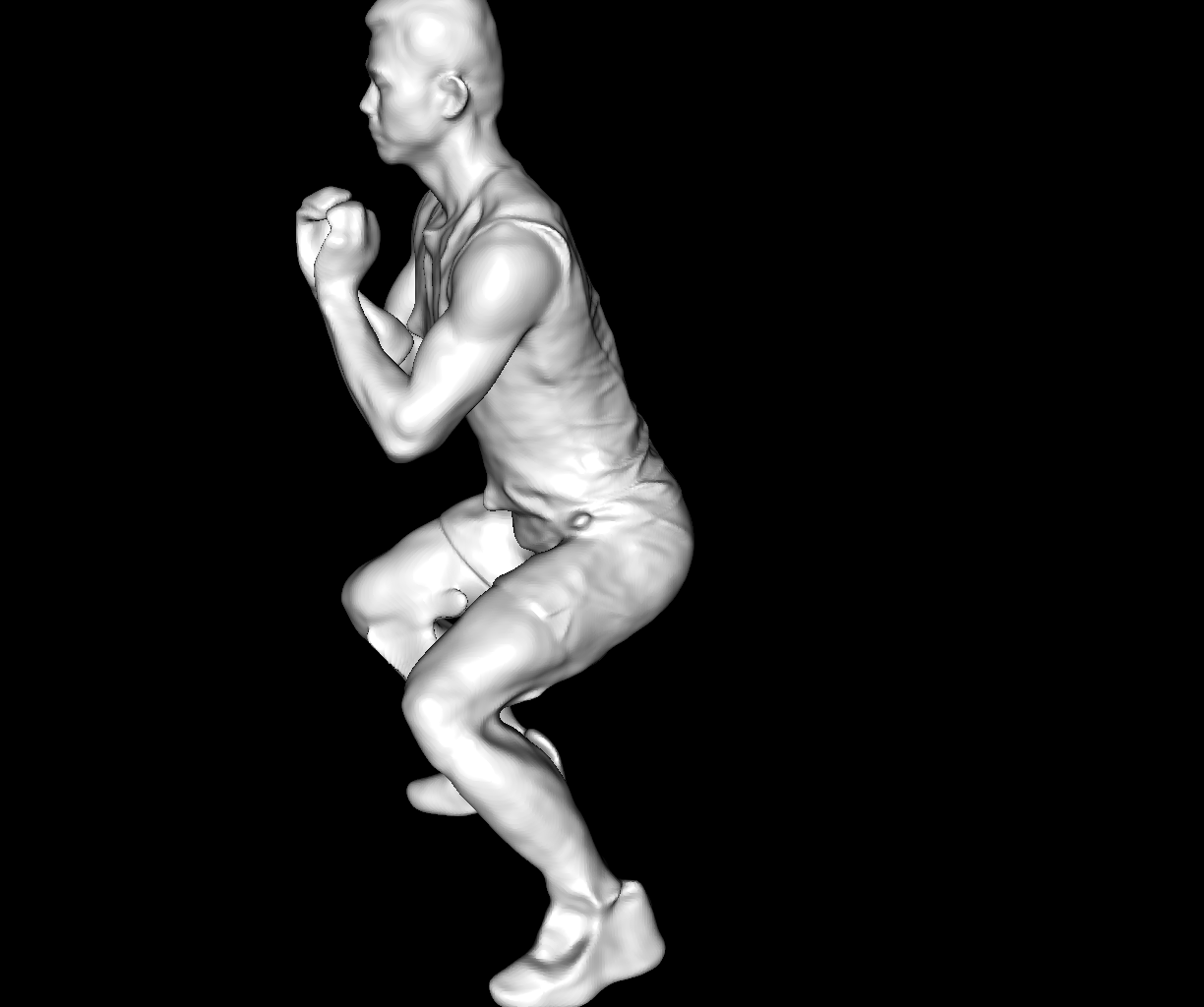} &
    \includegraphics[width=0.09\textwidth,trim=250 0 400 0,clip]{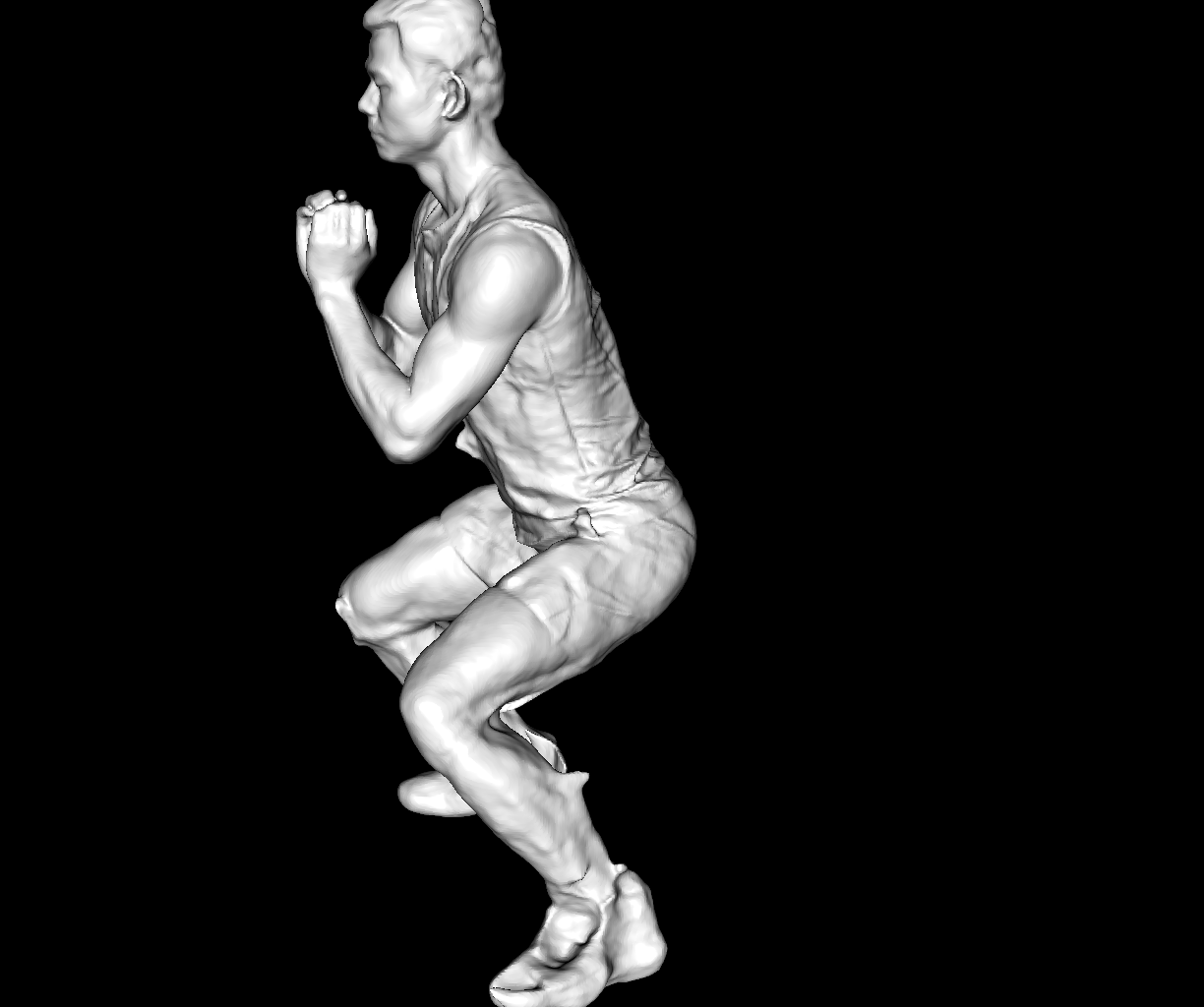} &
    \includegraphics[width=0.09\textwidth,trim=250 0 400 0,clip]{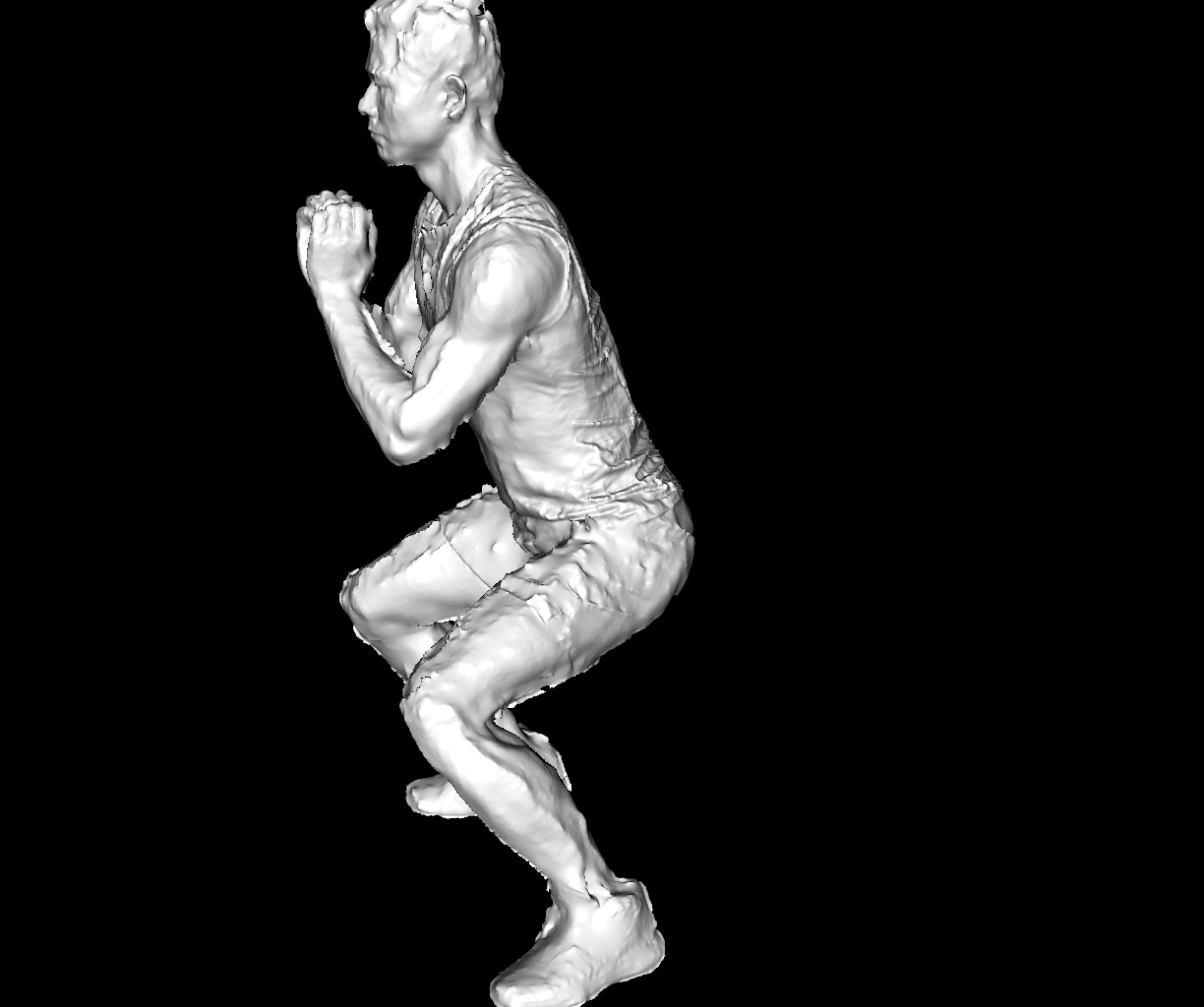} \\
    GT View & Metashape & IDR & NeuS & Ours
    \end{tabular}
    \caption{\textbf{Qualitative Comparison for Human Reconstruciton on NHR dataset~\cite{wu2020multi}.} We compare our method with Metashape~\cite{Metashape}, IDR~\cite{DBLP:conf/nips/idr20}, and NeuS~\cite{DBLP:conf/nips/neus21}.
    }
    \label{fig:nhr_ret}
    \vspace{-0.2in}
\end{figure}

\begin{figure*}
    \centering
    \begin{tabular}{@{\hskip2pt}c@{\hskip2pt}@{\hskip2pt}c@{\hskip2pt}@{\hskip2pt}c@{\hskip2pt}@{\hskip2pt}c@{\hskip2pt}@{\hskip2pt}c@{\hskip2pt}@{\hskip2pt}c@{\hskip2pt}@{\hskip2pt}c@{\hskip2pt}}
    \includegraphics[width=0.14\textwidth,trim=250 250 150 150,clip]{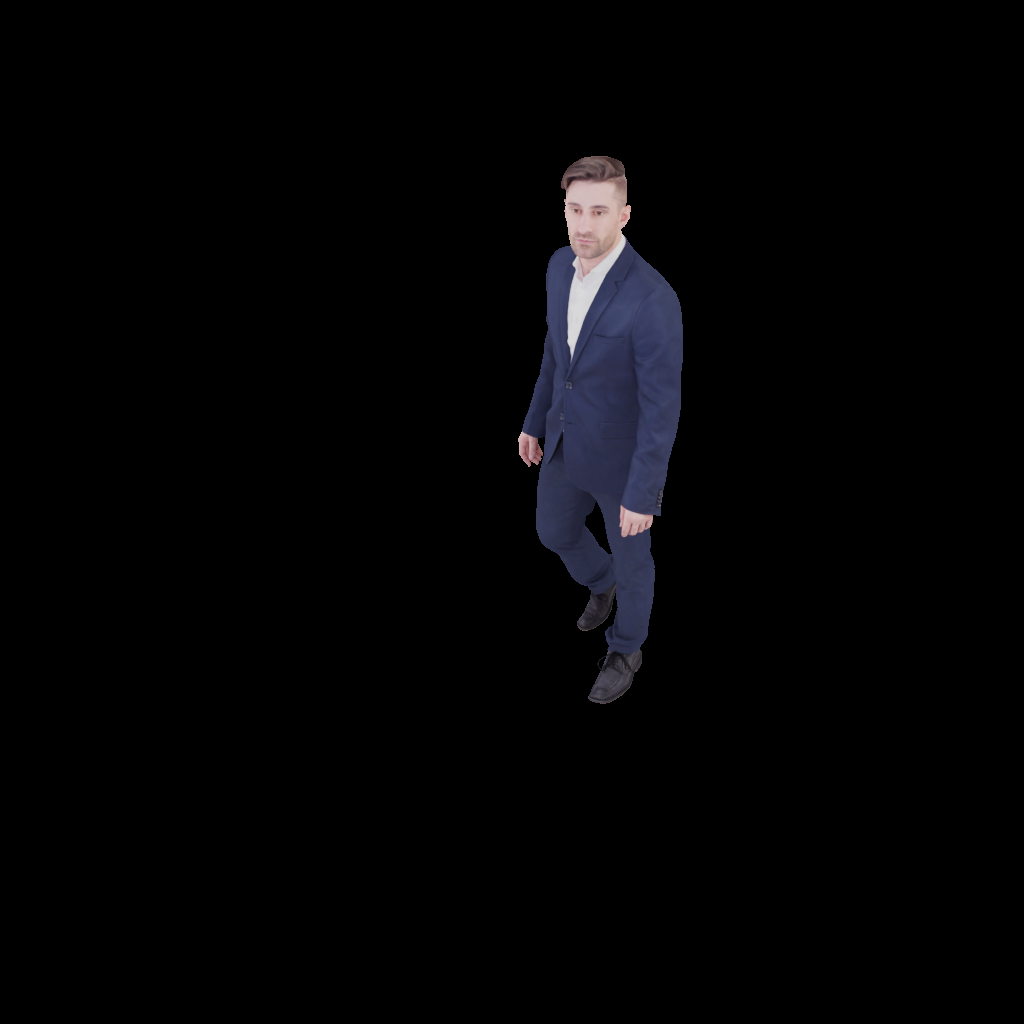} &
    \includegraphics[width=0.14\textwidth,trim=250 250 150 150,clip]{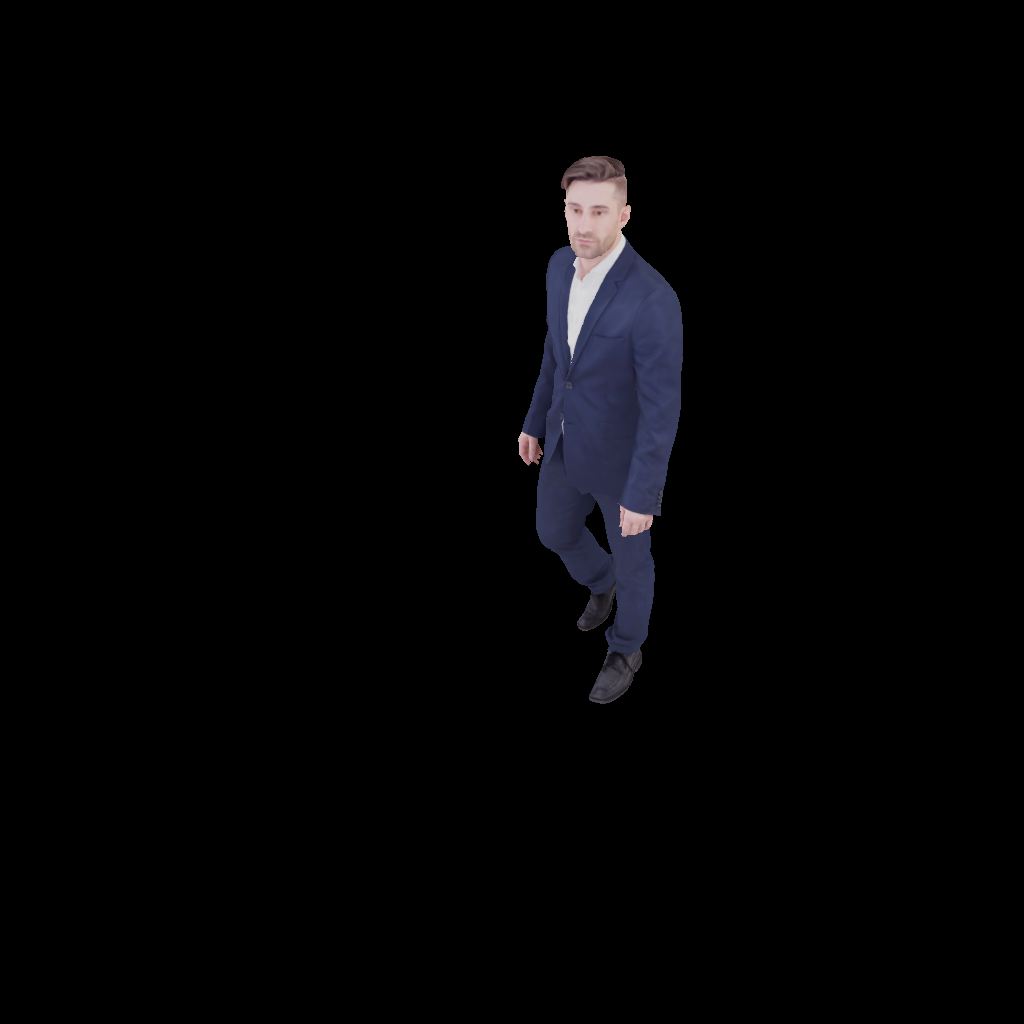} &
    \includegraphics[width=0.14\textwidth,trim=250 250 150 150,clip]{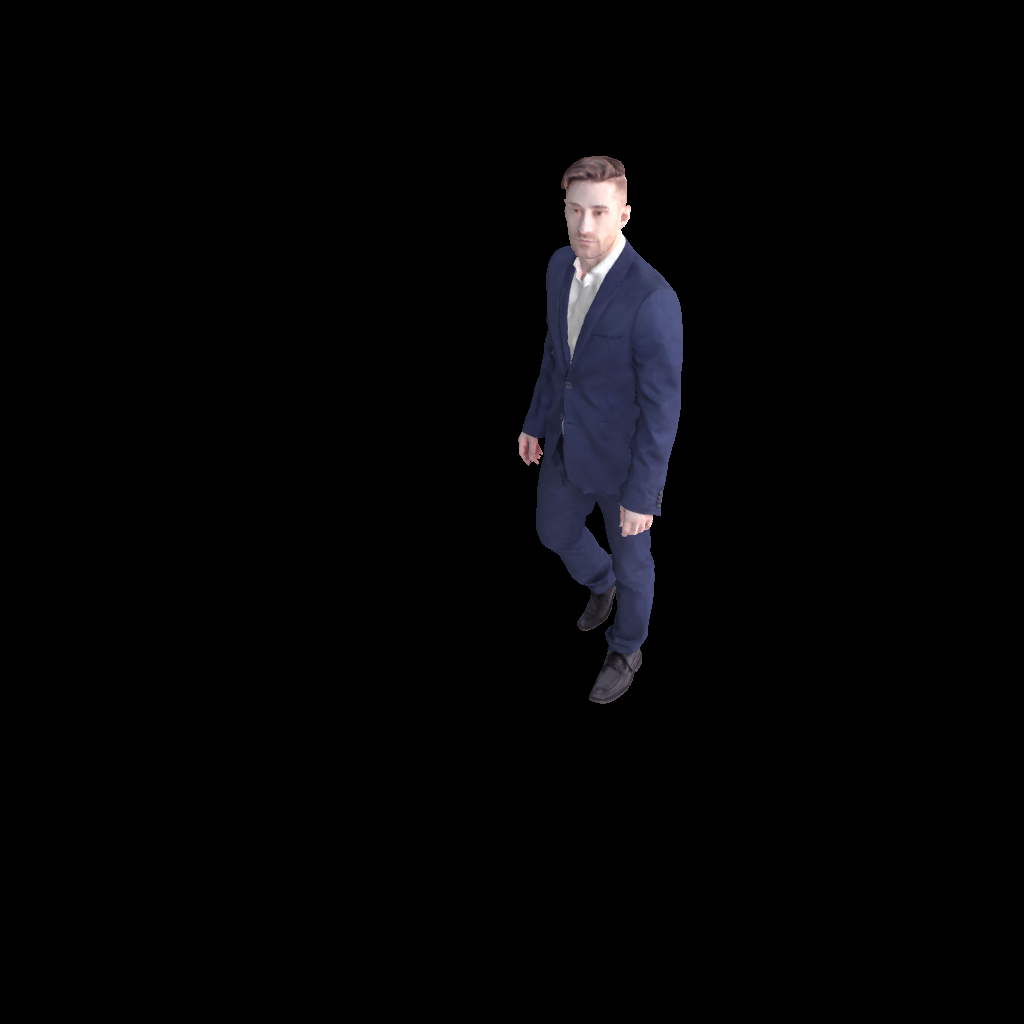} &
    \includegraphics[width=0.14\textwidth,trim=250 250 150 150,clip]{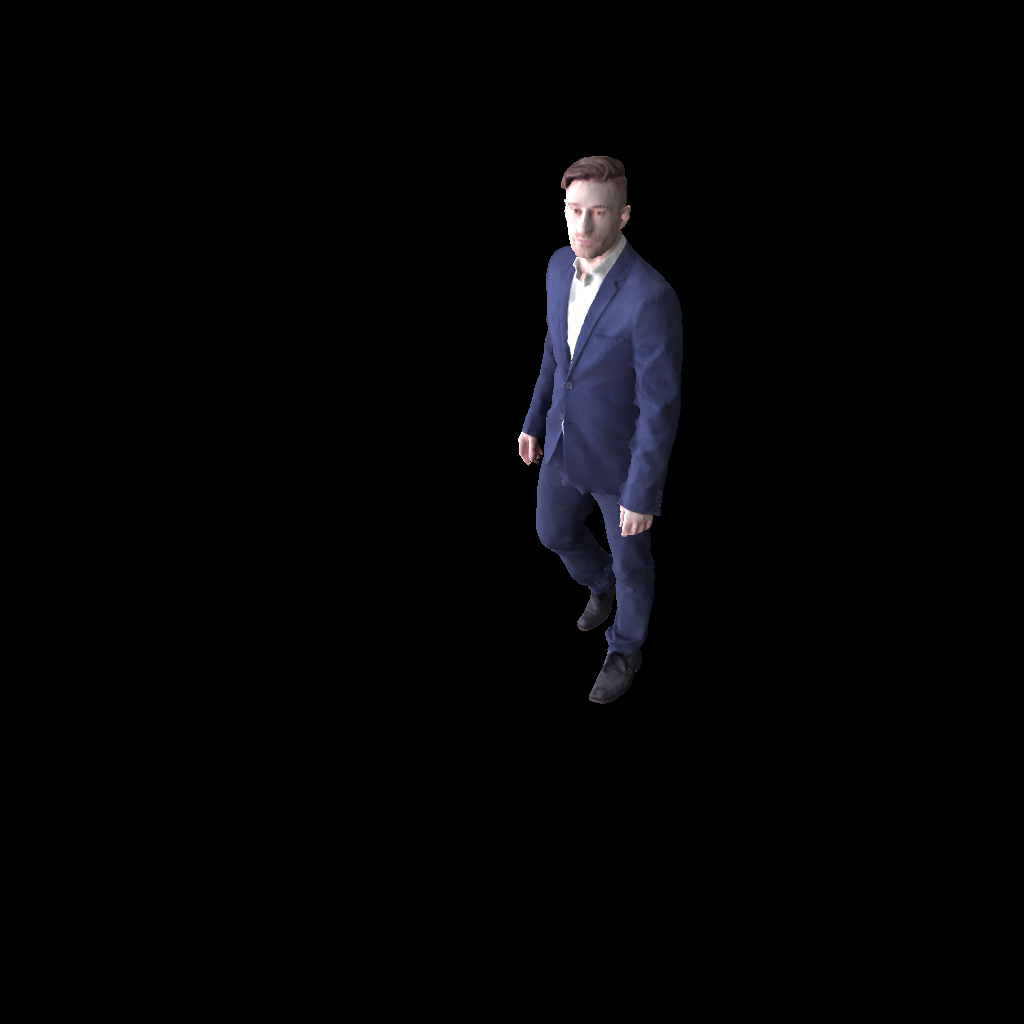} &
    \includegraphics[width=0.14\textwidth,trim=200 200 200 200,clip]{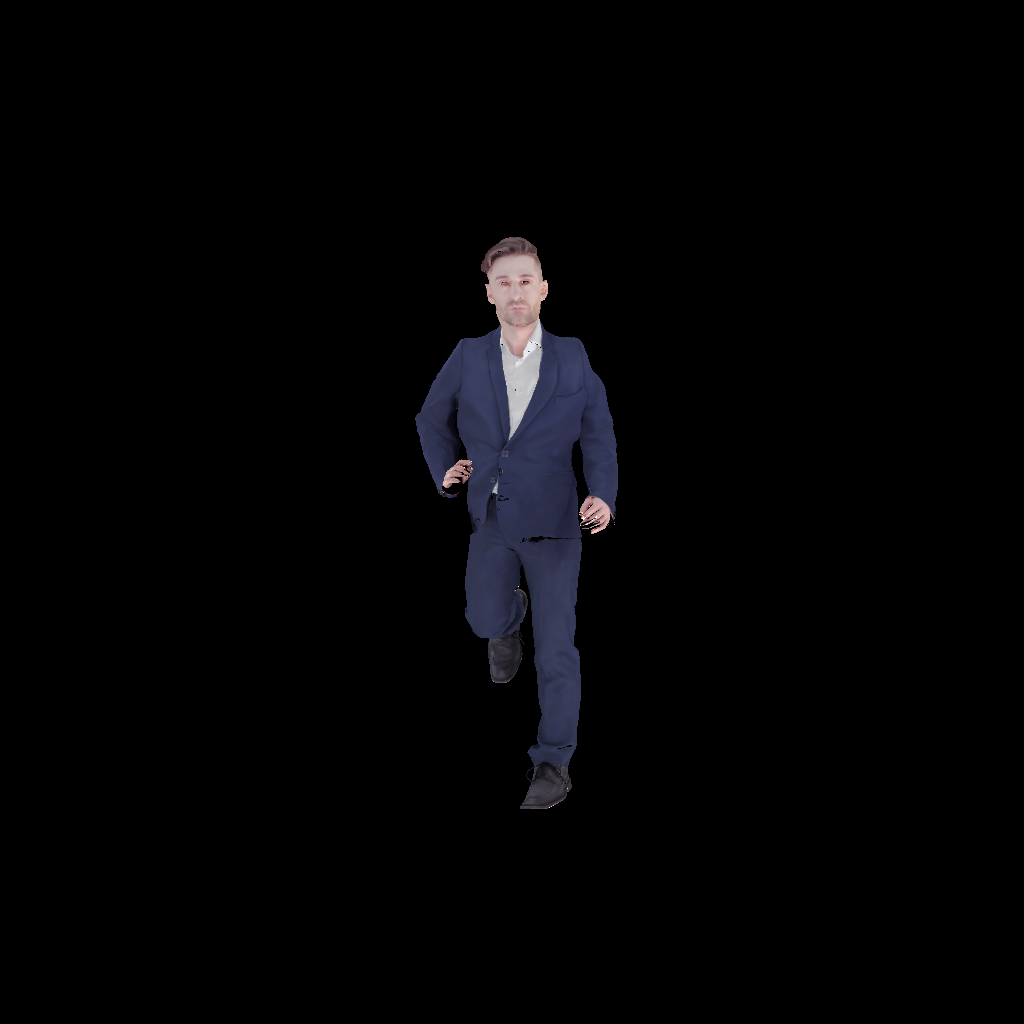} &
    \includegraphics[width=0.14\textwidth,trim=200 200 200 200,clip]{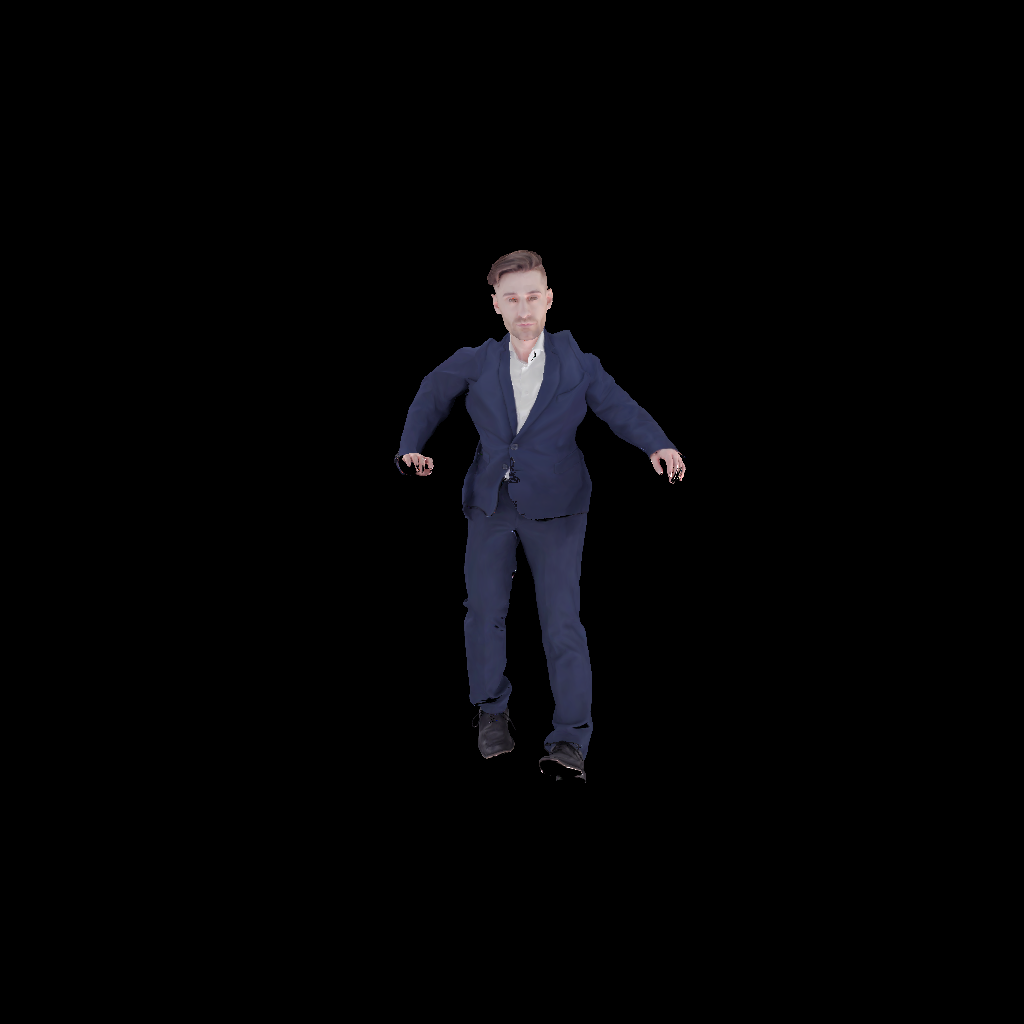} &
    \includegraphics[width=0.14\textwidth,trim=200 200 200 200,clip]{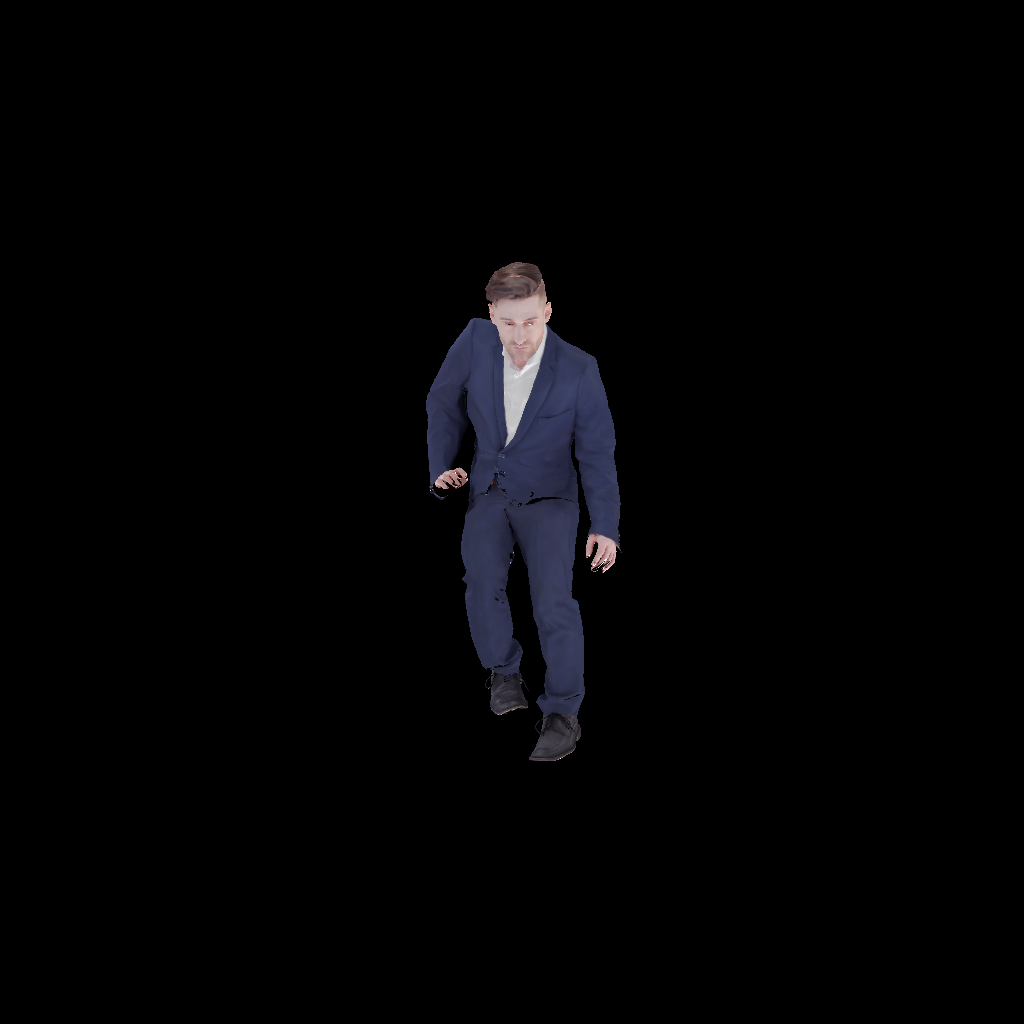} \\
    \includegraphics[width=0.14\textwidth,trim=100 0 100 0,clip]{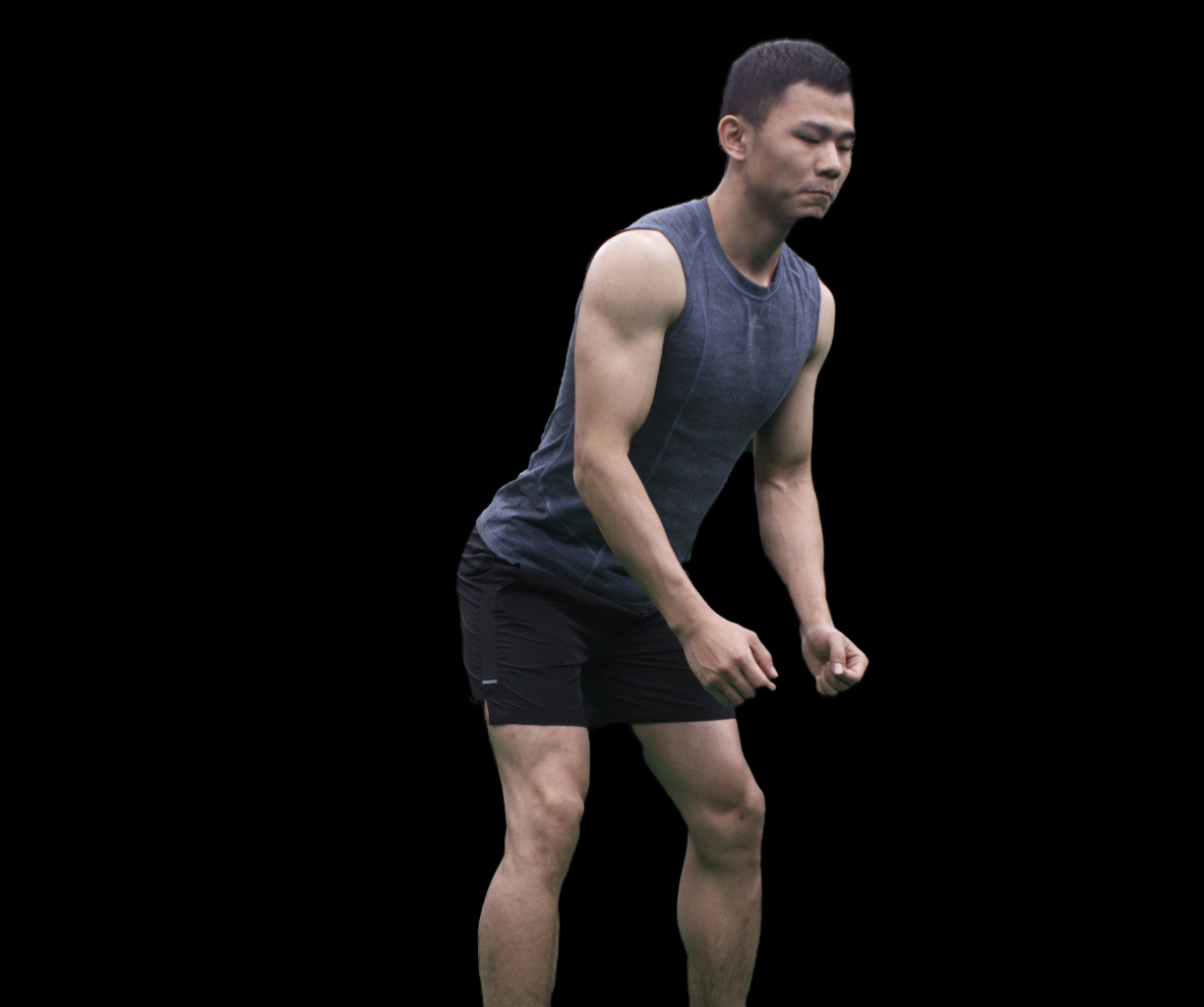} &
    \includegraphics[width=0.14\textwidth,trim=100 0 100 0,clip]{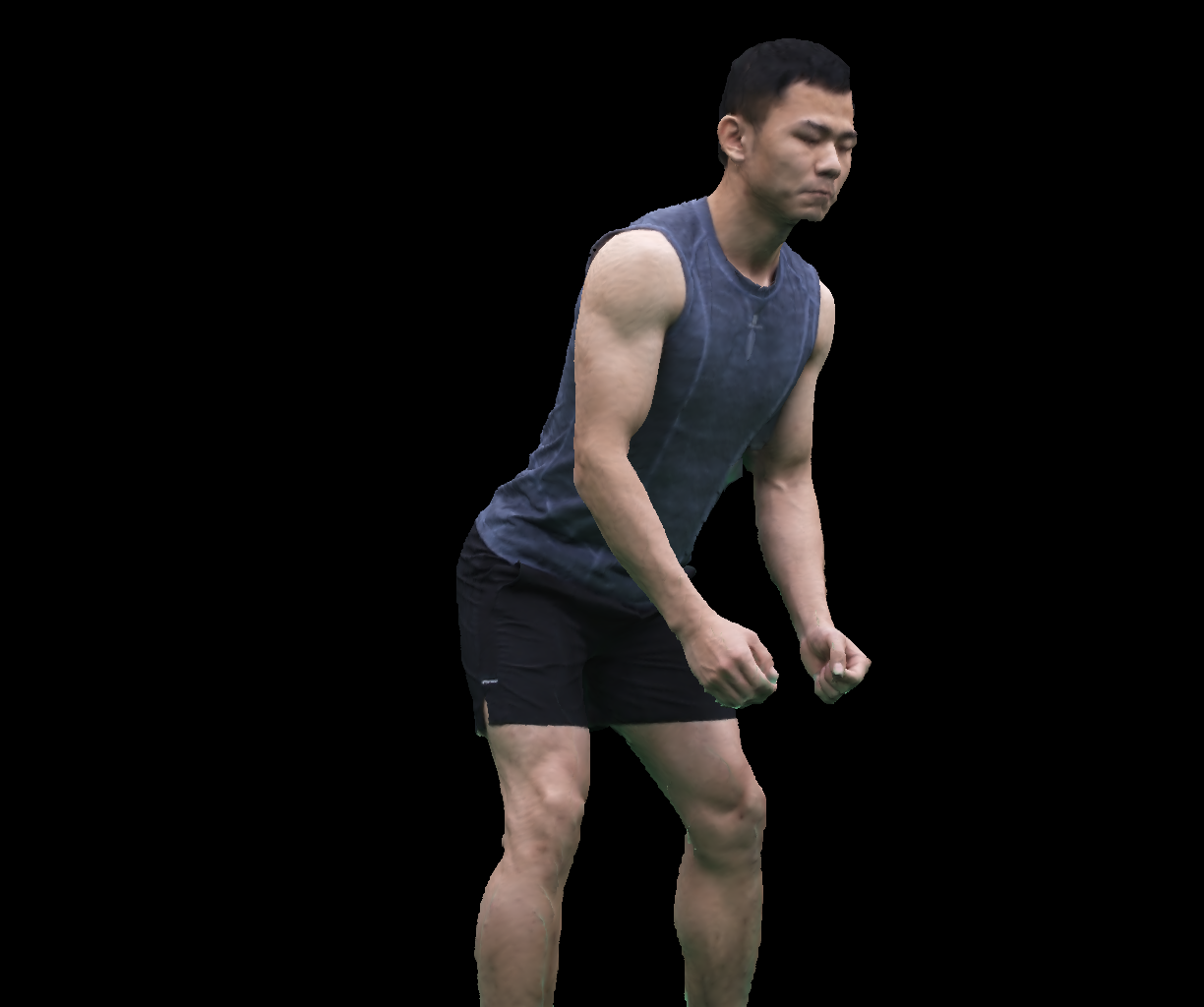} &
    \includegraphics[width=0.14\textwidth,trim=100 0 100 0,clip]{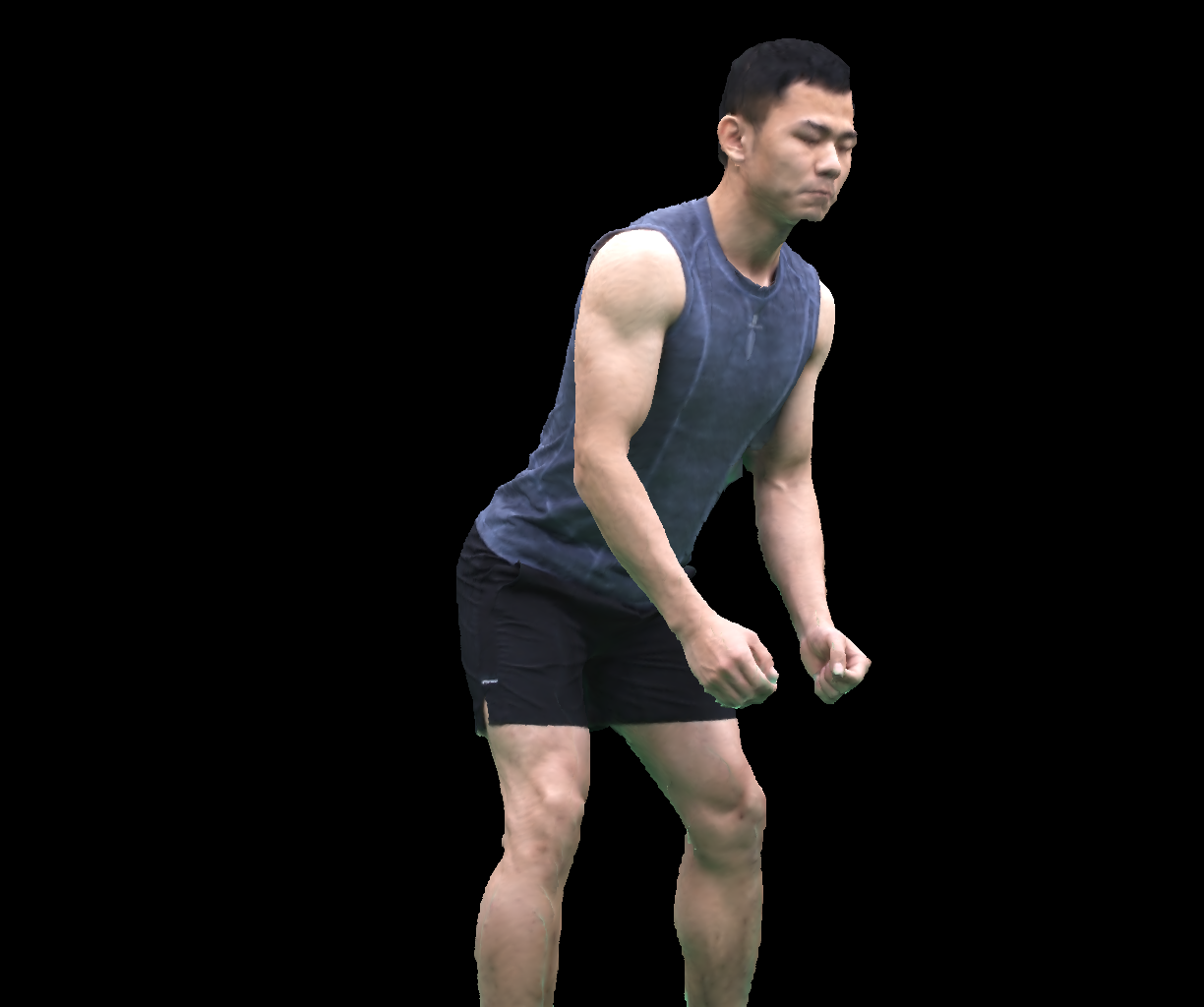} &
    \includegraphics[width=0.14\textwidth,trim=100 0 100 0,clip]{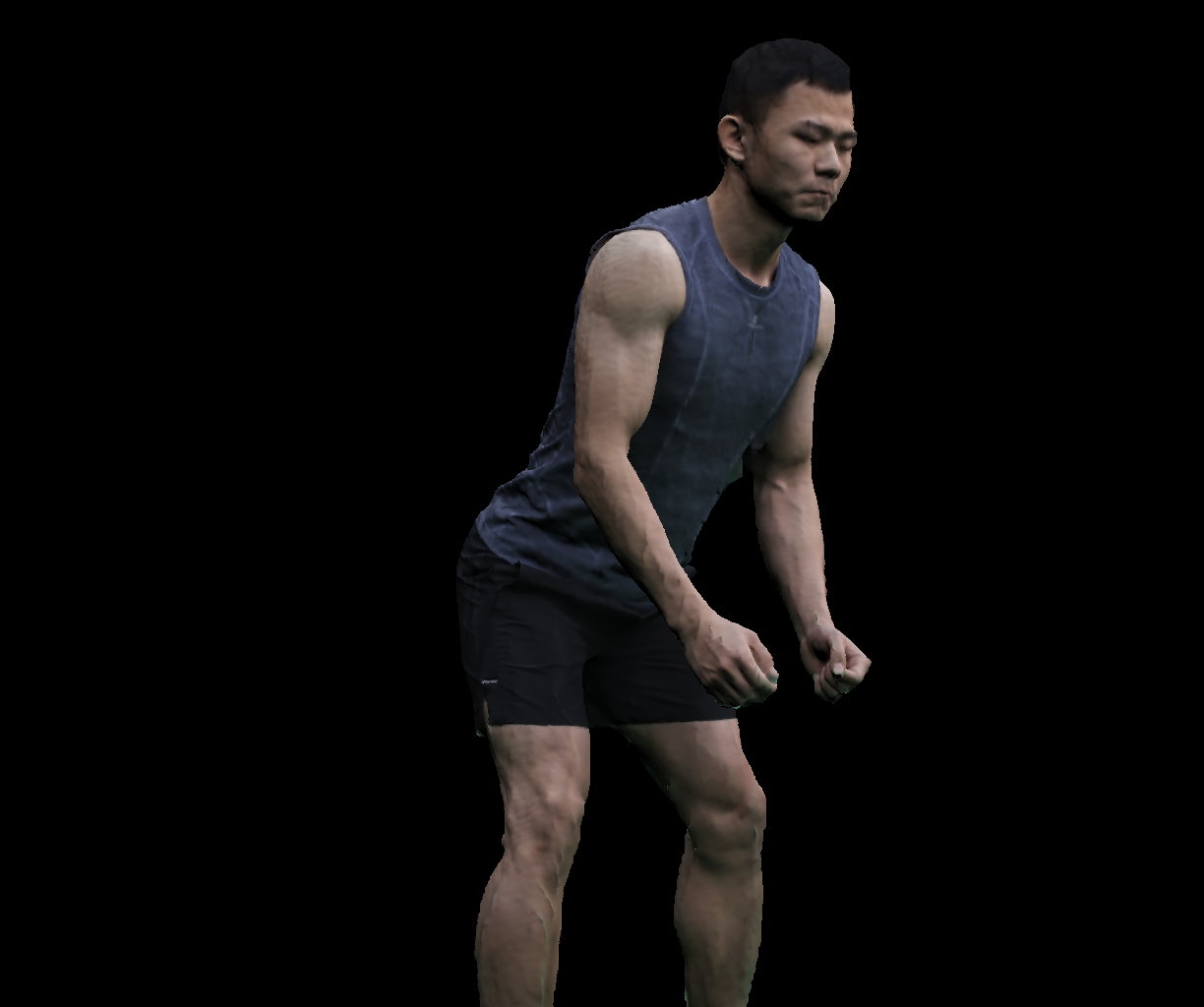} &
    \includegraphics[width=0.14\textwidth,trim=200 130 200 270,clip]{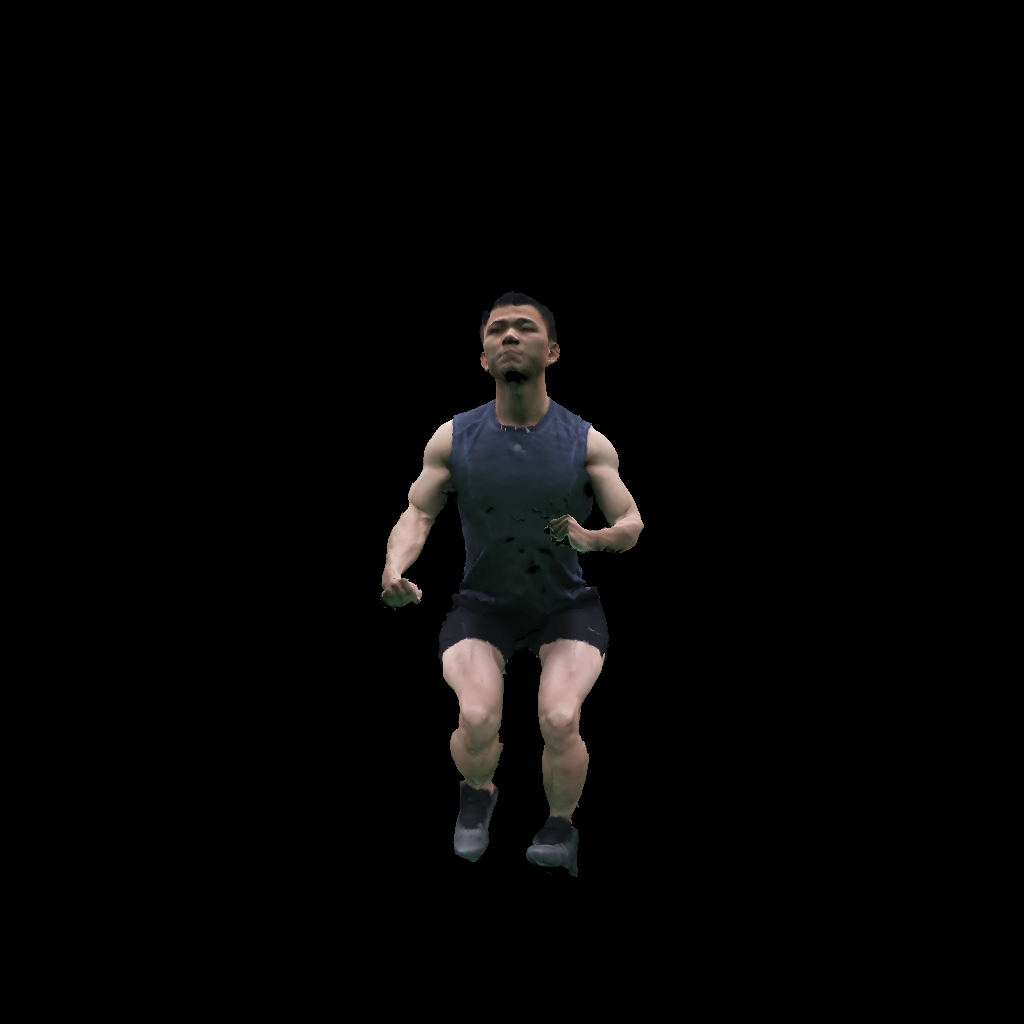} &
    \includegraphics[width=0.14\textwidth,trim=200 130 200 270,clip]{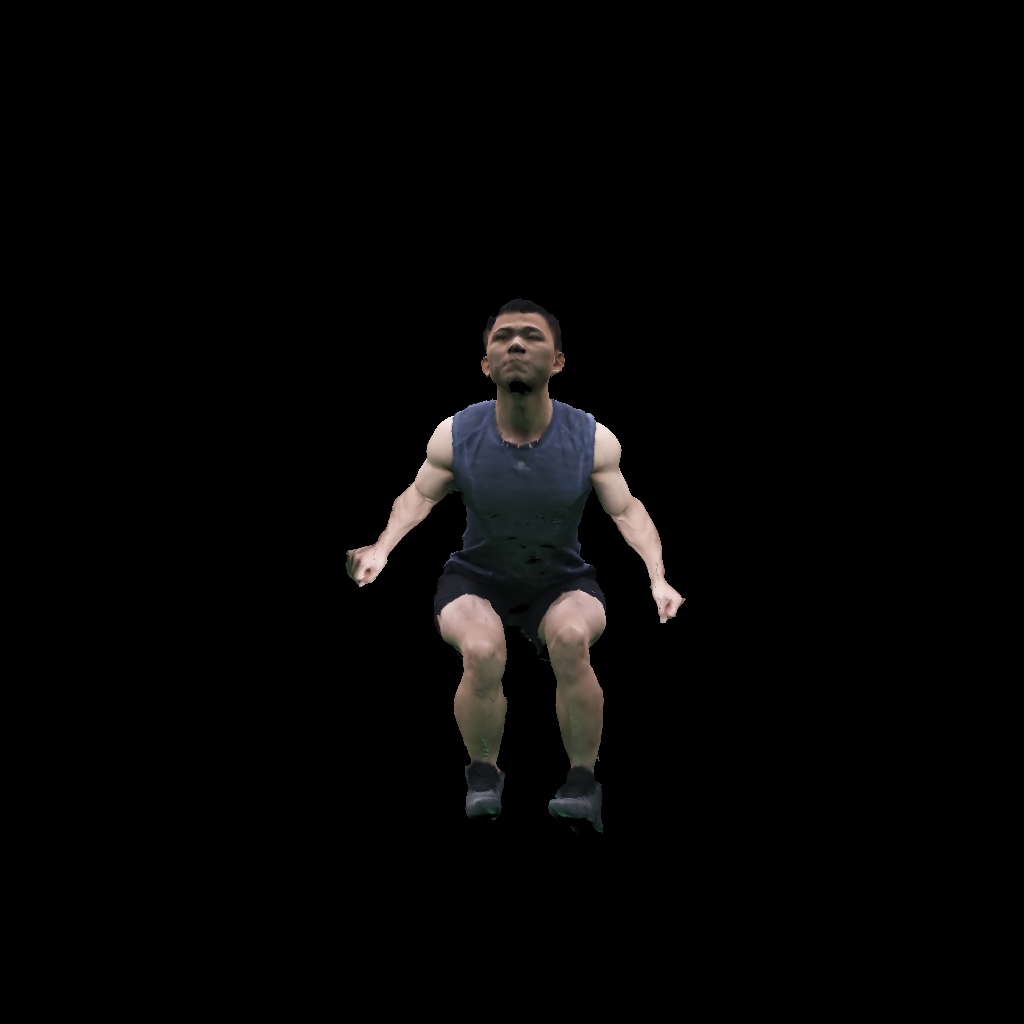} &
    \includegraphics[width=0.14\textwidth,trim=200 130 200 270,clip]{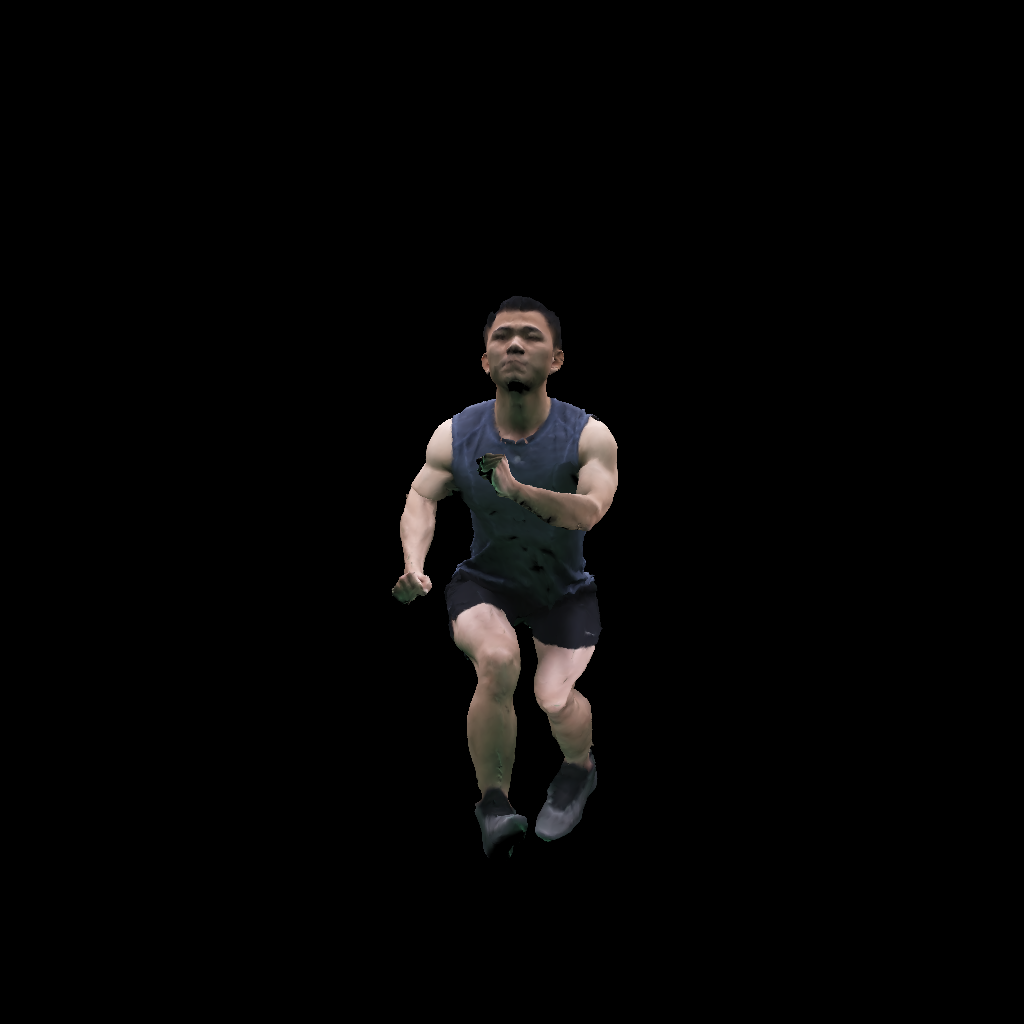} \\
    GT View & Our render & Relighting & Relighting & Reposing & Reposing & Respoing
    \end{tabular}
    \caption{\textbf{Relighting and Reposing Results on Our Reconstructed Meshes.} We show qualitative results of relighting and reposing. We register the SMPLX model on the reconstructed meshes. Hand and face parameters are estimated by ~\cite{DBLP:conf/cvpr/SMPLX19}.}
    \label{fig:repose}
\end{figure*}

\subsection{Results on Multi-view Videos}
We also conduct experiments on real world captured NHR dataset~\cite{wu2020multi}. NHR dataset is collected by a multi-camera dome system with up to 80 cameras arranged on a cylinder. All cameras are synchronized and capture at 25 frames per second. We conduct experiments on three sequences. We use 24 images with a resolution of $1224\times1024$ as input. Since there is no ground truth mesh for each frame, we only evaluate the quality of the image synthesis with metric PSNR. We qualitatively compare reconstruction results with IDR~\cite{DBLP:conf/nips/idr20}, NeuS~\cite{DBLP:conf/nips/neus21} and the point cloud reconstructed using Metashape~\cite{Metashape} provided by the dataset. The evaluation metrics are the same as in the previous subsection. for the quantitative and qualitative evaluation of general real-world objects on DTU~\cite{DBLP:conf/cvpr/DTU14}, please refer to the supp. mat.

Table ~\ref{tab:nhr_psnr} shows the image synthesis results. It can be seen that we can obtain the photo-realistic images using simple SH illumination and albedos, which greatly accelerate rendering. Fig.~\ref{fig:nhr_ret} shows the reconstruction results. Our proposed method can produce detailed mesh. Similar to the previous subsection, IDR and NeuS tend to produce smooth meshes, hand detail can not be recovered well. Moreover, the ambiguity of geometry and appearance can not be handled by rendering loss alone, which results in wrong reconstruction results of shoes.

\begin{table}
	\centering
	\begin{tabular}{c|ccc}
		\toprule
		 & Sport 1 &  Sport 2 & Sport 3 \\ \hline
		IDR~\cite{DBLP:conf/nips/idr20}  &  22.58  & 21.61   & 21.68    \\
		NeuS~\cite{DBLP:conf/iccv/animatenerf21} & 25.52 & 24.58 &  24.22  \\
	    Ours & \textbf{25.59} & \textbf{24.64} & \textbf{24.60}  \\
		\bottomrule
	\end{tabular}
    \caption{\textbf{Image Synthesis Evaluation on NHR Dataset~\cite{wu2020multi} in PSNR.} We conduct experiments on 3 sequences and our method show superior performance compared to other dynamic approach.}
 \vspace{-0.2in}
 \label{tab:nhr_psnr}
\end{table}

\begin{figure}
    \centering
    \begin{tabular}{@{\hskip2pt}c@{\hskip2pt}@{\hskip2pt}c@{\hskip2pt}@{\hskip2pt}c@{\hskip2pt}@{\hskip2pt}c@{\hskip2pt}@{\hskip2pt}c@{\hskip2pt}}
    \includegraphics[width=0.09\textwidth,trim=450 250 250 120,clip]{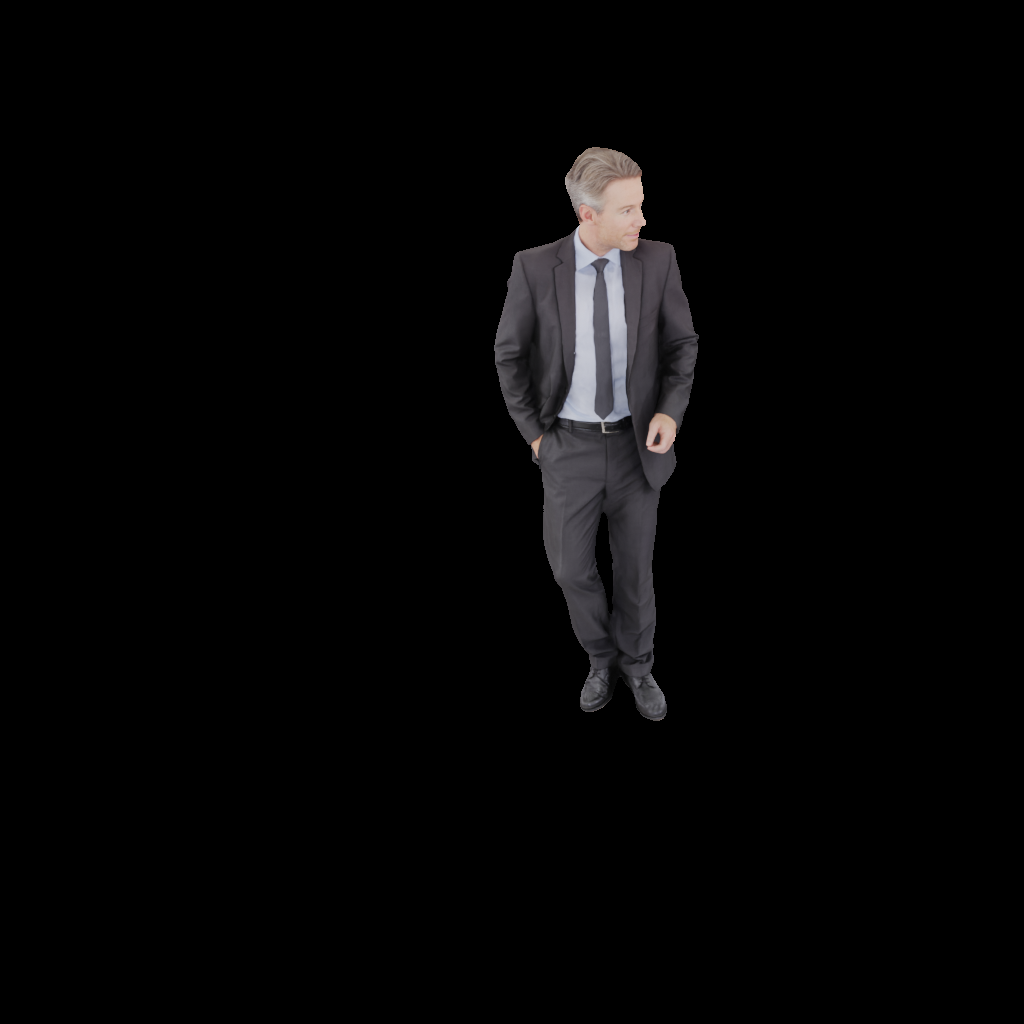} &
    \includegraphics[width=0.09\textwidth,trim=450 250 250 120,clip]{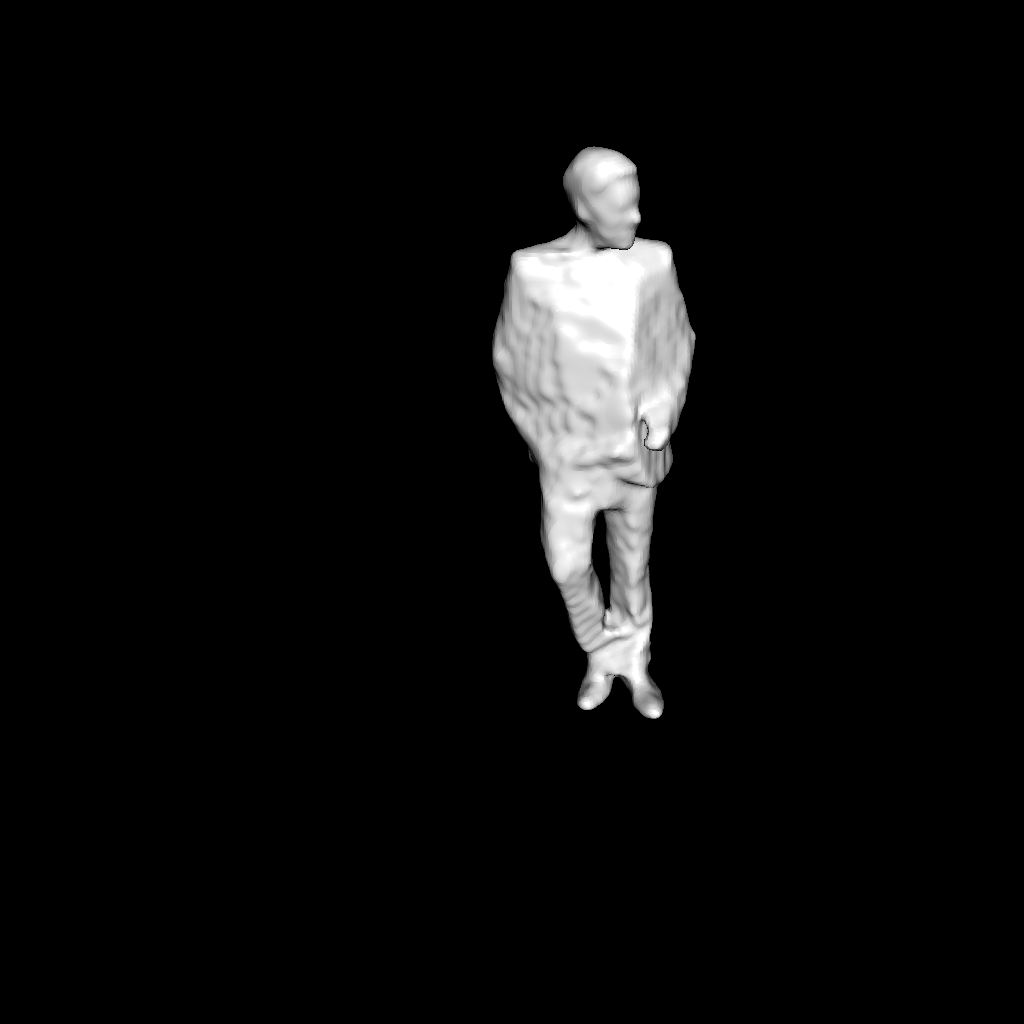} &
    \includegraphics[width=0.09\textwidth,trim=450 250 250 120,clip]{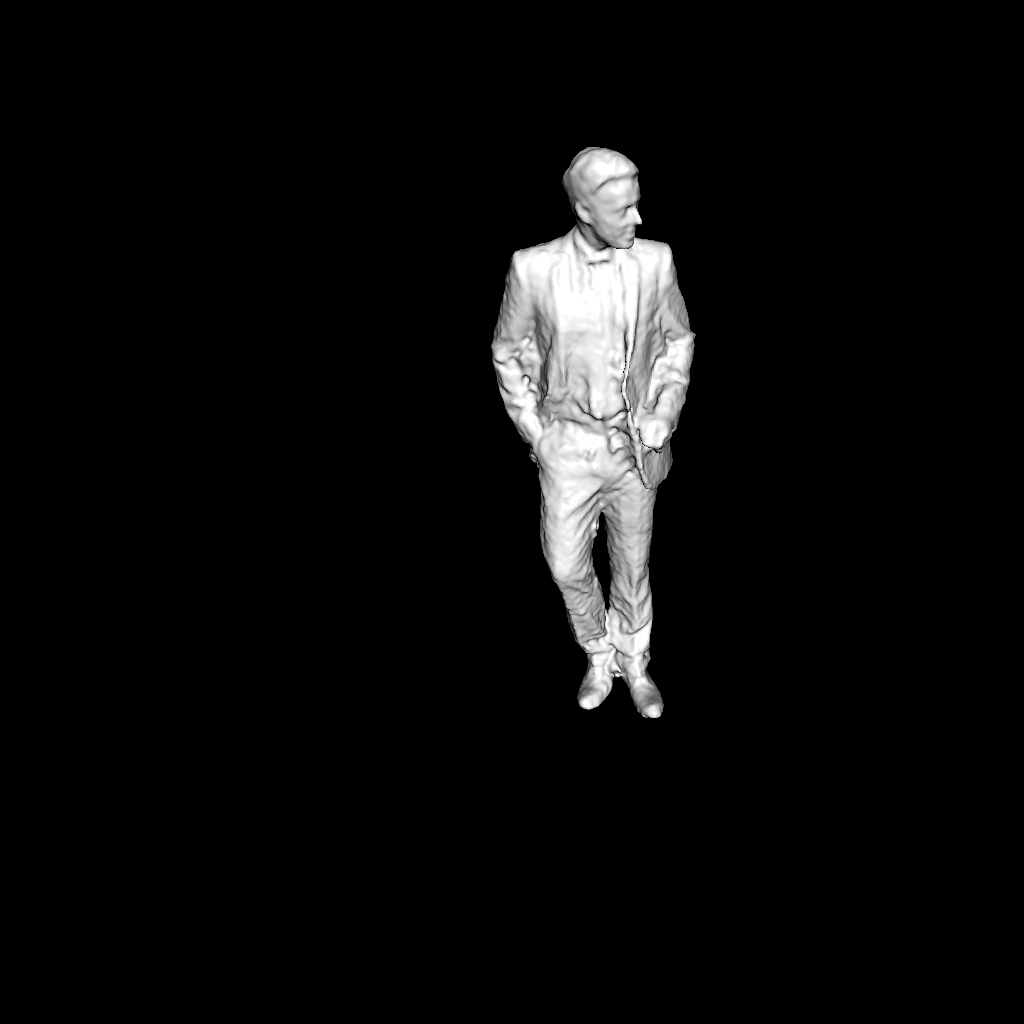} &
    \includegraphics[width=0.09\textwidth,trim=450 250 250 120,clip]{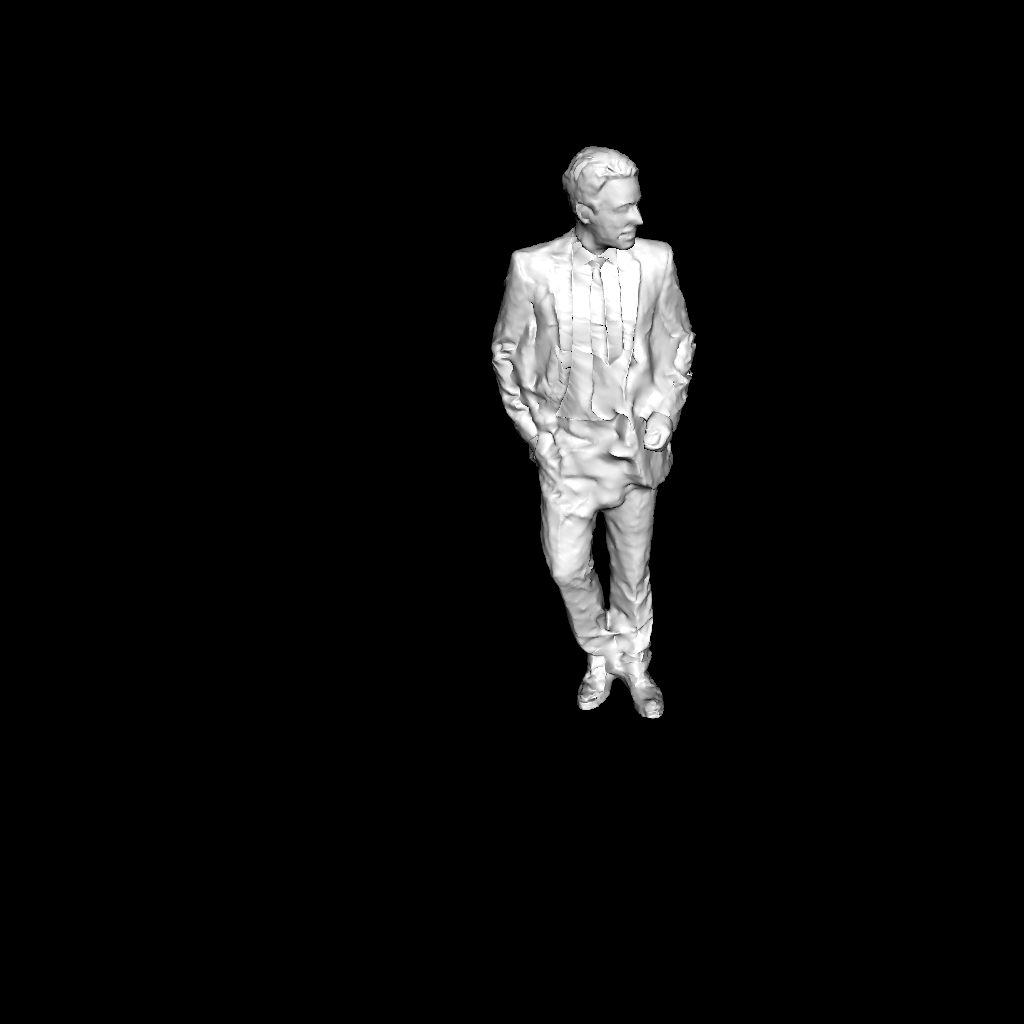} &
    \includegraphics[width=0.09\textwidth,trim=450 250 250 120,clip]{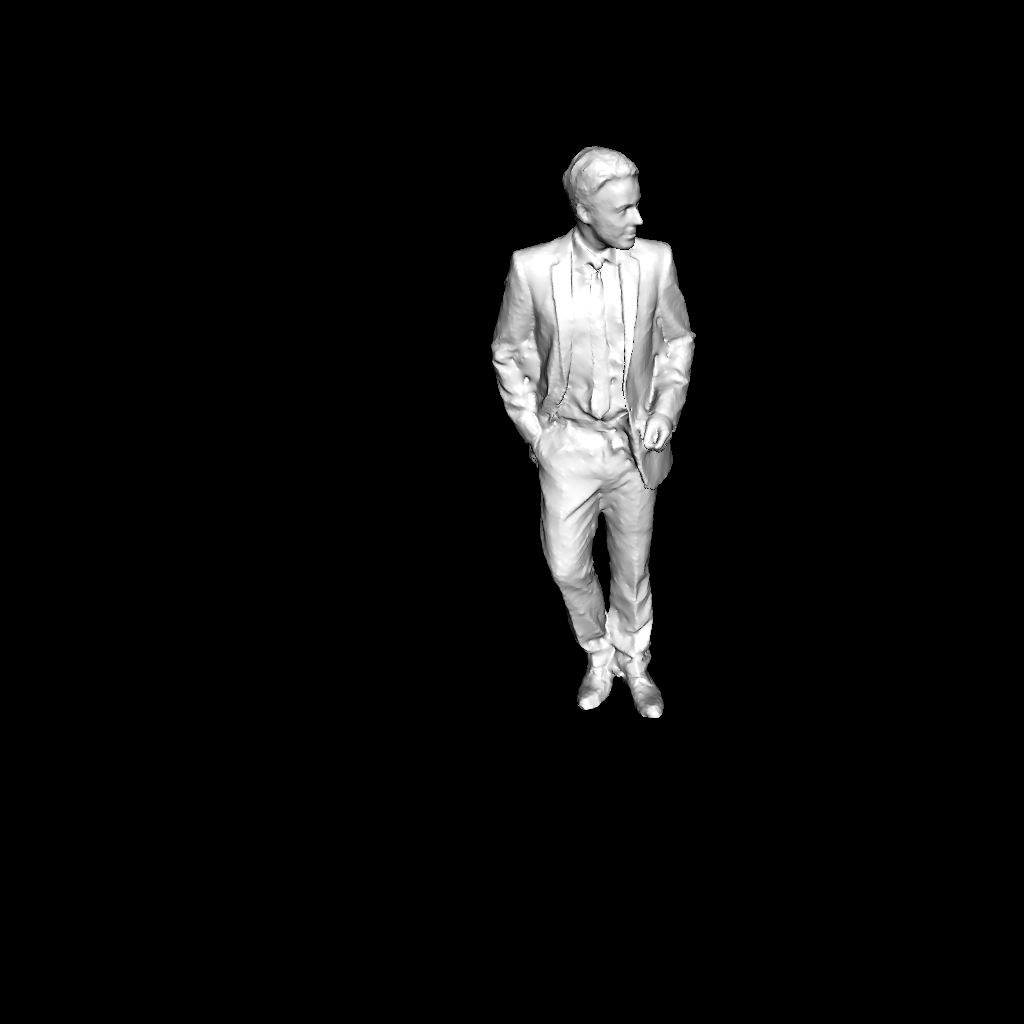} \\
     \includegraphics[width=0.09\textwidth,trim=450 0 250 140,clip]{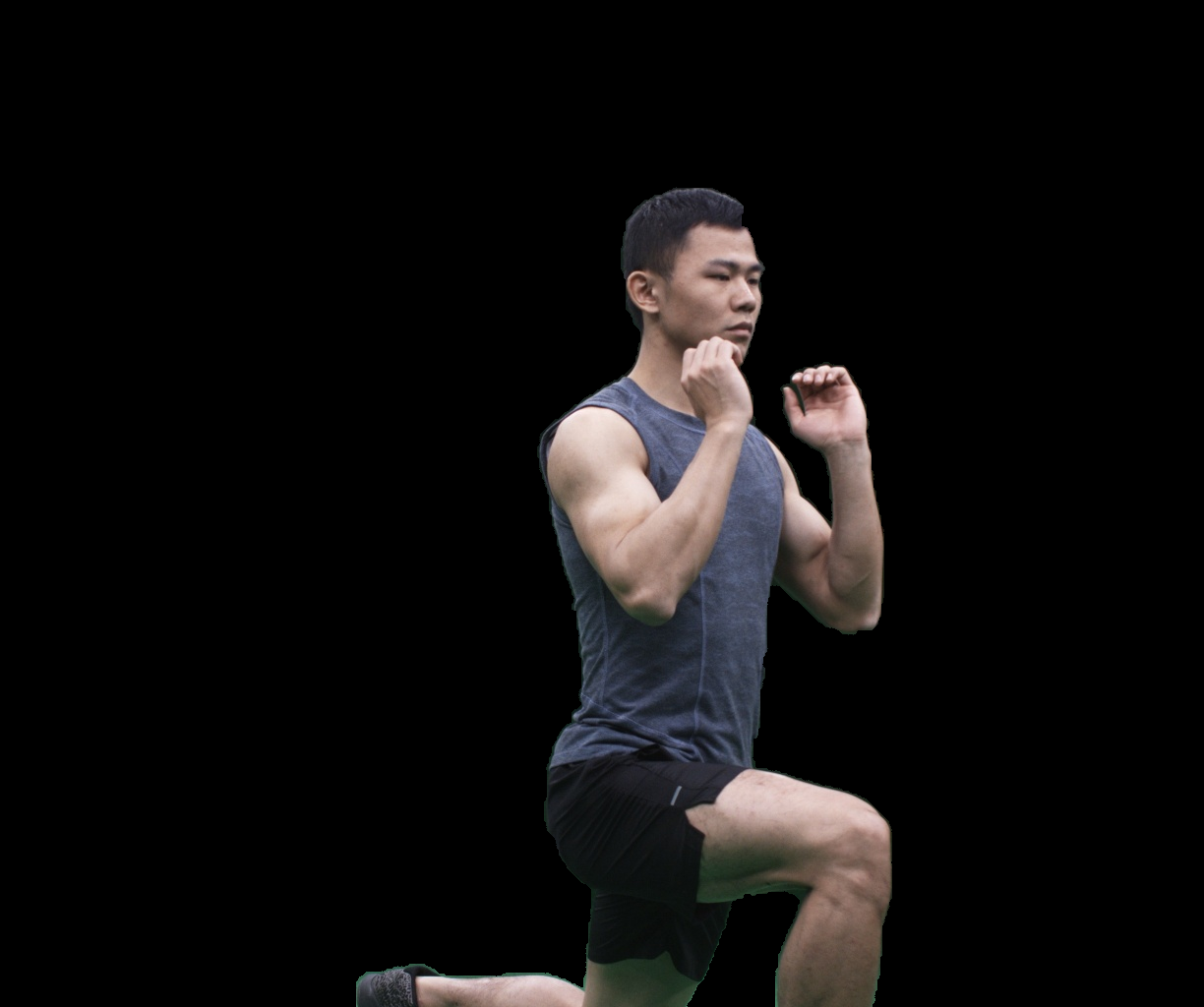} &
     \includegraphics[width=0.09\textwidth,trim=450 0 250 140,clip]{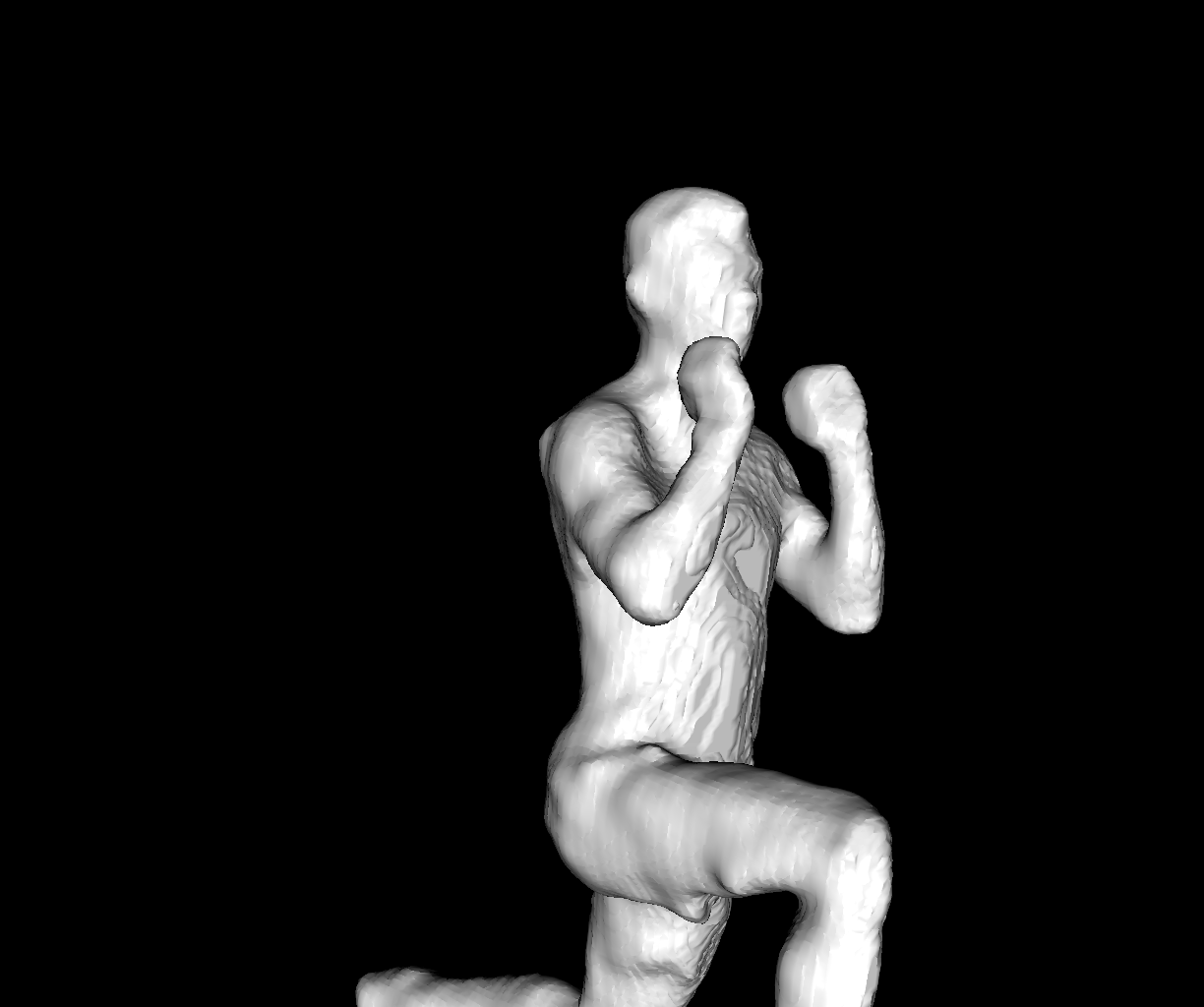} &
     \includegraphics[width=0.09\textwidth,trim=450 0 250 140,clip]{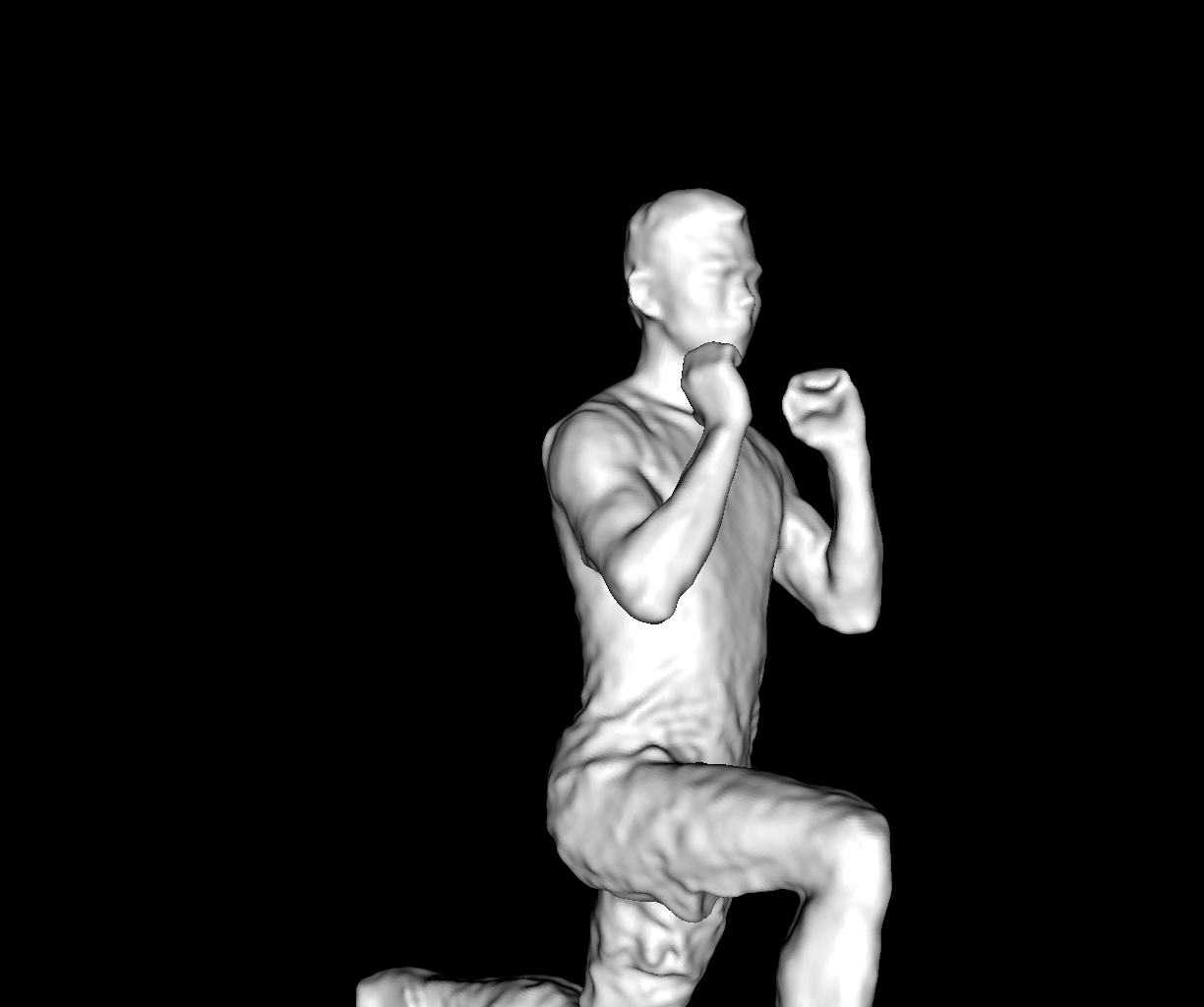} &
     \includegraphics[width=0.09\textwidth,trim=450 0 250 140,clip]{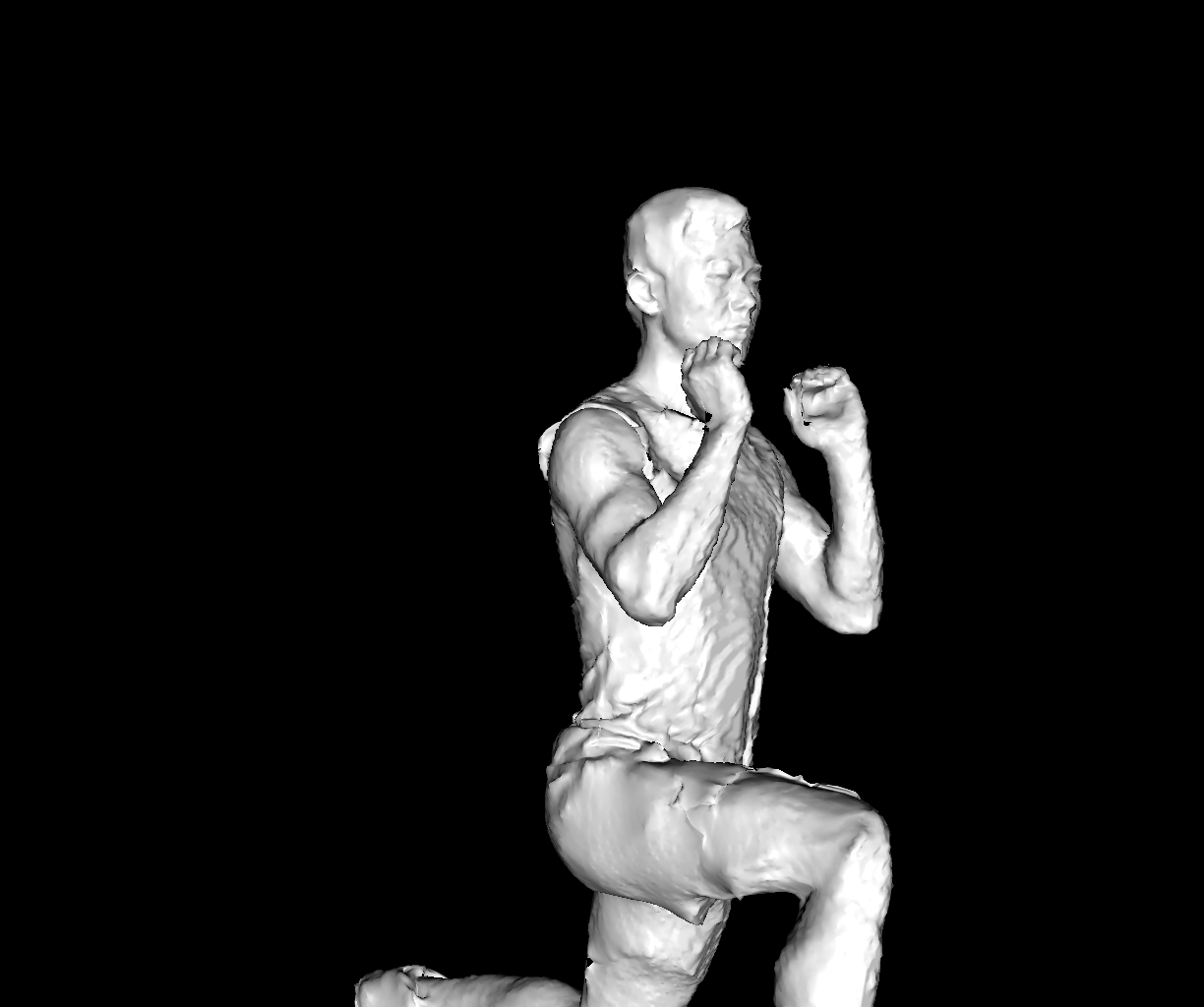} &
     \includegraphics[width=0.09\textwidth,trim=450 0 250 140,clip]{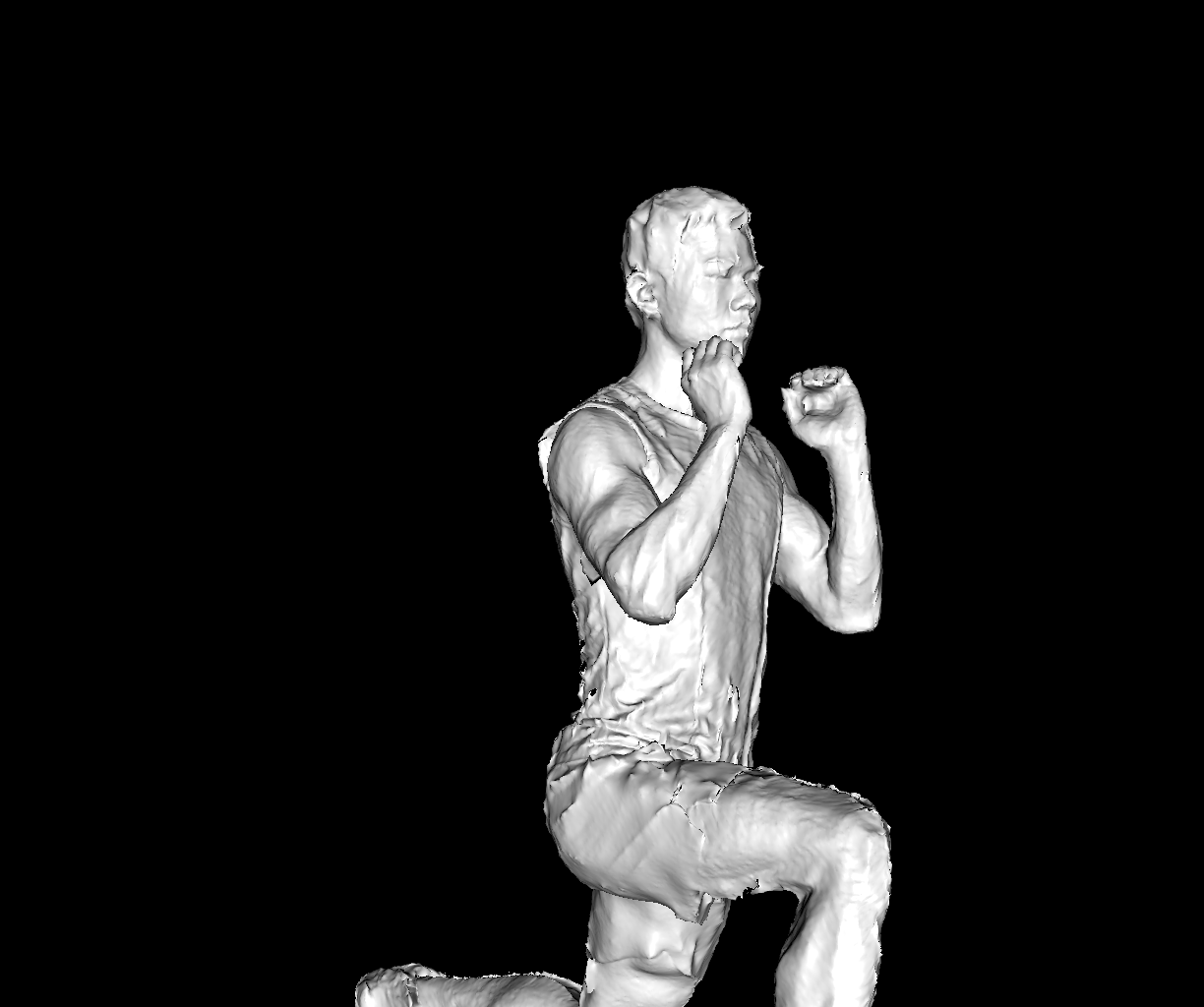} \\
     GT View & Visual Hull & NCC loss & SFS loss & NCC + SFS
    \end{tabular}
    
    \caption{\textbf{Qualitative Ablation Study on the Proposed Losses}. We show the results of visual hull initialization, only photometric consistency results, only shape from shading results and overall results, respectively.}
    \label{fig:ablation}
    \vspace{-0.2in}
\end{figure}

We can create a realistic human avatar by combining our reconstructed mesh with SMPL human model~\cite{DBLP:journals/tog/SMPL15} and deformation transfer~\cite{DBLP:journals/tog/deformtrans04}. We pre-select control points on the SMPL mesh and identify the closest points on our reconstructed mesh. By manipulating pose parameters, we can animate the registered SMPL mesh. Deformation can be efficiently transferred to our mesh due to the corresponding transformation of the control points. Additionally, we can synthesize relighting images by replacing the estimated SH coefficients. The results are shown in Fig.~\ref{fig:repose}, demonstrating our ability to generate realistic images with arbitrary lighting and poses.

\subsection{Ablation Studies}
In this section, we conduct ablation studies on the effect of the proposed NCC loss and SFS loss. We adopt visual hull initialization instead of deforming from the sphere for fast convergence. Table~\ref{tab:abl} shows the quantitative results. As can be noticed, our proposed NCC loss performs better than typical one, since our proposed mesh-based patch warping can obtain accurate 3D position of each pixel. Also, the geometry cannot be recovered correctly with only SFS loss, as visual hull initialization is far from the ground truth. Once the coarse geometry is optimized by NCC loss, SFS loss can finetune the results to get more accurate results. The qualitative results are shown in Fig.~\ref{fig:ablation}. Using only NCC loss, we can obtain the coarse results while some details like tie, hands are not recovered well. SFS loss alone may stuck at local optima. Combining NCC loss and SFS loss, we can get accurate results without losing details.

\begin{table}
    \small
    \setlength{\tabcolsep}{0.7mm}{
	\centering
	\begin{tabular}{cccc|cc}
		\toprule
		Visual hull & Typical NCC &  Our NCC & SFS & Normal & Chamfer \\ \hline
		$ \checkmark $  &    &   & & 0.15 & 0.55        \\
        $ \checkmark $  & $ \checkmark $ &  &   & 0.13  & 0.32 \\
		$ \checkmark $  &    & $ \checkmark $  &    & 0.08   & 0.21     \\
	    $ \checkmark $  &    &    &  $\checkmark$ & 0.14 & 0.42        \\
		$ \checkmark $  &    & $ \checkmark $ & $ \checkmark $  & \textbf{0.06}  &  \textbf{0.18}  \\
		\bottomrule
	\end{tabular}
        \caption{\textbf{Quantitative Ablation Study on the Proposed Loss}. We evaluate the effect of the photometric consistency and shape from shading refinement.}
    \vspace{-0.2in}
	\label{tab:abl}
 }
\end{table}

\section{Limitations and Conclusions}

Our proposed patch warping module, like other MVS methods, may face limitations in accurately reconstructing texture-less objects or under complex lighting conditions. To address this issue, we plan to incorporate deep features for multi-view matching in the future to enhance the module's robustness. Moreover, it's important to note that our SFS module assumes Lambertian surfaces, which may not be suitable for reflective regions.

In this paper, we introduce FastHuman, an efficient coarse-to-fine approach to reconstruct high-quality human bodies from multi-view images. Our method utilizes a mesh-based patch warping optimization technique that leverages orientation point clouds shape representation and multi-view photometric consistency constraints. As the human body and clothing mostly have diffuse characteristics, we employ an image formation model with SH illumination and propose a shading refinement algorithm to recover surface albedo and fine-scale surface detail. To evaluate our approach, we conducted experiments on both synthetic and real-world datasets. Our promising results demonstrated that our approach can reconstruct high-quality human meshes in just a few minutes.

{
    \small
    \bibliographystyle{ieeenat_fullname}
    \bibliography{main}

\begin{thebibliography}{75}
\providecommand{\natexlab}[1]{#1}
\providecommand{\url}[1]{\texttt{#1}}
\expandafter\ifx\csname urlstyle\endcsname\relax
  \providecommand{\doi}[1]{doi: #1}\else
  \providecommand{\doi}{doi: \begingroup \urlstyle{rm}\Url}\fi

\bibitem[Agisoft(2019)]{Metashape}
Agisoft.
\newblock Metashape software.
\newblock \emph{retrieved 20.05}, 2019.

\bibitem[Alldieck et~al.(2018)Alldieck, Magnor, Xu, Theobalt, and Pons-Moll]{Alldieck2018CVPR}
Thiemo Alldieck, Marcus Magnor, Weipeng Xu, Christian Theobalt, and Gerard Pons-Moll.
\newblock Video based reconstruction of 3d people models.
\newblock In \emph{Proc. IEEE Conf. on Computer Vision and Pattern Recognition (CVPR)}, 2018.

\bibitem[Anguelov et~al.(2005)Anguelov, Srinivasan, Koller, Thrun, Rodgers, and Davis]{DBLP:journals/tog/SCAPE05}
Dragomir Anguelov, Praveen Srinivasan, Daphne Koller, Sebastian Thrun, Jim Rodgers, and James Davis.
\newblock {SCAPE:} shape completion and animation of people.
\newblock \emph{ACM Trans. Graph.}, 24\penalty0 (3):\penalty0 408--416, 2005.

\bibitem[Blender(2018)]{blender}
Blender, 2018.
\newblock \url{https://www.blender.org}.

\bibitem[Bogo et~al.(2016)Bogo, Kanazawa, Lassner, Gehler, Romero, and Black]{DBLP:conf/eccv/Simplify16}
Federica Bogo, Angjoo Kanazawa, Christoph Lassner, Peter~V. Gehler, Javier Romero, and Michael~J. Black.
\newblock Keep it {SMPL:} automatic estimation of 3d human pose and shape from a single image.
\newblock In \emph{Proc. of the European Conf. on Computer Vision (ECCV)}, pages 561--578, 2016.

\bibitem[Chen et~al.(2021)Chen, Zheng, Black, Hilliges, and Geiger]{Chen2021ICCV}
Xu Chen, Yufeng Zheng, Michael~J Black, Otmar Hilliges, and Andreas Geiger.
\newblock Snarf: Differentiable forward skinning for animating non-rigid neural implicit shapes.
\newblock In \emph{Proc. of the IEEE International Conf. on Computer Vision (ICCV)}, 2021.

\bibitem[Cristiani and Falcone(2007)]{DBLP:journals/siamnum/CristianiF07}
Emiliano Cristiani and Maurizio Falcone.
\newblock Fast semi-lagrangian schemes for the eikonal equation and applications.
\newblock \emph{{SIAM} J. Numer. Anal.}, 45\penalty0 (5):\penalty0 1979--2011, 2007.

\bibitem[Darmon et~al.(2022)Darmon, Bascle, Devaux, Monasse, and Aubry]{DBLP:conf/cvpr/neuralwarp22}
Fran{\c{c}}ois Darmon, B{\'{e}}n{\'{e}}dicte Bascle, Jean{-}Cl{\'{e}}ment Devaux, Pascal Monasse, and Mathieu Aubry.
\newblock Improving neural implicit surfaces geometry with patch warping.
\newblock In \emph{Proc. IEEE Conf. on Computer Vision and Pattern Recognition (CVPR)}, pages 6250--6259, 2022.

\bibitem[Engine(2019)]{unreal}
Unreal Engine, 2019.
\newblock \url{https://www.unrealengine.com/}.

\bibitem[Fridovich-Keil et~al.(2022)Fridovich-Keil, Yu, Tancik, Chen, Recht, and Kanazawa]{yu2022plenoxels}
Sara Fridovich-Keil, Alex Yu, Matthew Tancik, Qinhong Chen, Benjamin Recht, and Angjoo Kanazawa.
\newblock Plenoxels: Radiance fields without neural networks.
\newblock In \emph{Proc. IEEE Conf. on Computer Vision and Pattern Recognition (CVPR)}, 2022.

\bibitem[Furukawa and Ponce(2010)]{DBLP:journals/pami/mvs10}
Yasutaka Furukawa and Jean Ponce.
\newblock Accurate, dense, and robust multiview stereopsis.
\newblock \emph{IEEE Trans. on Pattern Analysis and Machine Intelligence (PAMI)}, 32\penalty0 (8):\penalty0 1362--1376, 2010.

\bibitem[Haefner et~al.(2018)Haefner, Qu{\'e}au, M{\"o}llenhoff, and Cremers]{Haefner2018CVPR}
Bjoern Haefner, Yvain Qu{\'e}au, Thomas M{\"o}llenhoff, and Daniel Cremers.
\newblock Fight ill-posedness with ill-posedness: Single-shot variational depth super-resolution from shading.
\newblock In \emph{Proc. IEEE Conf. on Computer Vision and Pattern Recognition (CVPR)}, 2018.

\bibitem[Haefner et~al.(2019)Haefner, Peng, Verma, Qu{\'e}au, and Cremers]{Haefner2019PAMI}
Bjoern Haefner, Songyou Peng, Alok Verma, Yvain Qu{\'e}au, and Daniel Cremers.
\newblock Photometric depth super-resolution.
\newblock \emph{IEEE Trans. on Pattern Analysis and Machine Intelligence (PAMI)}, 42\penalty0 (10):\penalty0 2453--2464, 2019.

\bibitem[Horn(1970)]{DBLP:phd/us/sfs70}
Berthold K.~P. Horn.
\newblock \emph{Shape from shading; a method for obtaining the shape of a smooth opaque object from one view}.
\newblock PhD thesis, Massachusetts Institute of Technology, {USA}, 1970.

\bibitem[Horn and Brooks(1986)]{DBLP:journals/cvgip/HornB86}
Berthold K.~P. Horn and Michael~J. Brooks.
\newblock The variational approach to shape from shading.
\newblock \emph{Comput. Vis. Graph. Image Process.}, 33\penalty0 (2):\penalty0 174--208, 1986.

\bibitem[Huang et~al.(2018)Huang, Matzen, Kopf, Ahuja, and Huang]{DBLP:conf/cvpr/deepmvs18}
Po{-}Han Huang, Kevin Matzen, Johannes Kopf, Narendra Ahuja, and Jia{-}Bin Huang.
\newblock Deepmvs: Learning multi-view stereopsis.
\newblock In \emph{Proc. IEEE Conf. on Computer Vision and Pattern Recognition (CVPR)}, pages 2821--2830, 2018.

\bibitem[Jensen et~al.(2014)Jensen, Dahl, Vogiatzis, Tola, and Aan{\ae}s]{DBLP:conf/cvpr/DTU14}
Rasmus~Ramsb{\o}l Jensen, Anders~Lindbjerg Dahl, George Vogiatzis, Engin Tola, and Henrik Aan{\ae}s.
\newblock Large scale multi-view stereopsis evaluation.
\newblock In \emph{Proc. IEEE Conf. on Computer Vision and Pattern Recognition (CVPR)}, pages 406--413, 2014.

\bibitem[Joo et~al.(2018)Joo, Simon, and Sheikh]{DBLP:conf/cvpr/TotalCapture18}
Hanbyul Joo, Tomas Simon, and Yaser Sheikh.
\newblock Total capture: {A} 3d deformation model for tracking faces, hands, and bodies.
\newblock In \emph{Proc. IEEE Conf. on Computer Vision and Pattern Recognition (CVPR)}, pages 8320--8329, 2018.

\bibitem[Kajiya(1986)]{DBLP:conf/siggraph/req86}
James~T. Kajiya.
\newblock The rendering equation.
\newblock In \emph{Proceedings of the Conference on Computer Graphics and Interactive Techniques, (SIGGRAPH)}, pages 143--150, 1986.

\bibitem[Kanazawa et~al.(2018)Kanazawa, Black, Jacobs, and Malik]{DBLP:conf/cvpr/HMR18}
Angjoo Kanazawa, Michael~J. Black, David~W. Jacobs, and Jitendra Malik.
\newblock End-to-end recovery of human shape and pose.
\newblock In \emph{Proc. IEEE Conf. on Computer Vision and Pattern Recognition (CVPR)}, pages 7122--7131, 2018.

\bibitem[Kazhdan and Hoppe(2013)]{DBLP:journals/tog/spr13}
Michael~M. Kazhdan and Hugues Hoppe.
\newblock Screened poisson surface reconstruction.
\newblock \emph{ACM Trans. Graph.}, 32\penalty0 (3):\penalty0 29:1--29:13, 2013.

\bibitem[Kazhdan et~al.(2006)Kazhdan, Bolitho, and Hoppe]{DBLP:conf/sgp/psr06}
Michael~M. Kazhdan, Matthew Bolitho, and Hugues Hoppe.
\newblock Poisson surface reconstruction.
\newblock In \emph{Proceedings of the Fourth Eurographics Symposium on Geometry Processing (SGP)}, pages 61--70, 2006.

\bibitem[Kingma and Ba(2015)]{DBLP:journals/corr/adam14}
Diederik~P. Kingma and Jimmy Ba.
\newblock Adam: {A} method for stochastic optimization.
\newblock In \emph{International Conference on Learning Representations, {ICLR}}, 2015.

\bibitem[Kolotouros et~al.(2019)Kolotouros, Pavlakos, Black, and Daniilidis]{DBLP:conf/iccv/SPIN19}
Nikos Kolotouros, Georgios Pavlakos, Michael~J. Black, and Kostas Daniilidis.
\newblock Learning to reconstruct 3d human pose and shape via model-fitting in the loop.
\newblock In \emph{Proc. of the IEEE International Conf. on Computer Vision (ICCV)}, pages 2252--2261, 2019.

\bibitem[Laine et~al.(2020)Laine, Hellsten, Karras, Seol, Lehtinen, and Aila]{Laine2020diffrast}
Samuli Laine, Janne Hellsten, Tero Karras, Yeongho Seol, Jaakko Lehtinen, and Timo Aila.
\newblock Modular primitives for high-performance differentiable rendering.
\newblock \emph{ACM Trans. Graph.}, 39\penalty0 (6), 2020.

\bibitem[Laurentini(1994)]{DBLP:journals/pami/visualhull94}
Aldo Laurentini.
\newblock The visual hull concept for silhouette-based image understanding.
\newblock \emph{IEEE Trans. on Pattern Analysis and Machine Intelligence (PAMI)}, 16\penalty0 (2):\penalty0 150--162, 1994.

\bibitem[Lin et~al.(2022)Lin, Yang, Saleemi, and Sengupta]{DBLP:conf/wacv/rvm22}
Shanchuan Lin, Linjie Yang, Imran Saleemi, and Soumyadip Sengupta.
\newblock Robust high-resolution video matting with temporal guidance.
\newblock In \emph{Proc. of Winter Conf. on Applications of Computer Vision (WACV)}, pages 3132--3141, 2022.

\bibitem[Lionar et~al.(2021)Lionar, Emtsev, Svilarkovic, and Peng]{Lionar2021WACV}
Stefan Lionar, Daniil Emtsev, Dusan Svilarkovic, and Songyou Peng.
\newblock Dynamic plane convolutional occupancy networks.
\newblock In \emph{Proc. of Winter Conf. on Applications of Computer Vision (WACV)}, 2021.

\bibitem[Liu et~al.(2020)Liu, Zhang, Peng, Shi, Pollefeys, and Cui]{Liu2020CVPR}
Shaohui Liu, Yinda Zhang, Songyou Peng, Boxin Shi, Marc Pollefeys, and Zhaopeng Cui.
\newblock Dist: Rendering deep implicit signed distance function with differentiable sphere tracing.
\newblock In \emph{Proc. IEEE Conf. on Computer Vision and Pattern Recognition (CVPR)}, 2020.

\bibitem[Liu et~al.(2009)Liu, Dai, and Xu]{Liu2009TVCG}
Yebin Liu, Qionghai Dai, and Wenli Xu.
\newblock A point-cloud-based multiview stereo algorithm for free-viewpoint video.
\newblock \emph{IEEE Trans. Vis. Comput. Graph.}, 2009.

\bibitem[Loper et~al.(2015)Loper, Mahmood, Romero, Pons{-}Moll, and Black]{DBLP:journals/tog/SMPL15}
Matthew Loper, Naureen Mahmood, Javier Romero, Gerard Pons{-}Moll, and Michael~J. Black.
\newblock {SMPL:} a skinned multi-person linear model.
\newblock \emph{ACM Trans. Graph.}, 34\penalty0 (6):\penalty0 248:1--248:16, 2015.

\bibitem[Lorensen and Cline(1987)]{DBLP:conf/siggraph/Marchingcubes87}
William~E. Lorensen and Harvey~E. Cline.
\newblock Marching cubes: {A} high resolution 3d surface construction algorithm.
\newblock In \emph{Proceedings of the Conference on Computer Graphics and Interactive Techniques, (SIGGRAPH)}, pages 163--169, 1987.

\bibitem[Mescheder et~al.(2019)Mescheder, Oechsle, Niemeyer, Nowozin, and Geiger]{Mescheder2019CVPR}
Lars Mescheder, Michael Oechsle, Michael Niemeyer, Sebastian Nowozin, and Andreas Geiger.
\newblock Occupancy networks: Learning 3d reconstruction in function space.
\newblock In \emph{Proc. IEEE Conf. on Computer Vision and Pattern Recognition (CVPR)}, 2019.

\bibitem[Mildenhall et~al.(2020)Mildenhall, Srinivasan, Tancik, Barron, Ramamoorthi, and Ng]{DBLP:conf/eccv/nerf20}
Ben Mildenhall, Pratul~P. Srinivasan, Matthew Tancik, Jonathan~T. Barron, Ravi Ramamoorthi, and Ren Ng.
\newblock Nerf: Representing scenes as neural radiance fields for view synthesis.
\newblock In \emph{Proc. of the European Conf. on Computer Vision (ECCV)}, 2020.

\bibitem[M\"uller et~al.(2022)M\"uller, Evans, Schied, and Keller]{mueller2022instant}
Thomas M\"uller, Alex Evans, Christoph Schied, and Alexander Keller.
\newblock Instant neural graphics primitives with a multiresolution hash encoding.
\newblock \emph{ACM Trans. Graph.}, 41\penalty0 (4):\penalty0 102:1--102:15, 2022.

\bibitem[Niemeyer et~al.(2020)Niemeyer, Mescheder, Oechsle, and Geiger]{Niemeyer2020CVPR}
Michael Niemeyer, Lars Mescheder, Michael Oechsle, and Andreas Geiger.
\newblock Differentiable volumetric rendering: Learning implicit 3d representations without 3d supervision.
\newblock In \emph{Proc. IEEE Conf. on Computer Vision and Pattern Recognition (CVPR)}, 2020.

\bibitem[Oechsle et~al.(2021)Oechsle, Peng, and Geiger]{Oechsle2021ICCV}
Michael Oechsle, Songyou Peng, and Andreas Geiger.
\newblock Unisurf: Unifying neural implicit surfaces and radiance fields for multi-view reconstruction.
\newblock In \emph{Proc. of the IEEE International Conf. on Computer Vision (ICCV)}, 2021.

\bibitem[Or{-}El et~al.(2015)Or{-}El, Rosman, Wetzler, Kimmel, and Bruckstein]{DBLP:conf/cvpr/sfs15}
Roy Or{-}El, Guy Rosman, Aaron Wetzler, Ron Kimmel, and Alfred~M. Bruckstein.
\newblock Rgbd-fusion: Real-time high precision depth recovery.
\newblock In \emph{Proc. IEEE Conf. on Computer Vision and Pattern Recognition (CVPR)}, pages 5407--5416, 2015.

\bibitem[Park et~al.(2019)Park, Florence, Straub, Newcombe, and Lovegrove]{Park2019CVPR}
Jeong~Joon Park, Peter Florence, Julian Straub, Richard Newcombe, and Steven Lovegrove.
\newblock Deepsdf: Learning continuous signed distance functions for shape representation.
\newblock In \emph{Proc. IEEE Conf. on Computer Vision and Pattern Recognition (CVPR)}, 2019.

\bibitem[Pavlakos et~al.(2019)Pavlakos, Choutas, Ghorbani, Bolkart, Osman, Tzionas, and Black]{DBLP:conf/cvpr/SMPLX19}
Georgios Pavlakos, Vasileios Choutas, Nima Ghorbani, Timo Bolkart, Ahmed A.~A. Osman, Dimitrios Tzionas, and Michael~J. Black.
\newblock Expressive body capture: 3d hands, face, and body from a single image.
\newblock In \emph{Proc. IEEE Conf. on Computer Vision and Pattern Recognition (CVPR)}, pages 10975--10985, 2019.

\bibitem[Peng et~al.(2017)Peng, Haefner, Qu{\'e}au, and Cremers]{Peng2017ICCVW}
Songyou Peng, Bjoern Haefner, Yvain Qu{\'e}au, and Daniel Cremers.
\newblock Depth super-resolution meets uncalibrated photometric stereo.
\newblock In \emph{Proc. of the IEEE International Conf. on Computer Vision Workshops (ICCVW)}, 2017.

\bibitem[Peng et~al.(2020)Peng, Niemeyer, Mescheder, Pollefeys, and Geiger]{Peng2020ECCV}
Songyou Peng, Michael Niemeyer, Lars Mescheder, Marc Pollefeys, and Andreas Geiger.
\newblock Convolutional occupancy networks.
\newblock In \emph{Proc. of the European Conf. on Computer Vision (ECCV)}, 2020.

\bibitem[Peng et~al.(2021{\natexlab{a}})Peng, Dong, Wang, Zhang, Shuai, Zhou, and Bao]{DBLP:conf/iccv/animatenerf21}
Sida Peng, Junting Dong, Qianqian Wang, Shangzhan Zhang, Qing Shuai, Xiaowei Zhou, and Hujun Bao.
\newblock Animatable neural radiance fields for modeling dynamic human bodies.
\newblock In \emph{Proc. of the IEEE International Conf. on Computer Vision (ICCV)}, pages 14294--14303, 2021{\natexlab{a}}.

\bibitem[Peng et~al.(2021{\natexlab{b}})Peng, Jiang, Liao, Niemeyer, Pollefeys, and Geiger]{Peng2021SAP}
Songyou Peng, Chiyu~"Max" Jiang, Yiyi Liao, Michael Niemeyer, Marc Pollefeys, and Andreas Geiger.
\newblock Shape as points: A differentiable poisson solver.
\newblock In \emph{Advances in Neural Information Processing Systems (NeurIPS)}, 2021{\natexlab{b}}.

\bibitem[Peng et~al.(2021{\natexlab{c}})Peng, Zhang, Xu, Wang, Shuai, Bao, and Zhou]{peng2021neural}
Sida Peng, Yuanqing Zhang, Yinghao Xu, Qianqian Wang, Qing Shuai, Hujun Bao, and Xiaowei Zhou.
\newblock Neural body: Implicit neural representations with structured latent codes for novel view synthesis of dynamic humans.
\newblock In \emph{Proc. IEEE Conf. on Computer Vision and Pattern Recognition (CVPR)}, 2021{\natexlab{c}}.

\bibitem[Qu{\'{e}}au et~al.(2017)Qu{\'{e}}au, M{\'{e}}lou, Castan, Cremers, and Durou]{DBLP:conf/emmcvpr/QueauMCCD17}
Yvain Qu{\'{e}}au, Jean M{\'{e}}lou, Fabien Castan, Daniel Cremers, and Jean{-}Denis Durou.
\newblock A variational approach to shape-from-shading under natural illumination.
\newblock In \emph{IEEE Conf. Comput. Vis. Pattern Recog. Worksh.}, pages 342--357, 2017.

\bibitem[Reiser et~al.(2021)Reiser, Peng, Liao, and Geiger]{Reiser2021ICCV}
Christian Reiser, Songyou Peng, Yiyi Liao, and Andreas Geiger.
\newblock Kilonerf: Speeding up neural radiance fields with thousands of tiny mlps.
\newblock In \emph{Proceedings of the IEEE/CVF International Conference on Computer Vision}, 2021.

\bibitem[Remelli et~al.(2020)Remelli, Lukoianov, Richter, Guillard, Bagautdinov, Baqu{\'{e}}, and Fua]{DBLP:conf/nips/meshsdf20}
Edoardo Remelli, Artem Lukoianov, Stephan~R. Richter, Beno{\^{\i}}t Guillard, Timur~M. Bagautdinov, Pierre Baqu{\'{e}}, and Pascal Fua.
\newblock Meshsdf: Differentiable iso-surface extraction.
\newblock In \emph{Advances in Neural Information Processing Systems (NeurIPS)}, 2020.

\bibitem[Renderpeople(2018)]{renderpeople}
Renderpeople, 2018.
\newblock \url{https://renderpeople.com/3d-people}.

\bibitem[Saito et~al.(2019)Saito, Huang, Natsume, Morishima, Li, and Kanazawa]{DBLP:conf/iccv/PIFu19}
Shunsuke Saito, Zeng Huang, Ryota Natsume, Shigeo Morishima, Hao Li, and Angjoo Kanazawa.
\newblock Pifu: Pixel-aligned implicit function for high-resolution clothed human digitization.
\newblock In \emph{Proc. of the IEEE International Conf. on Computer Vision (ICCV)}, pages 2304--2314, 2019.

\bibitem[Saito et~al.(2020)Saito, Simon, Saragih, and Joo]{DBLP:conf/cvpr/pifuhd20}
Shunsuke Saito, Tomas Simon, Jason~M. Saragih, and Hanbyul Joo.
\newblock Pifuhd: Multi-level pixel-aligned implicit function for high-resolution 3d human digitization.
\newblock In \emph{Proc. IEEE Conf. on Computer Vision and Pattern Recognition (CVPR)}, pages 81--90, 2020.

\bibitem[Sang et~al.(2020)Sang, Haefner, and Cremers]{Sang2020WACV}
Lu Sang, Bjoern Haefner, and Daniel Cremers.
\newblock Inferring super-resolution depth from a moving light-source enhanced rgb-d sensor: a variational approach.
\newblock In \emph{Proc. of Winter Conf. on Applications of Computer Vision (WACV)}, 2020.

\bibitem[Sch{\"{o}}nberger et~al.(2016)Sch{\"{o}}nberger, Zheng, Frahm, and Pollefeys]{DBLP:conf/eccv/colmap16}
Johannes~L. Sch{\"{o}}nberger, Enliang Zheng, Jan{-}Michael Frahm, and Marc Pollefeys.
\newblock Pixelwise view selection for unstructured multi-view stereo.
\newblock In \emph{Proc. of the European Conf. on Computer Vision (ECCV)}, pages 501--518, 2016.

\bibitem[Shao et~al.(2022{\natexlab{a}})Shao, Zhang, Zhang, Chen, Cao, Yu, and Liu]{DBLP:conf/cvpr/doublefieldL22}
Ruizhi Shao, Hongwen Zhang, He Zhang, Mingjia Chen, Yanpei Cao, Tao Yu, and Yebin Liu.
\newblock Doublefield: Bridging the neural surface and radiance fields for high-fidelity human reconstruction and rendering.
\newblock In \emph{Proc. IEEE Conf. on Computer Vision and Pattern Recognition (CVPR)}, pages 15851--15861, 2022{\natexlab{a}}.

\bibitem[Shao et~al.(2022{\natexlab{b}})Shao, Zheng, Zhang, Sun, and Liu]{shao2022diffustereo}
Ruizhi Shao, Zerong Zheng, Hongwen Zhang, Jingxiang Sun, and Yebin Liu.
\newblock Diffustereo: High quality human reconstruction via diffusion-based stereo using sparse cameras.
\newblock In \emph{Proc. of the European Conf. on Computer Vision (ECCV)}, 2022{\natexlab{b}}.

\bibitem[Starck and Hilton(2007)]{Starck2007surface}
Jonathan Starck and Adrian Hilton.
\newblock Surface capture for performance-based animation.
\newblock \emph{IEEE computer graphics and applications}, 2007.

\bibitem[Sumner and Popovic(2004)]{DBLP:journals/tog/deformtrans04}
Robert~W. Sumner and Jovan Popovic.
\newblock Deformation transfer for triangle meshes.
\newblock \emph{ACM Trans. Graph.}, 23\penalty0 (3):\penalty0 399--405, 2004.

\bibitem[Vlasic et~al.(2009)Vlasic, Peers, Baran, Debevec, Popovi{\'c}, Rusinkiewicz, and Matusik]{Vlasic2009SIGGRAPH}
Daniel Vlasic, Pieter Peers, Ilya Baran, Paul Debevec, Jovan Popovi{\'c}, Szymon Rusinkiewicz, and Wojciech Matusik.
\newblock Dynamic shape capture using multi-view photometric stereo.
\newblock \emph{ACM SIGGRAPH Asia}, 2009.

\bibitem[Wang et~al.(2021)Wang, Liu, Liu, Theobalt, Komura, and Wang]{DBLP:conf/nips/neus21}
Peng Wang, Lingjie Liu, Yuan Liu, Christian Theobalt, Taku Komura, and Wenping Wang.
\newblock Neus: Learning neural implicit surfaces by volume rendering for multi-view reconstruction.
\newblock In \emph{Advances in Neural Information Processing Systems (NeurIPS)}, pages 27171--27183, 2021.

\bibitem[Wang et~al.(2022)Wang, Schwarz, Geiger, and Tang]{Wang2022ECCV}
Shaofei Wang, Katja Schwarz, Andreas Geiger, and Siyu Tang.
\newblock Arah: Animatable volume rendering of articulated human sdfs.
\newblock In \emph{Proc. of the European Conf. on Computer Vision (ECCV)}, 2022.

\bibitem[Weng et~al.(2022)Weng, Curless, Srinivasan, Barron, and Kemelmacher-Shlizerman]{humannerf22}
Chung-Yi Weng, Brian Curless, Pratul~P. Srinivasan, Jonathan~T. Barron, and Ira Kemelmacher-Shlizerman.
\newblock Human{N}e{RF}: Free-viewpoint rendering of moving people from monocular video.
\newblock In \emph{Proc. IEEE Conf. on Computer Vision and Pattern Recognition (CVPR)}, pages 16210--16220, 2022.

\bibitem[Woodham(1980)]{woodham1980photometric}
Robert~J Woodham.
\newblock Photometric method for determining surface orientation from multiple images.
\newblock \emph{Optical engineering}, 19\penalty0 (1):\penalty0 139--144, 1980.

\bibitem[Wu et~al.(2011)Wu, Varanasi, Liu, Seidel, and Theobalt]{Wu2011ICCV}
Chenglei Wu, Kiran Varanasi, Yebin Liu, Hans-Peter Seidel, and Christian Theobalt.
\newblock Shading-based dynamic shape refinement from multi-view video under general illumination.
\newblock In \emph{Proc. of the IEEE International Conf. on Computer Vision (ICCV)}, 2011.

\bibitem[Wu et~al.(2014)Wu, Zollh\"{o}fer, Nie\ss{}ner, Stamminger, Izadi, and Theobalt]{10.1145/sfs14}
Chenglei Wu, Michael Zollh\"{o}fer, Matthias Nie\ss{}ner, Marc Stamminger, Shahram Izadi, and Christian Theobalt.
\newblock Real-time shading-based refinement for consumer depth cameras.
\newblock \emph{ACM Trans. Graph.}, 33, 2014.

\bibitem[Wu et~al.(2020)Wu, Wang, Hu, and Yu]{wu2020multi}
Minye Wu, Yuehao Wang, Qiang Hu, and Jingyi Yu.
\newblock Multi-view neural human rendering.
\newblock In \emph{Proc. IEEE Conf. on Computer Vision and Pattern Recognition (CVPR)}, pages 1682--1691, 2020.

\bibitem[Xie et~al.(2022)Xie, Takikawa, Saito, Litany, Yan, Khan, Tombari, Tompkin, Sitzmann, and Sridhar]{Xie2022EUROGRPHICS}
Yiheng Xie, Towaki Takikawa, Shunsuke Saito, Or Litany, Shiqin Yan, Numair Khan, Federico Tombari, James Tompkin, Vincent Sitzmann, and Srinath Sridhar.
\newblock Neural fields in visual computing and beyond.
\newblock In \emph{Computer Graphics Forum}, 2022.

\bibitem[Xu et~al.(2020)Xu, Bazavan, Zanfir, Freeman, Sukthankar, and Sminchisescu]{Xu2020CVPR}
Hongyi Xu, Eduard~Gabriel Bazavan, Andrei Zanfir, William~T. Freeman, Rahul Sukthankar, and Cristian Sminchisescu.
\newblock Ghum \& ghuml: Generative 3d human shape and articulated pose models.
\newblock In \emph{Proc. IEEE Conf. on Computer Vision and Pattern Recognition (CVPR)}, 2020.

\bibitem[Yao et~al.(2018)Yao, Luo, Li, Fang, and Quan]{DBLP:conf/eccv/mvsnet18}
Yao Yao, Zixin Luo, Shiwei Li, Tian Fang, and Long Quan.
\newblock Mvsnet: Depth inference for unstructured multi-view stereo.
\newblock In \emph{Proc. of the European Conf. on Computer Vision (ECCV)}, pages 785--801, 2018.

\bibitem[Yariv et~al.(2020)Yariv, Kasten, Moran, Galun, Atzmon, Basri, and Lipman]{DBLP:conf/nips/idr20}
Lior Yariv, Yoni Kasten, Dror Moran, Meirav Galun, Matan Atzmon, Ronen Basri, and Yaron Lipman.
\newblock Multiview neural surface reconstruction by disentangling geometry and appearance.
\newblock In \emph{Advances in Neural Information Processing Systems (NeurIPS)}, 2020.

\bibitem[Yariv et~al.(2021)Yariv, Gu, Kasten, and Lipman]{yariv2021volume}
Lior Yariv, Jiatao Gu, Yoni Kasten, and Yaron Lipman.
\newblock Volume rendering of neural implicit surfaces.
\newblock In \emph{Advances in Neural Information Processing Systems (NeurIPS)}, 2021.

\bibitem[Yu et~al.(2013)Yu, Yeung, Tai, and Lin]{DBLP:conf/cvpr/sfs13}
Lap{-}Fai Yu, Sai~Kit Yeung, Yu{-}Wing Tai, and Stephen Lin.
\newblock Shading-based shape refinement of {RGB-D} images.
\newblock In \emph{Proc. IEEE Conf. on Computer Vision and Pattern Recognition (CVPR)}, pages 1415--1422, 2013.

\bibitem[Yu et~al.(2022)Yu, Peng, Niemeyer, Sattler, and Geiger]{Yu2022NEURIPS}
Zehao Yu, Songyou Peng, Michael Niemeyer, Torsten Sattler, and Andreas Geiger.
\newblock Monosdf: Exploring monocular geometric cues for neural implicit surface reconstruction.
\newblock \emph{Advances in Neural Information Processing Systems (NeurIPS)}, 2022.

\bibitem[Zhang et~al.(1999)Zhang, Tsai, Cryer, and Shah]{Zhang1999PAMI}
Ruo Zhang, Ping-Sing Tsai, James~Edwin Cryer, and Mubarak Shah.
\newblock Shape-from-shading: a survey.
\newblock \emph{IEEE Trans. on Pattern Analysis and Machine Intelligence (PAMI)}, 1999.

\bibitem[Zhu et~al.(2022)Zhu, Peng, Larsson, Xu, Bao, Cui, Oswald, and Pollefeys]{Zhu2022CVPR}
Zihan Zhu, Songyou Peng, Viktor Larsson, Weiwei Xu, Hujun Bao, Zhaopeng Cui, Martin~R Oswald, and Marc Pollefeys.
\newblock Nice-slam: Neural implicit scalable encoding for slam.
\newblock In \emph{Proc. IEEE Conf. on Computer Vision and Pattern Recognition (CVPR)}, 2022.

\bibitem[Zollh{\"{o}}fer et~al.(2015)Zollh{\"{o}}fer, Dai, Innmann, Wu, Stamminger, Theobalt, and Nie{\ss}ner]{DBLP:journals/tog/ZollhoferDIWSTN15}
Michael Zollh{\"{o}}fer, Angela Dai, Matthias Innmann, Chenglei Wu, Marc Stamminger, Christian Theobalt, and Matthias Nie{\ss}ner.
\newblock Shading-based refinement on volumetric signed distance functions.
\newblock \emph{ACM Trans. Graph.}, 34\penalty0 (4):\penalty0 96:1--96:14, 2015.

\end{thebibliography}
}
\clearpage
\maketitlesupplementary

\begin{figure}
	\centering
	\begin{tabular}
		{@{\hskip2pt}c@{\hskip2pt}@{\hskip2pt}c@{\hskip2pt}@{\hskip2pt}c@{\hskip2pt}@{\hskip2pt}c@{\hskip2pt}}
		\includegraphics[width=0.11\textwidth,trim=250 150 400 100,clip]{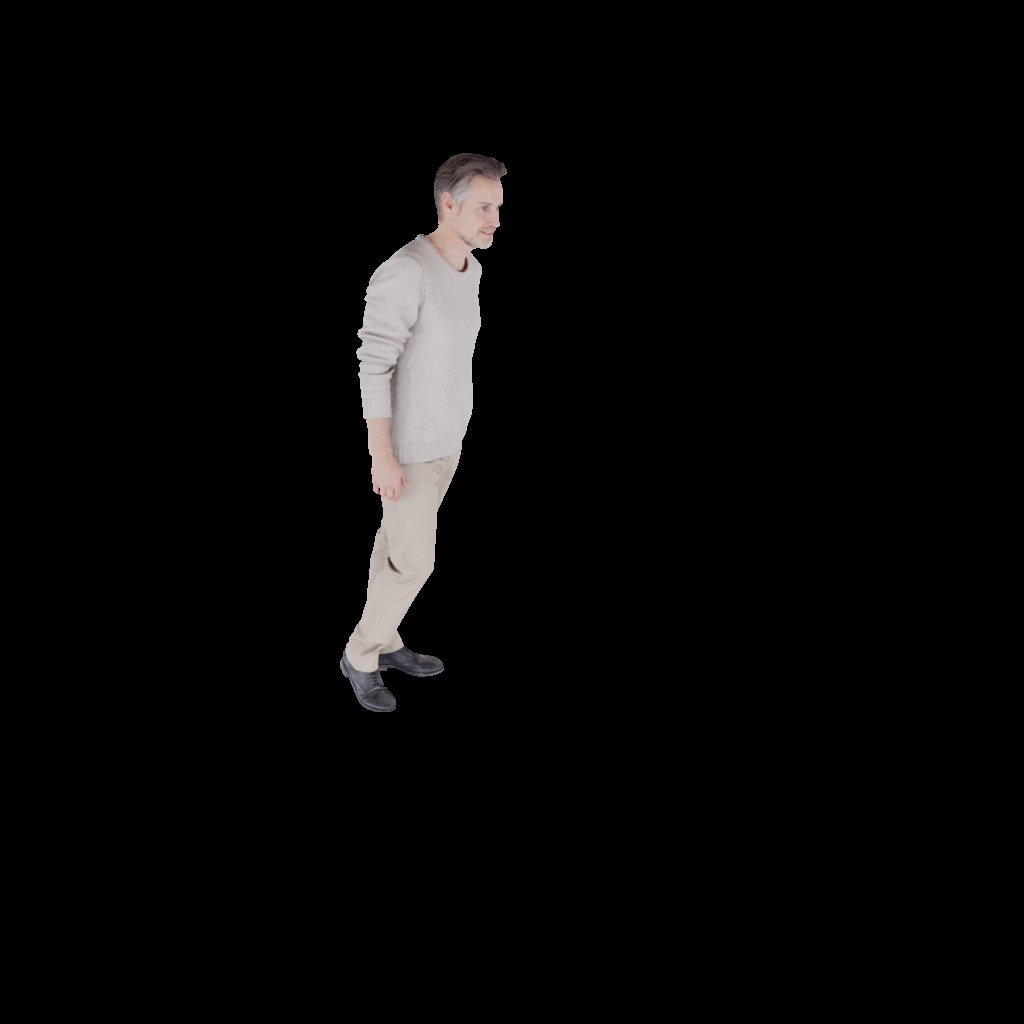} &
		\includegraphics[width=0.11\textwidth,trim=450 150 200 100,clip]{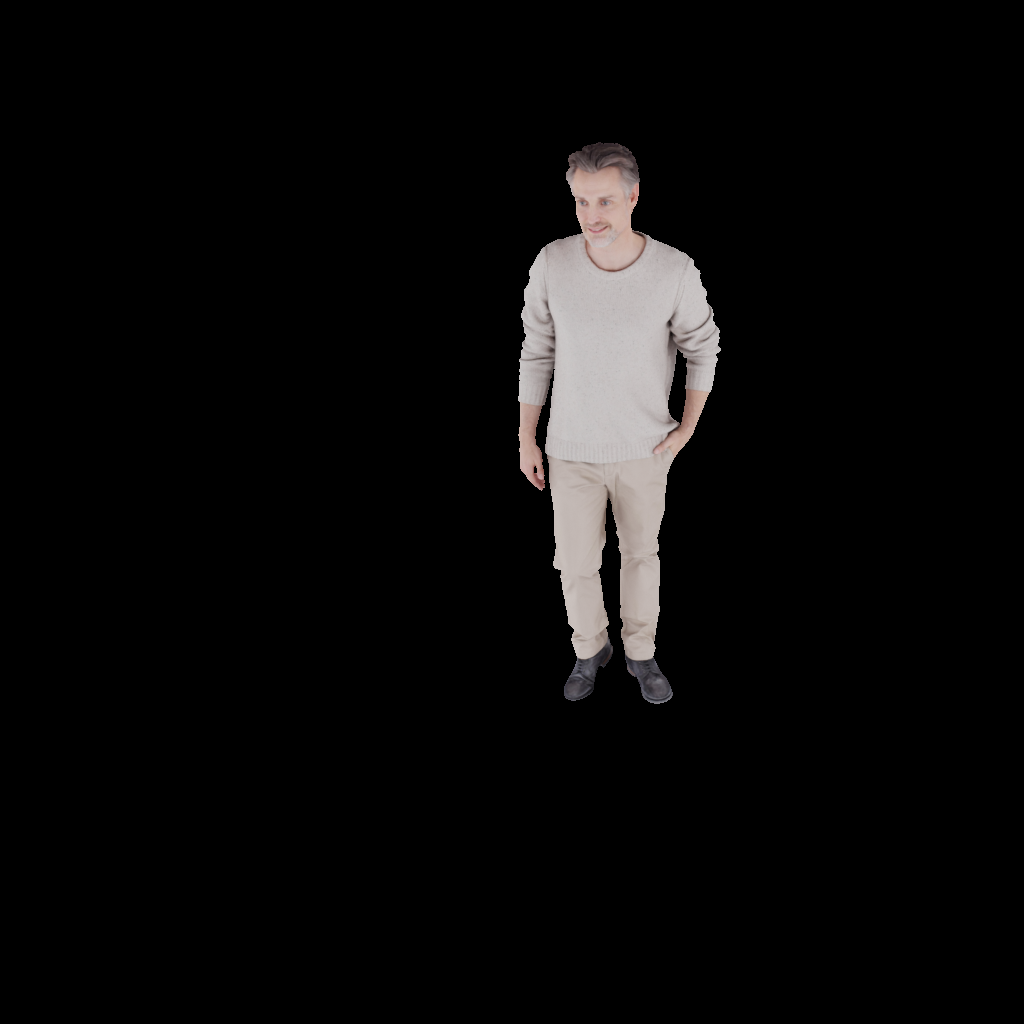} &
		\includegraphics[width=0.11\textwidth,trim=500 150 150 100,clip]{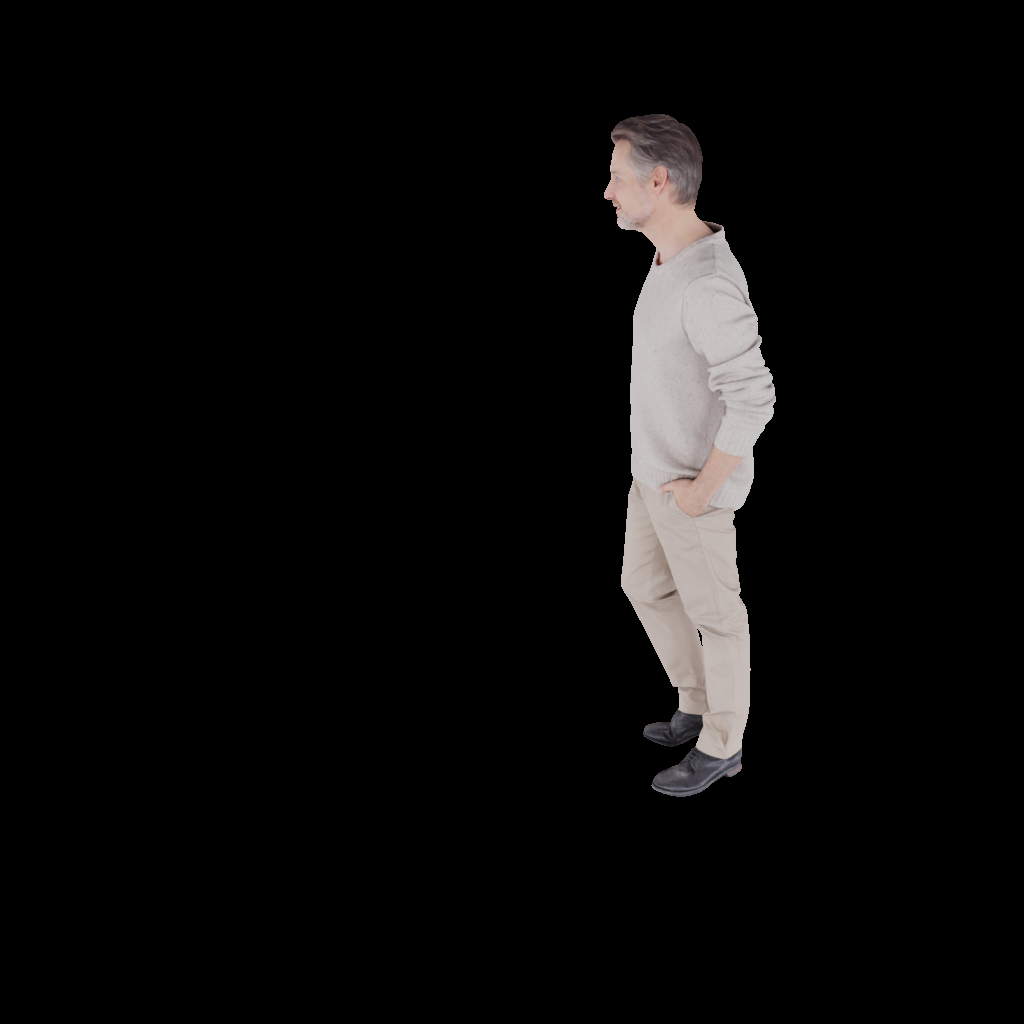} &
		\includegraphics[width=0.11\textwidth,trim=250 50 400 200,clip]{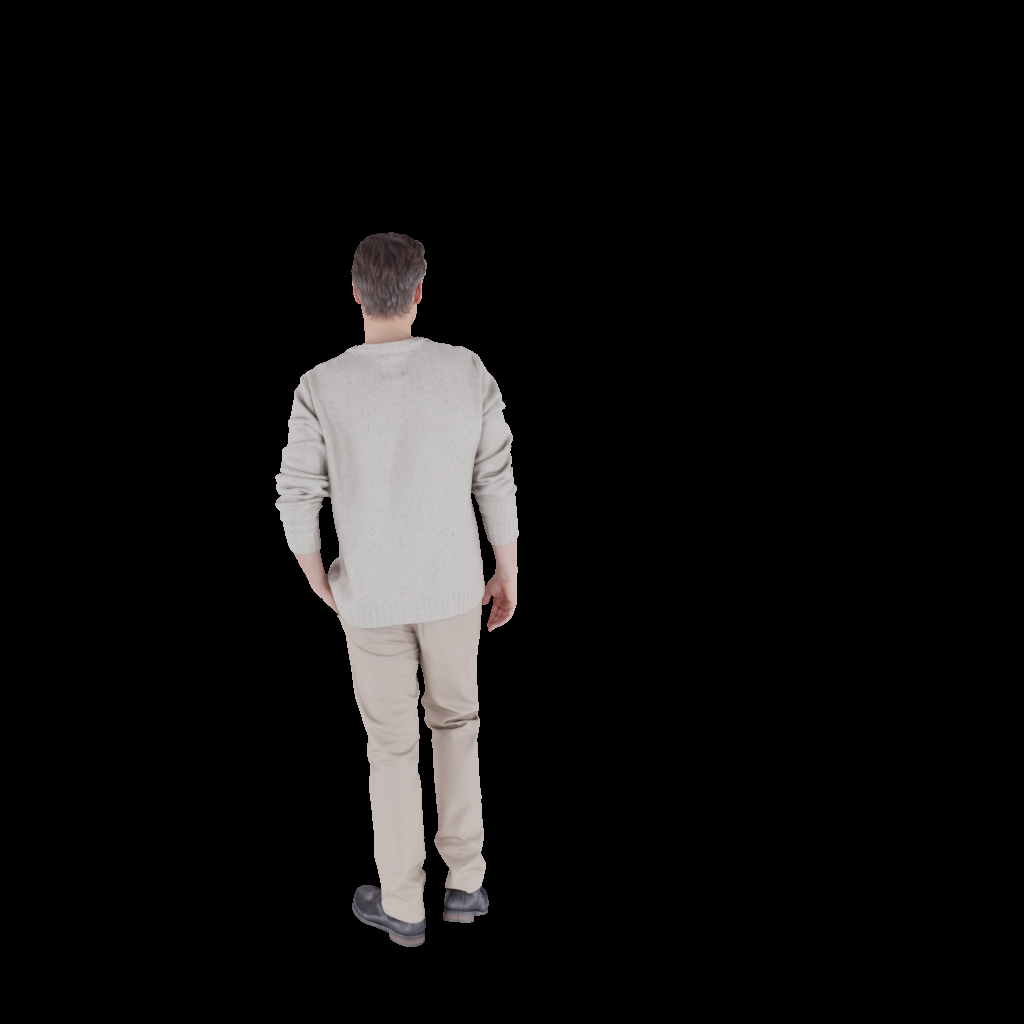} \\
		\includegraphics[width=0.11\textwidth,trim=250 150 400 100,clip]{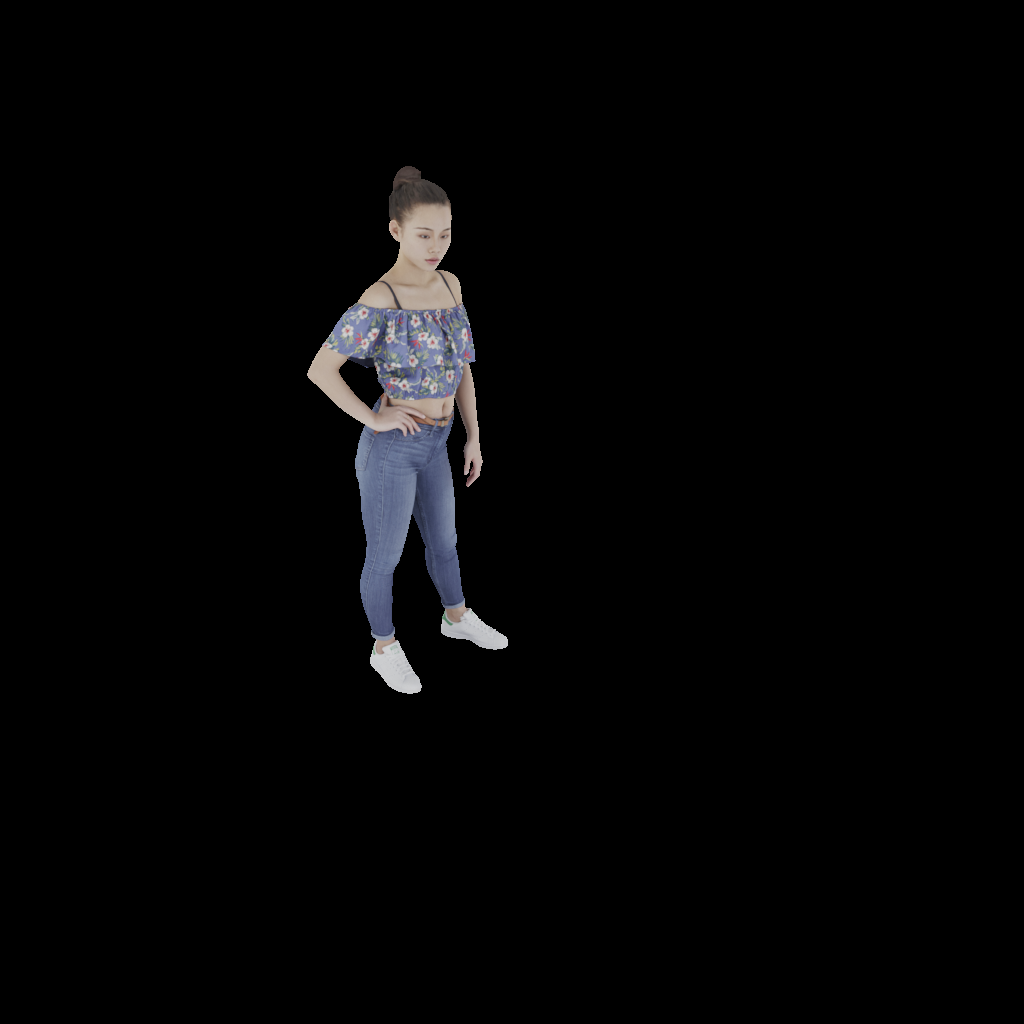} &
		\includegraphics[width=0.11\textwidth,trim=450 150 200 100,clip]{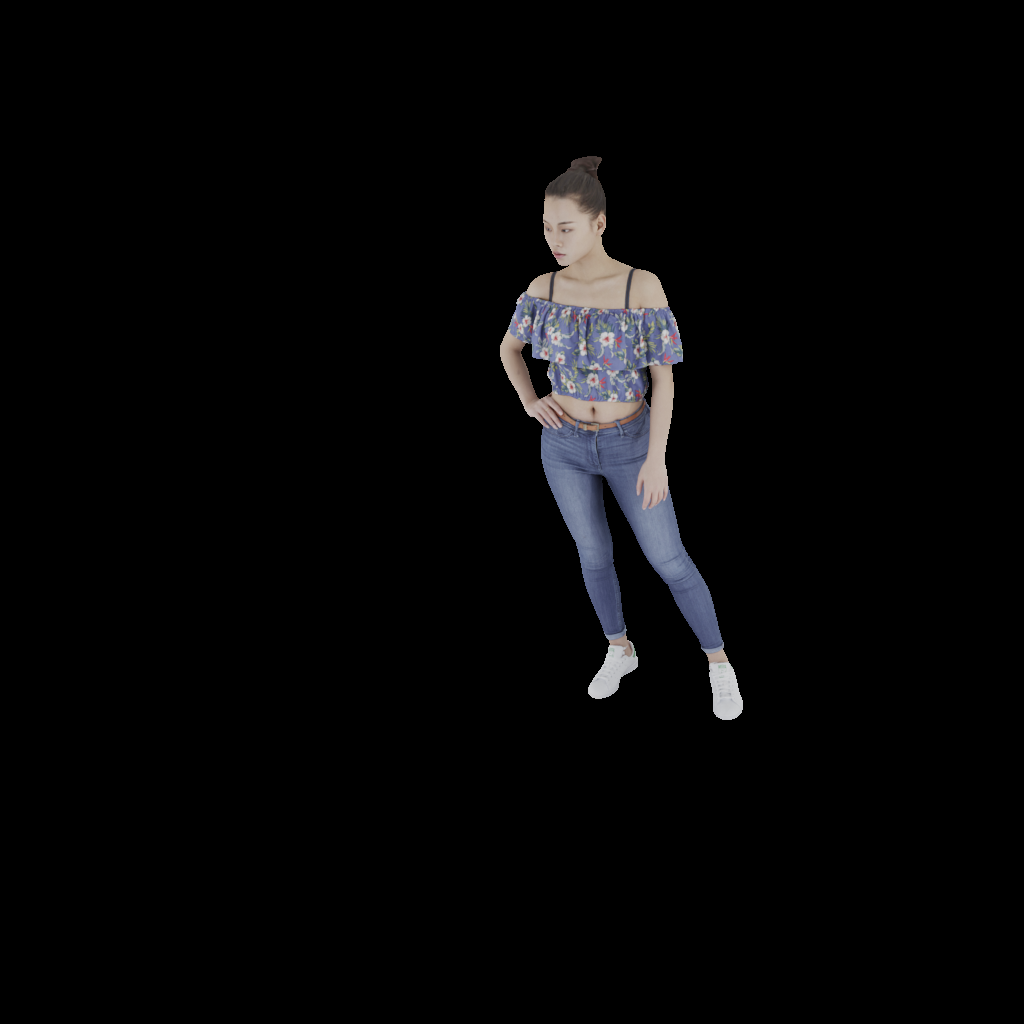} &
		\includegraphics[width=0.11\textwidth,trim=500 150 150 100,clip]{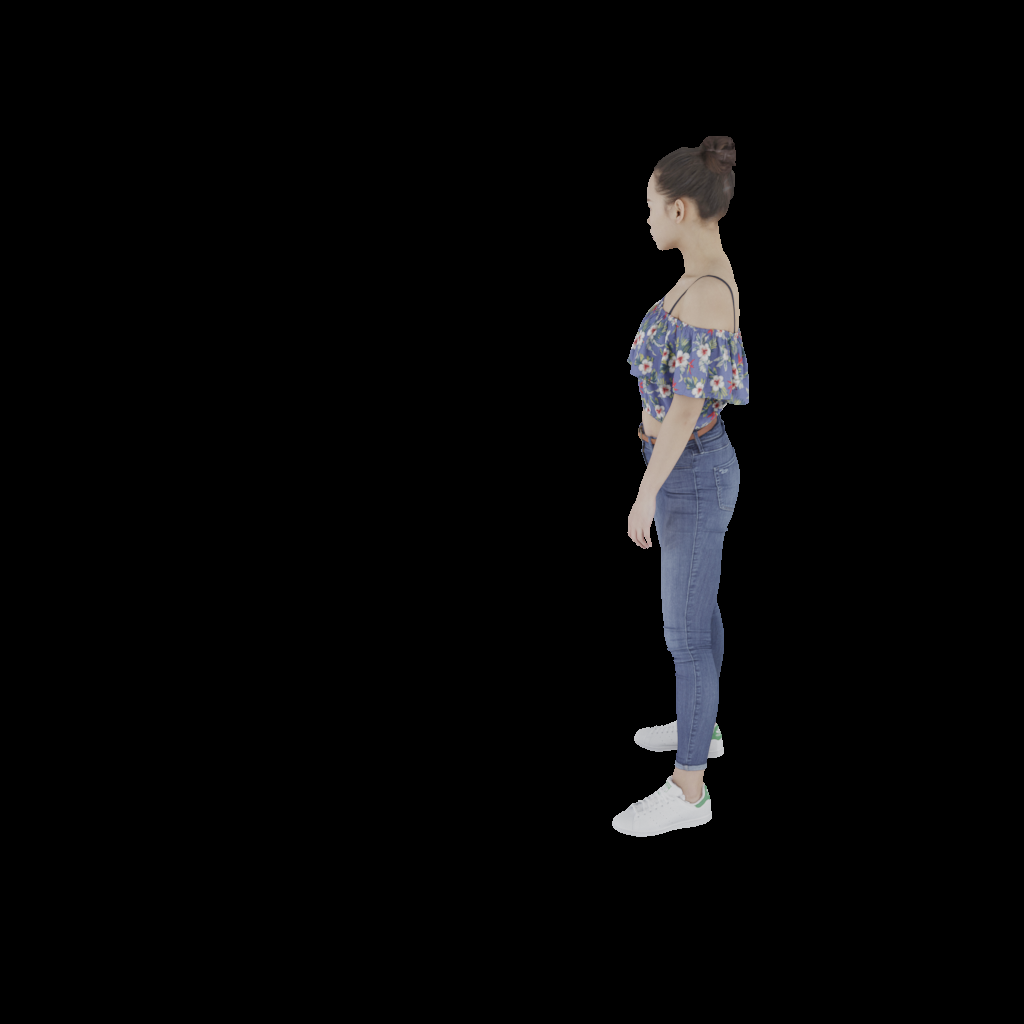} &
		\includegraphics[width=0.11\textwidth,trim=250 50 400 200,clip]{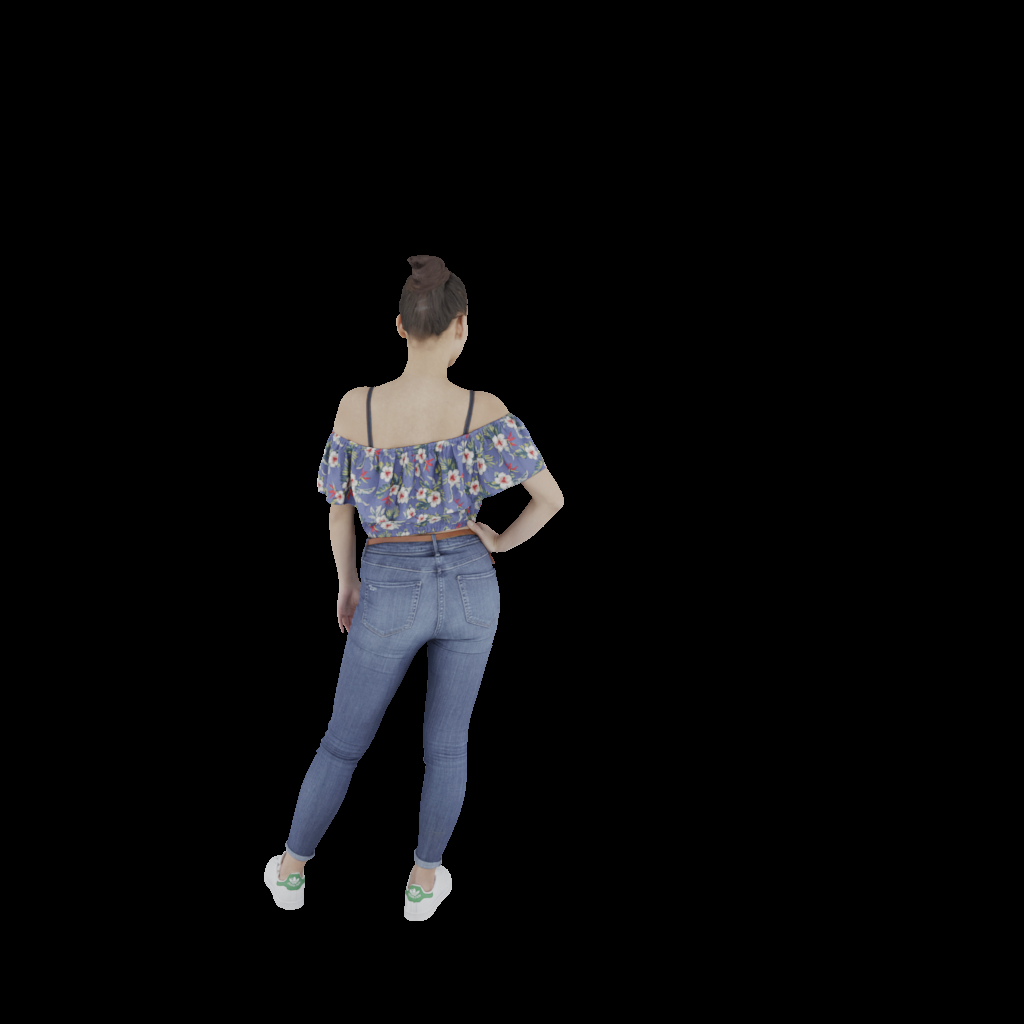} \\
	\end{tabular}
	
	\caption{Example of our synthesized dataset. Our synthesized dataset can provide realistic images.}
	\label{fig:rp}
	\vspace{-0.15in}
\end{figure}

\section{Ablation Studies}

In this section, we present our synthesized dataset and results of ablation studies. Fig~\ref{fig:rp} shows the images of our synthesized dataset. It can be seen taht our synthesized datset provide realistic images. We conduct ablation studies on the number of oriented points, sig and  patch size. In our paper we use 50K oriented points, sig=4 and the patch size is $11\times11$. Table~\ref{tab:abl} shows that our proposed module is robust to the parameters. Different parameters have little effect on the reconstruction results.

\begin{table}
	\centering
	\begin{tabular}{c|cc}
		\toprule
		& Normal C. &  Chamfer-$L_1$ \\ \hline
		baseline  &  0.08  & 0.21    \\
		10K points & 0.09 & 0.21  \\
		100K points &  0.08  & 0.21 \\
		sig = 1 & 0.10 & 0.24 \\
		sig = 10 & 0.13 & 0.31 \\
		$5\times5$ patch size &  0.09  & 0.22 \\
		$15\times15$ patch size &0.08 & 0.22 \\
		\bottomrule
	\end{tabular}
	\caption{Quantitative Ablation Study on the parameter. We evaluate the effect of the number of oriented points and patch size.}
	\vspace{-0.15in}
	\label{tab:abl}
\end{table}

\begin{figure}
	\centering
	\begin{tabular}
		{@{\hskip2pt}c@{\hskip2pt}@{\hskip2pt}c@{\hskip2pt}@{\hskip2pt}c@{\hskip2pt}}
		\includegraphics[width=0.15\textwidth]{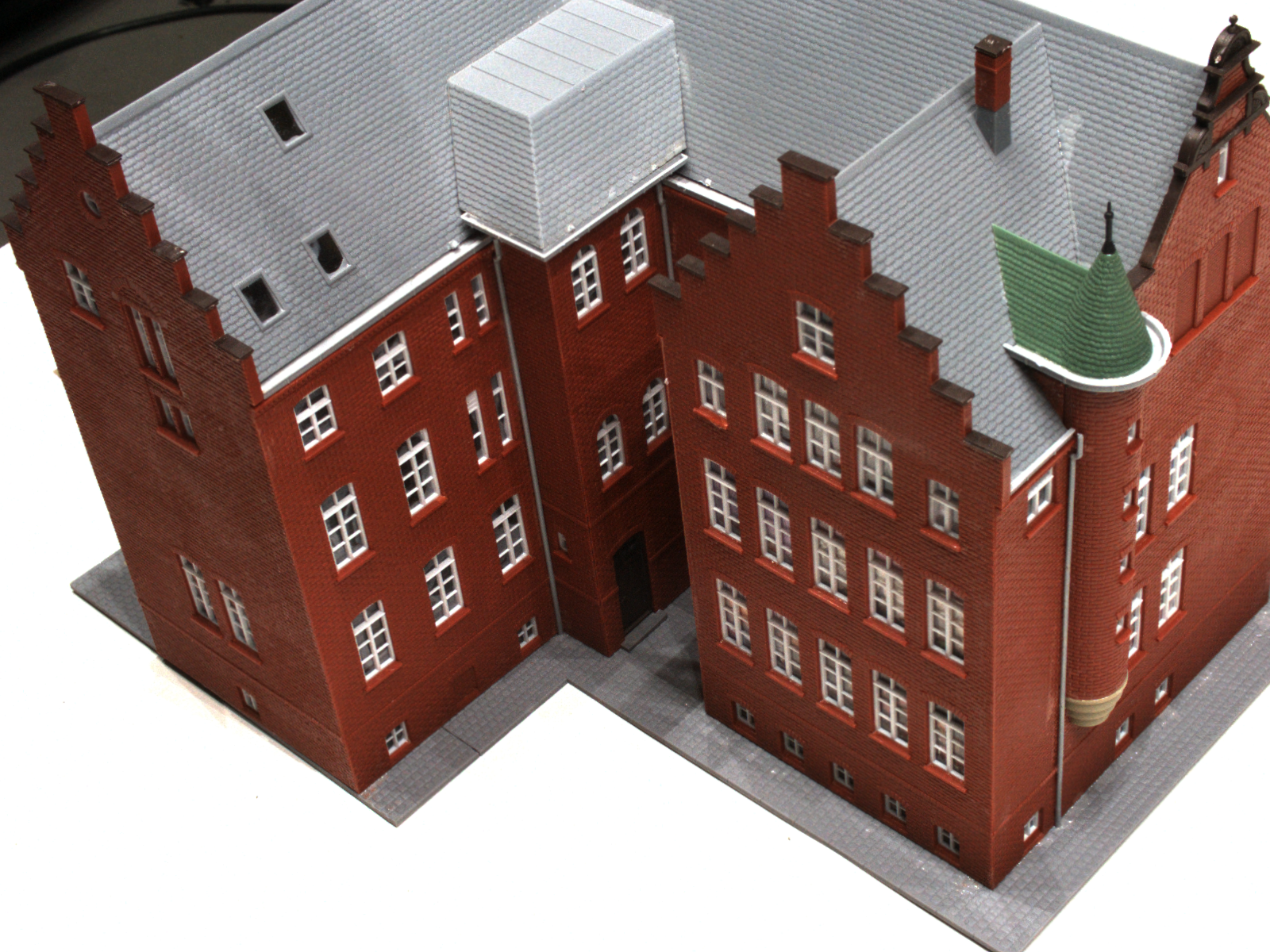} &
		\includegraphics[width=0.15\textwidth,trim=50 10 70 20,clip]{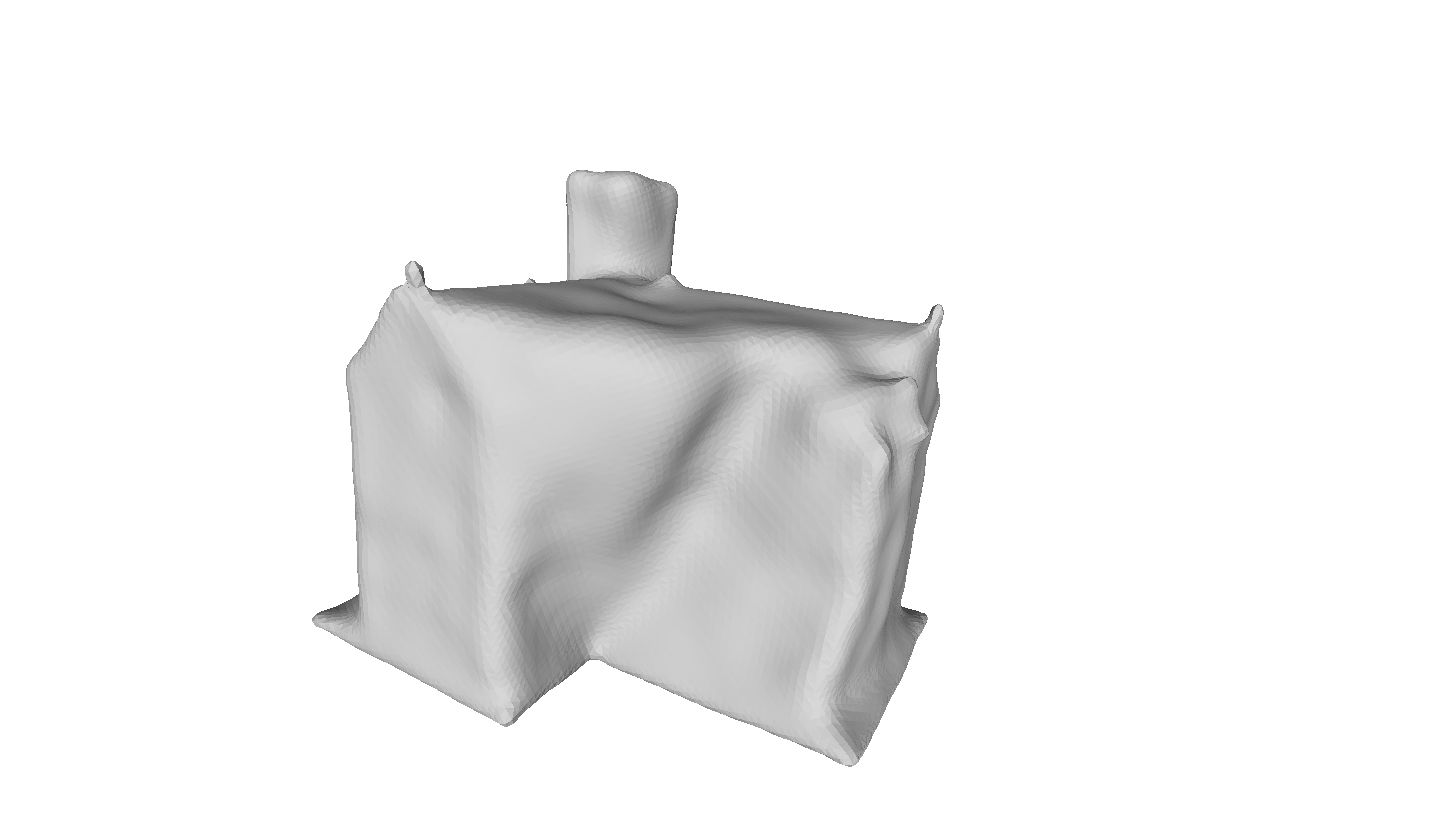} &
		\includegraphics[width=0.15\textwidth,trim=50 10 70 20,clip]{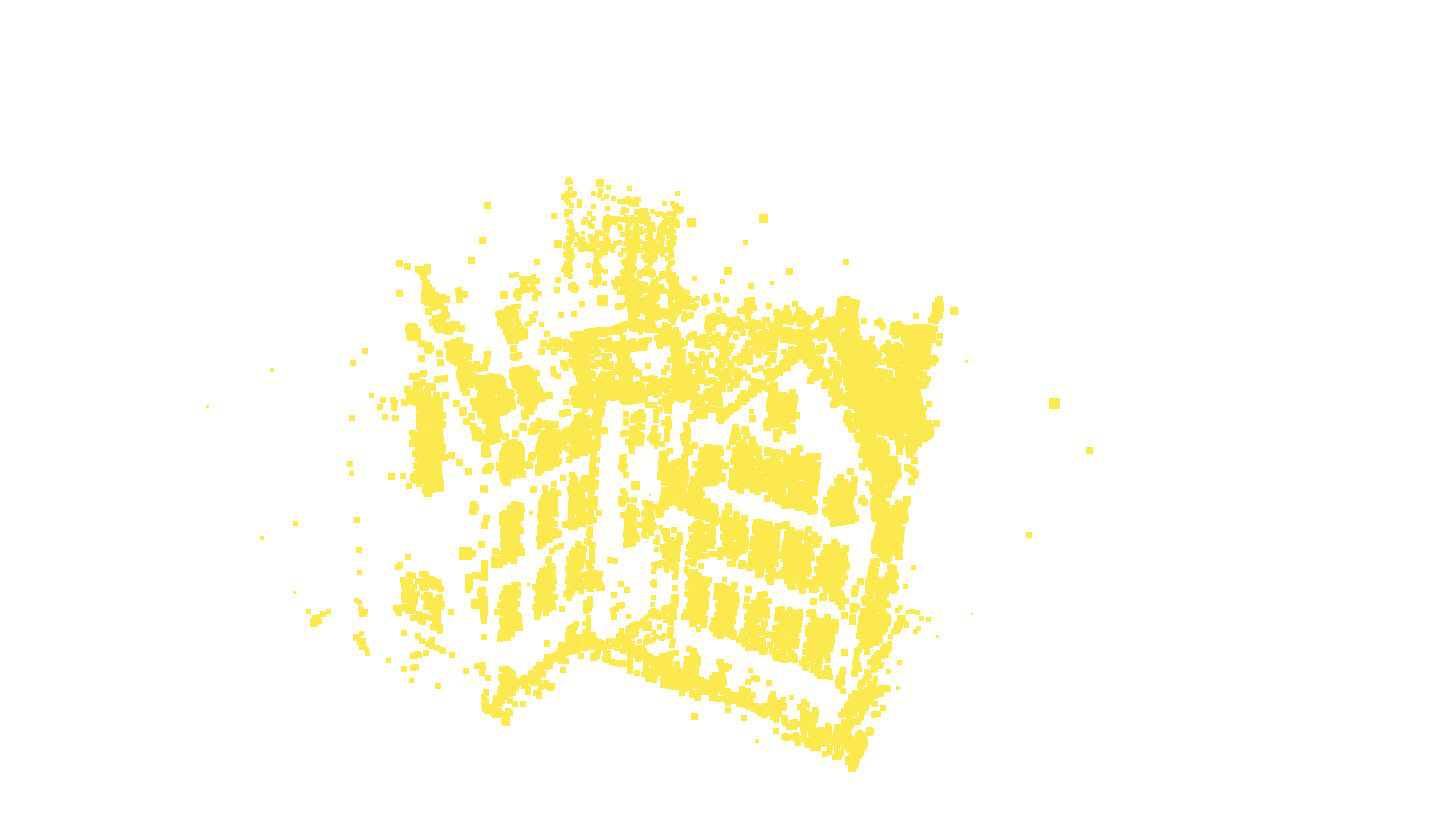} \\
		\includegraphics[width=0.15\textwidth]{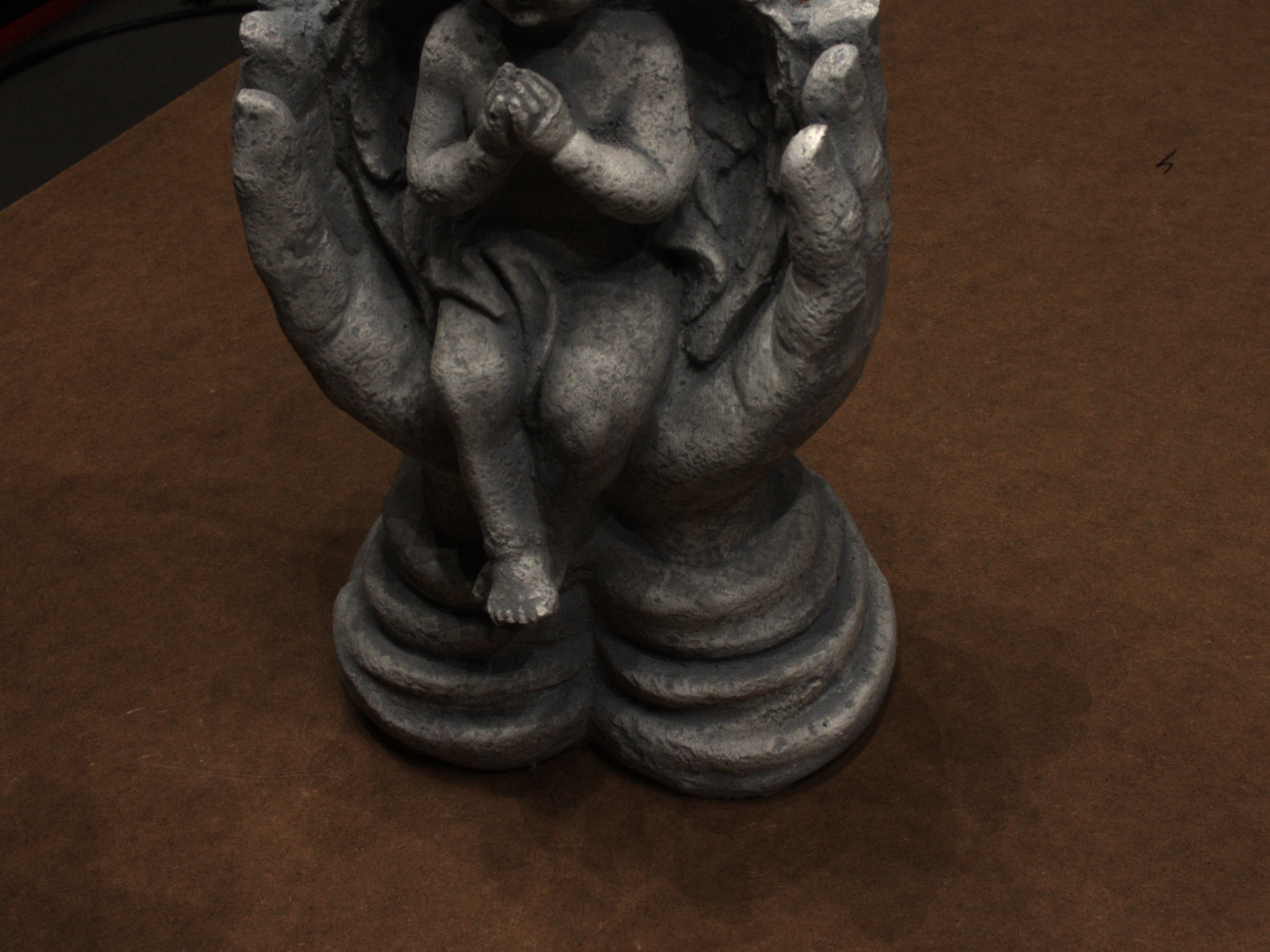} &
		\includegraphics[width=0.15\textwidth,trim=50 10 70 20,clip]{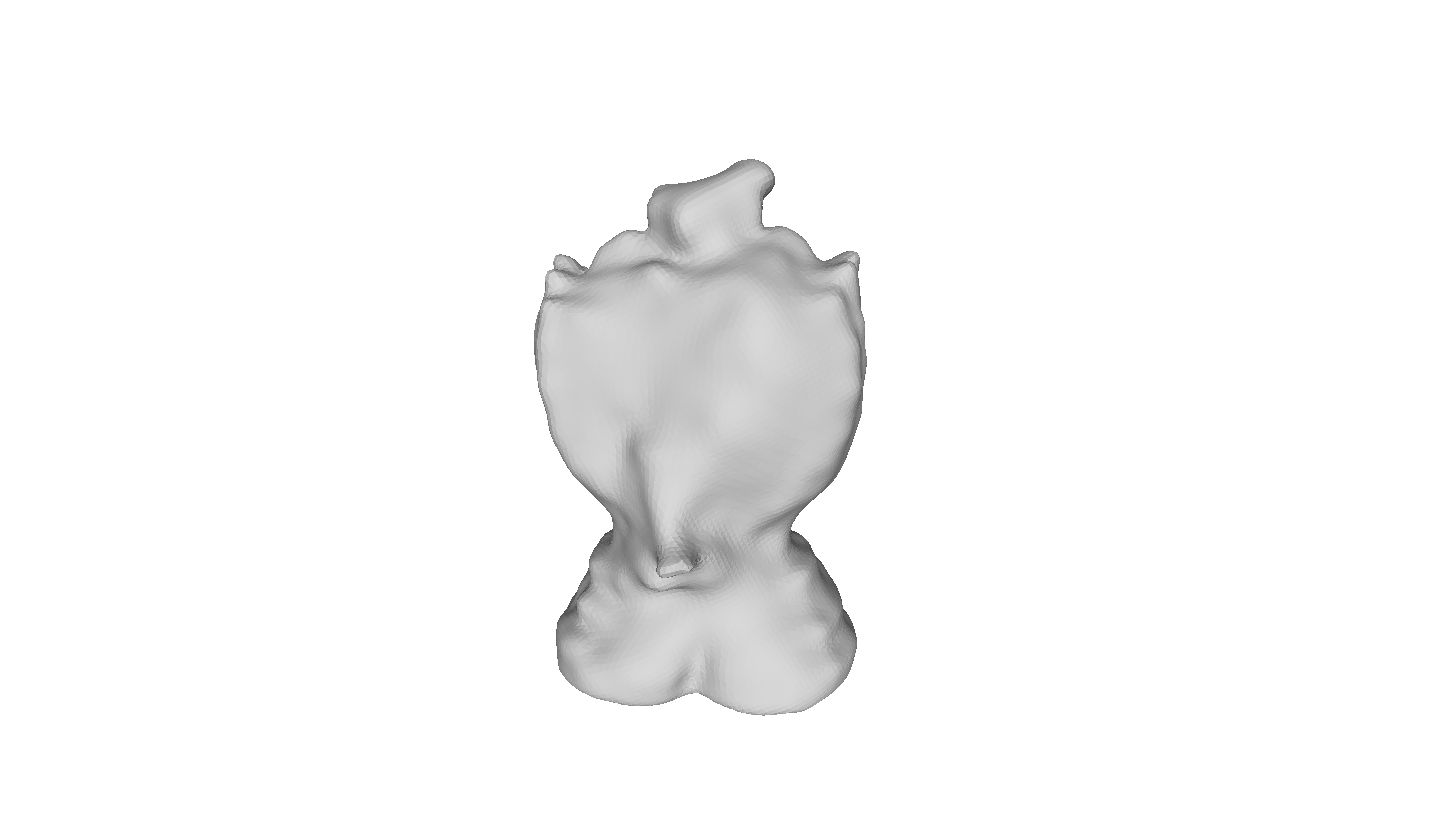} &
		\includegraphics[width=0.15\textwidth,trim=50 10 70 20,clip]{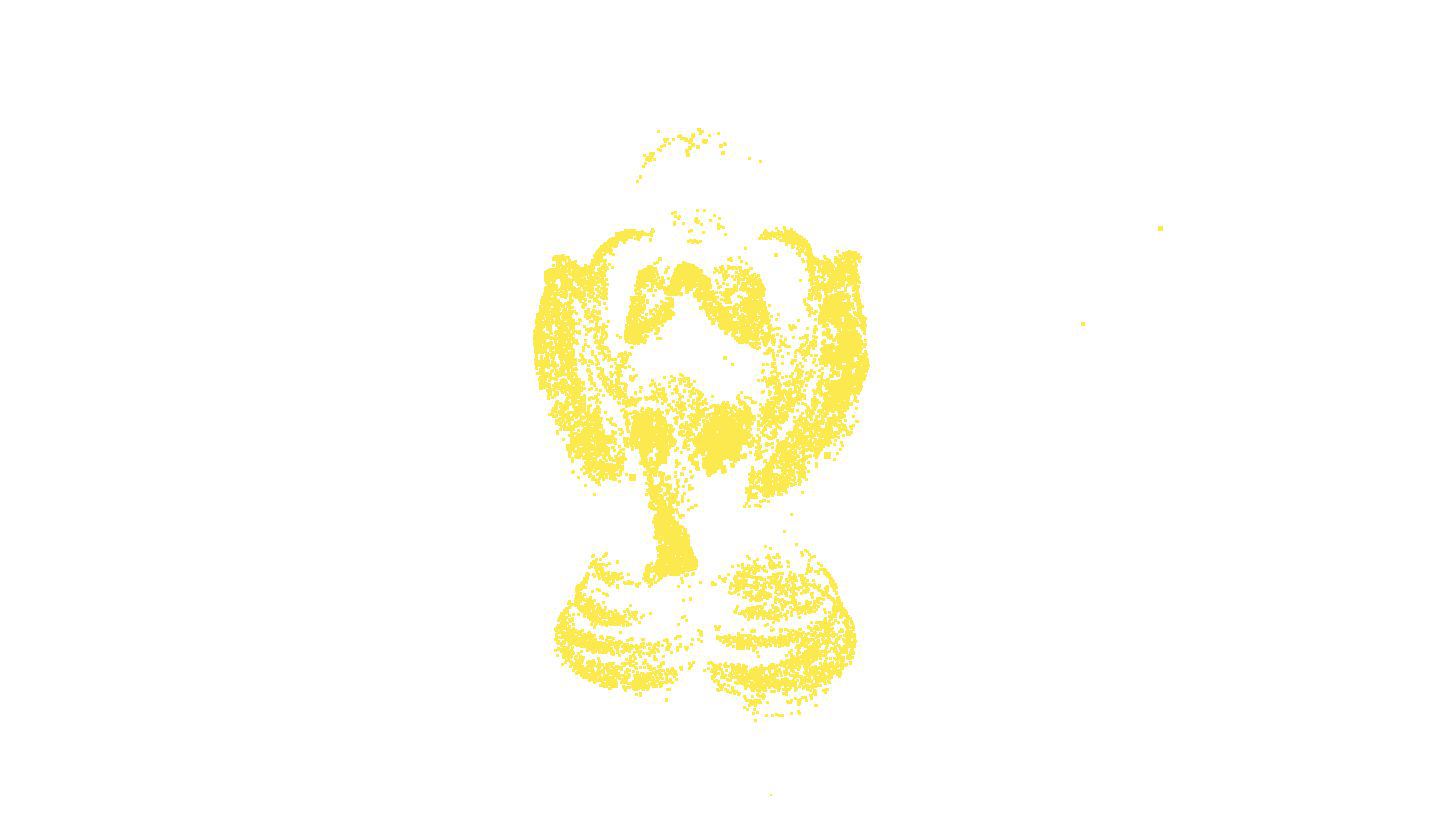} \\
		GT View & Visual Hull & Sparse Points
	\end{tabular}
	
	\caption{Examples on DTU dataset. We present images, Visual Hull results and sparse points generated by Colmap~\cite{DBLP:conf/eccv/colmap16}.}
	\label{fig:exp}
	\vspace{-0.15in}
\end{figure}

\begin{table*}
	\centering
	\caption{Quantitative results on DTU dataset. Our method achieves comparable mean Chamfer distance against NeuralWarp~\cite{DBLP:conf/cvpr/neuralwarp22}}
	\resizebox{\linewidth}{!}{%
		\begin{tabular}{l||cccc||cccc||cccc}
			\toprule
			& \multicolumn{4}{c||}{Accuracy (mm)}         & \multicolumn{4}{c||}{Completeness (mm)}     & \multicolumn{4}{c}{Chamfer (mm)}              \\ \cline{2-13}
			& IDR~\cite{DBLP:conf/nips/idr20} & NeuS~\cite{DBLP:conf/nips/neus21} & NeuralWarp~\cite{DBLP:conf/cvpr/neuralwarp22} & Ours & IDR~\cite{DBLP:conf/nips/idr20} & NeuS~\cite{DBLP:conf/nips/neus21} & NeuralWarp~\cite{DBLP:conf/cvpr/neuralwarp22} & Ours & IDR~\cite{DBLP:conf/nips/idr20} & NeuS~\cite{DBLP:conf/nips/neus21} & NeuralWarp~\cite{DBLP:conf/cvpr/neuralwarp22} & Ours  \\ \hline
			24   & 1.76 & 0.90 & \textbf{0.52} & 0.81 	& 1.50 & 0.75 & \textbf{0.46} & 0.48 	& 1.63 & 0.83 & \textbf{0.49} & 0.64 \\
			37   & 2.16 & 1.09 & \textbf{0.82} & 1.36 	& 1.55 & 0.88 & \textbf{0.61} & 0.81 	& 1.86 & 0.98 & \textbf{0.71} & 1.09 \\
			40   & 0.65 & 0.58 & \textbf{0.39} & 0.38 	& 0.61 & 0.54 & 0.37 & \textbf{0.31} 	& 0.63 & 0.56 & 0.38 & \textbf{0.35} \\
			55   & 0.57 & 0.40 & 0.40 & \textbf{0.35} 	& 0.37 & 0.34 & 0.37 & \textbf{0.27} 	& 0.47 & \textbf{0.37} & 0.38 & \textbf{0.31} \\
			63   & 1.43 & 1.62 & \textbf{1.01} & 1.42 	& 0.63 & 0.64 & 0.58 & \textbf{0.43} 	& 1.03 & 1.13 & \textbf{0.79} & 0.93 \\
			65   & 0.88 & \textbf{0.68} & 0.81 & 0.75 	& 0.69 & \textbf{0.51} & 0.81 & 0.76 	& 0.78 & \textbf{0.59} & 0.81 & 0.75 \\
			69   & 0.88 & \textbf{0.68} & 0.92 & 0.70 	& 0.66 & 0.52 & 0.72 & \textbf{0.49} 	& 0.77 & \textbf{0.60} & 0.82 & \textbf{0.60} \\
			83   & 1.10 & 1.33 & \textbf{0.85} & 1.11 	& 1.55 & 1.57 & 1.55 & \textbf{0.95} 	& 1.32 & 1.45 & 1.20 & \textbf{1.01} \\
			97   & 1.31 & 1.07 & \textbf{0.84} & 1.06 	& 0.99 & \textbf{0.84} & 1.28 & 1.44 	& 1.15 & \textbf{0.95} & 1.06 & 1.27 \\
			105  & 0.86 & 0.78 & \textbf{0.59} & 0.64 	& \textbf{0.66} & 0.78 & 0.77 & \textbf{0.66} 	& 0.76 & 0.78 & 0.68 & \textbf{0.65} \\
			106  & 0.71 & \textbf{0.53} & 0.57 & 0.64 	& 0.60 & \textbf{0.52} & 0.74 & 0.66 	& 0.66 & \textbf{0.52} & 0.66 & 0.65 \\
			110  & 1.09 & 1.71 & \textbf{0.92} & 1.06 	& 0.68 & 1.16 & 0.56 & \textbf{0.41} 	& 0.89 & 1.44 & 0.74 & \textbf{0.73} \\
			114  & 0.45 & 0.34 & 0.41 & \textbf{0.32} 	& 0.38 & 0.38 & 0.40 & \textbf{0.34} 	& 0.41 & 0.36 & 0.41 & \textbf{0.33} \\
			118  & 0.55 & \textbf{0.48} & 0.73 & 0.52 	& 0.46 & \textbf{0.43} & 0.54 & 0.44 	& 0.50 & \textbf{0.45} & 0.63 & 0.48 \\
			122  & 0.71 & 0.57 & 0.55 & \textbf{0.54} 	& 0.43 & 0.41 & 0.46 & \textbf{0.34} 	& 0.57 & 0.49 & 0.51 & \textbf{0.44} \\ \hline
			Mean & 1.01 & 0.85 & \textbf{0.69} & 0.78 	& 0.78 & 0.68 & 0.68 & \textbf{0.59} 	& 0.90 & 0.77 & \textbf{0.68} & \textbf{0.68} \\
			\bottomrule
	\end{tabular}}
	\label{tab:dtu}
\end{table*}

\begin{figure*}
	\centering
	\begin{tabular}
		{@{\hskip2pt}c@{\hskip2pt}@{\hskip2pt}c@{\hskip2pt}@{\hskip2pt}c@{\hskip2pt}@{\hskip2pt}c@{\hskip2pt}@{\hskip2pt}c@{\hskip2pt}@{\hskip2pt}c@{\hskip2pt}}
		\includegraphics[width=0.2\textwidth]{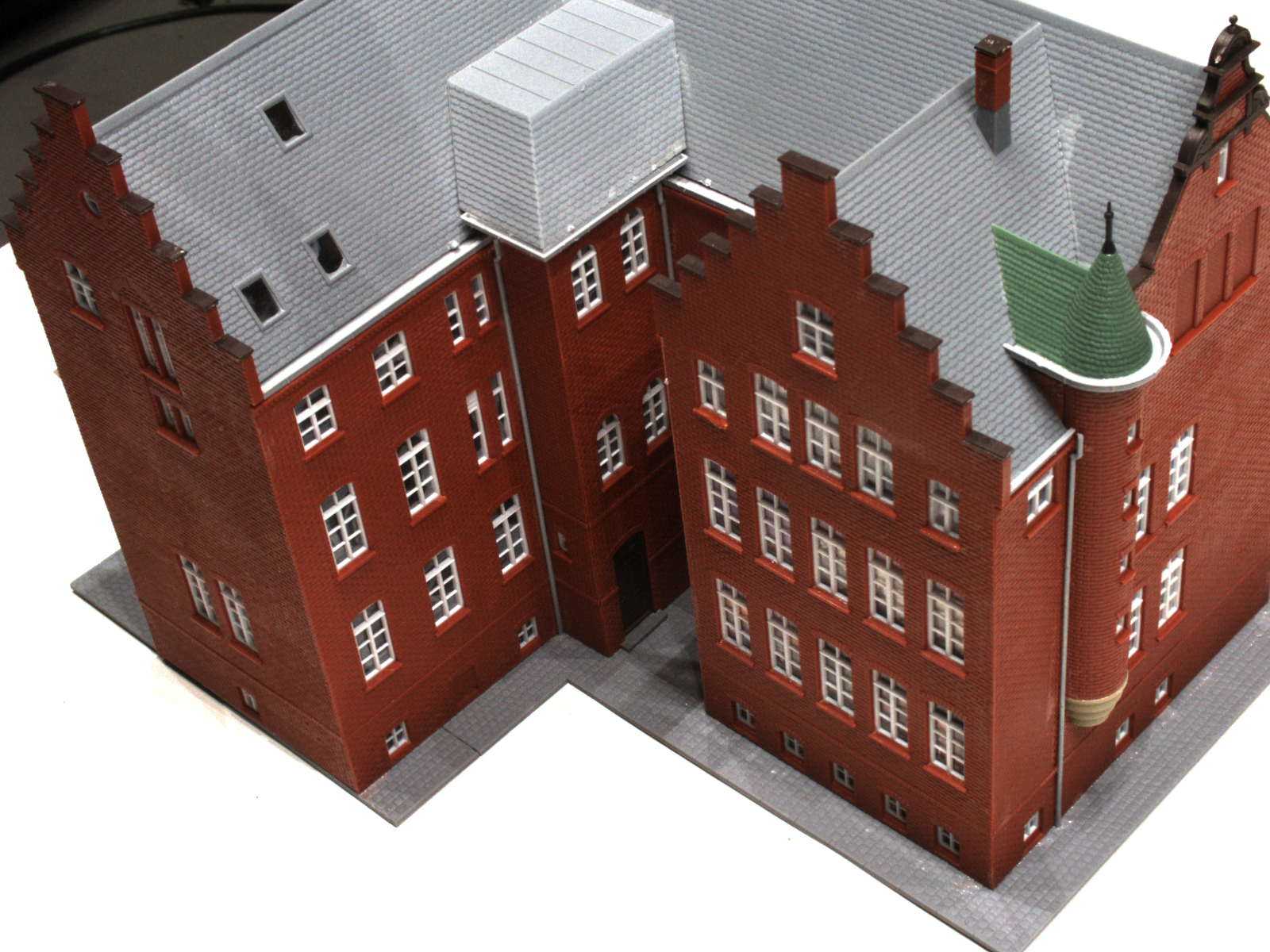} &
		\includegraphics[width=0.2\textwidth]{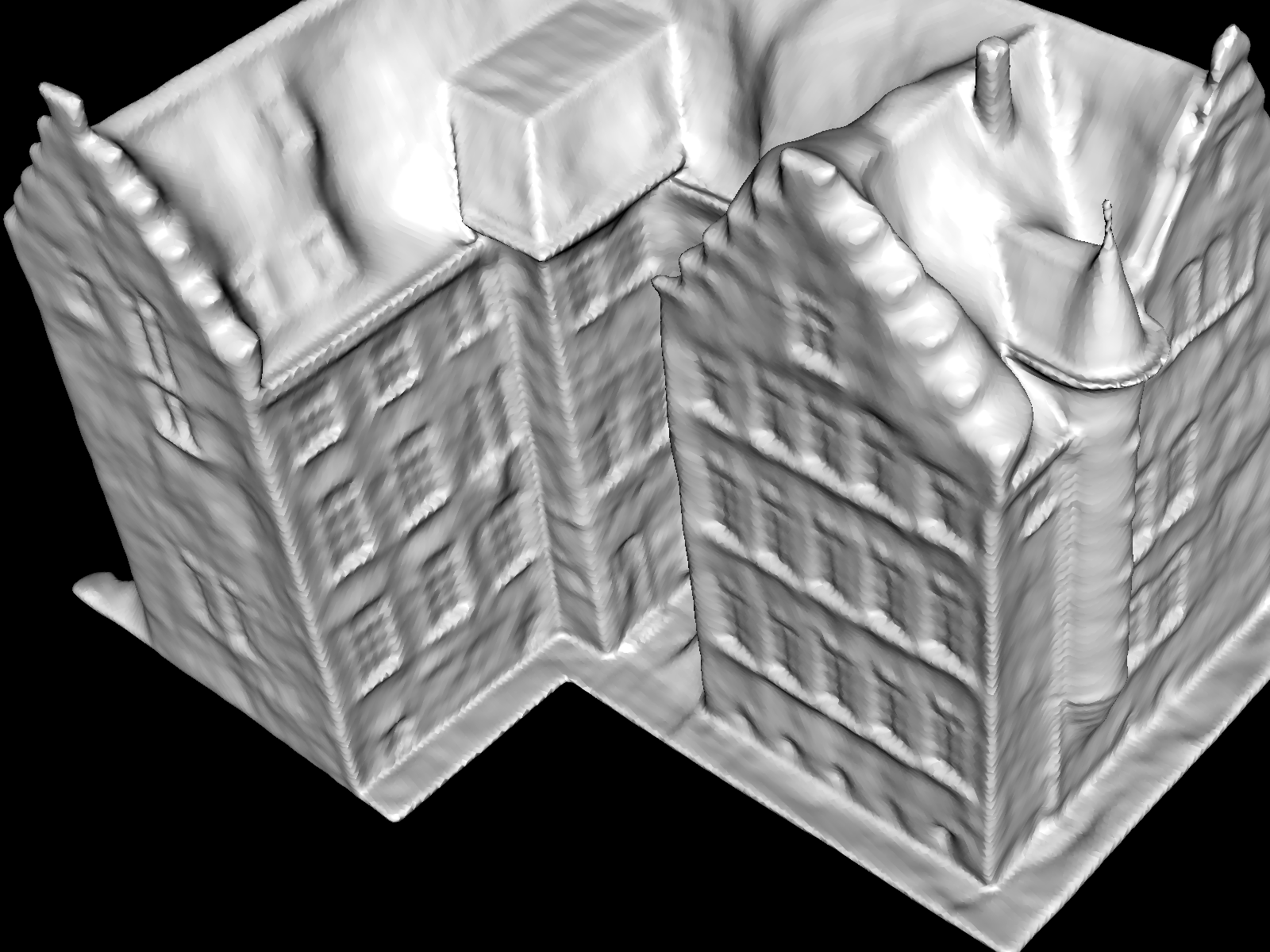} &
		\includegraphics[width=0.2\textwidth]{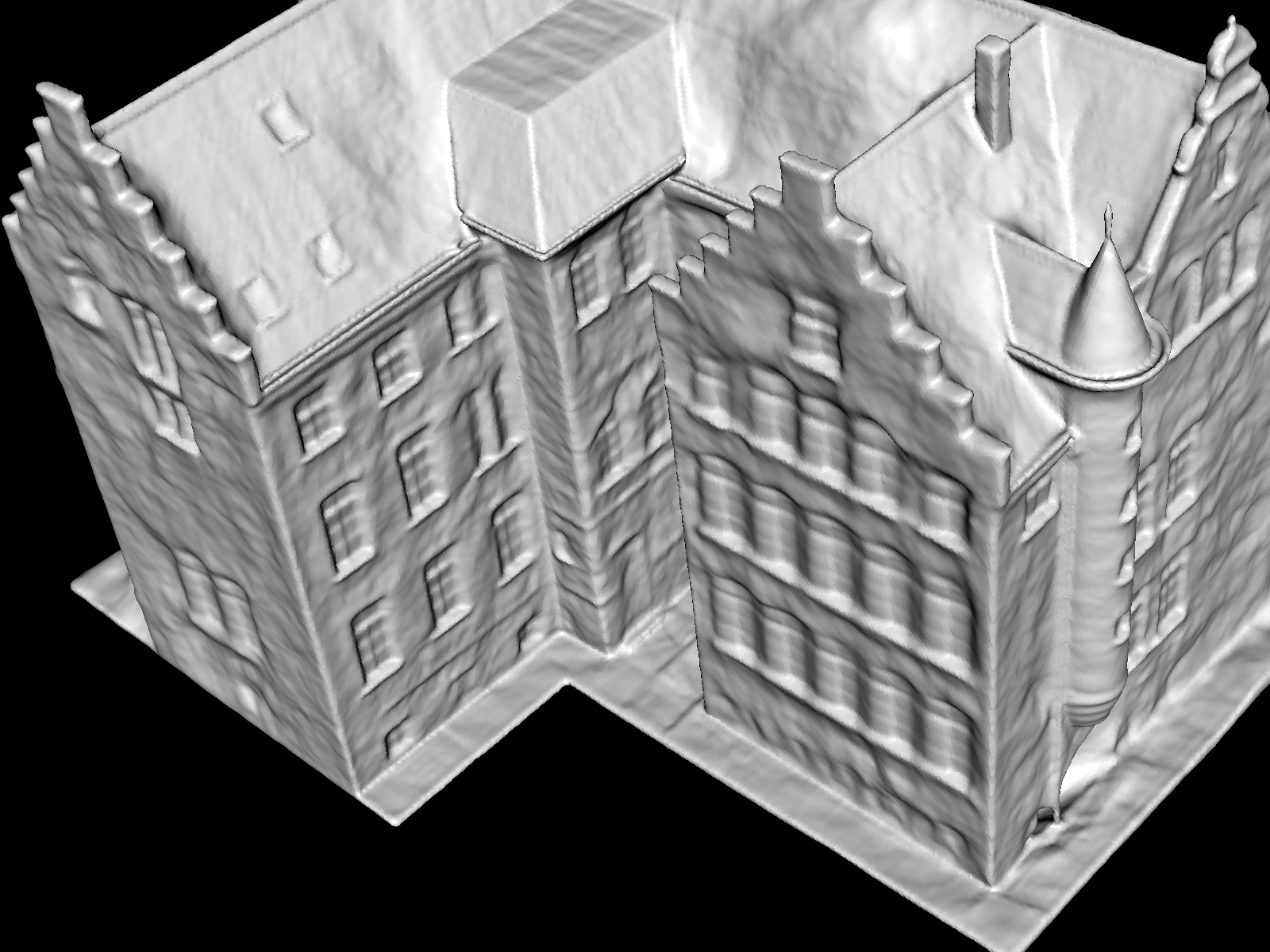} &
		\includegraphics[width=0.2\textwidth]{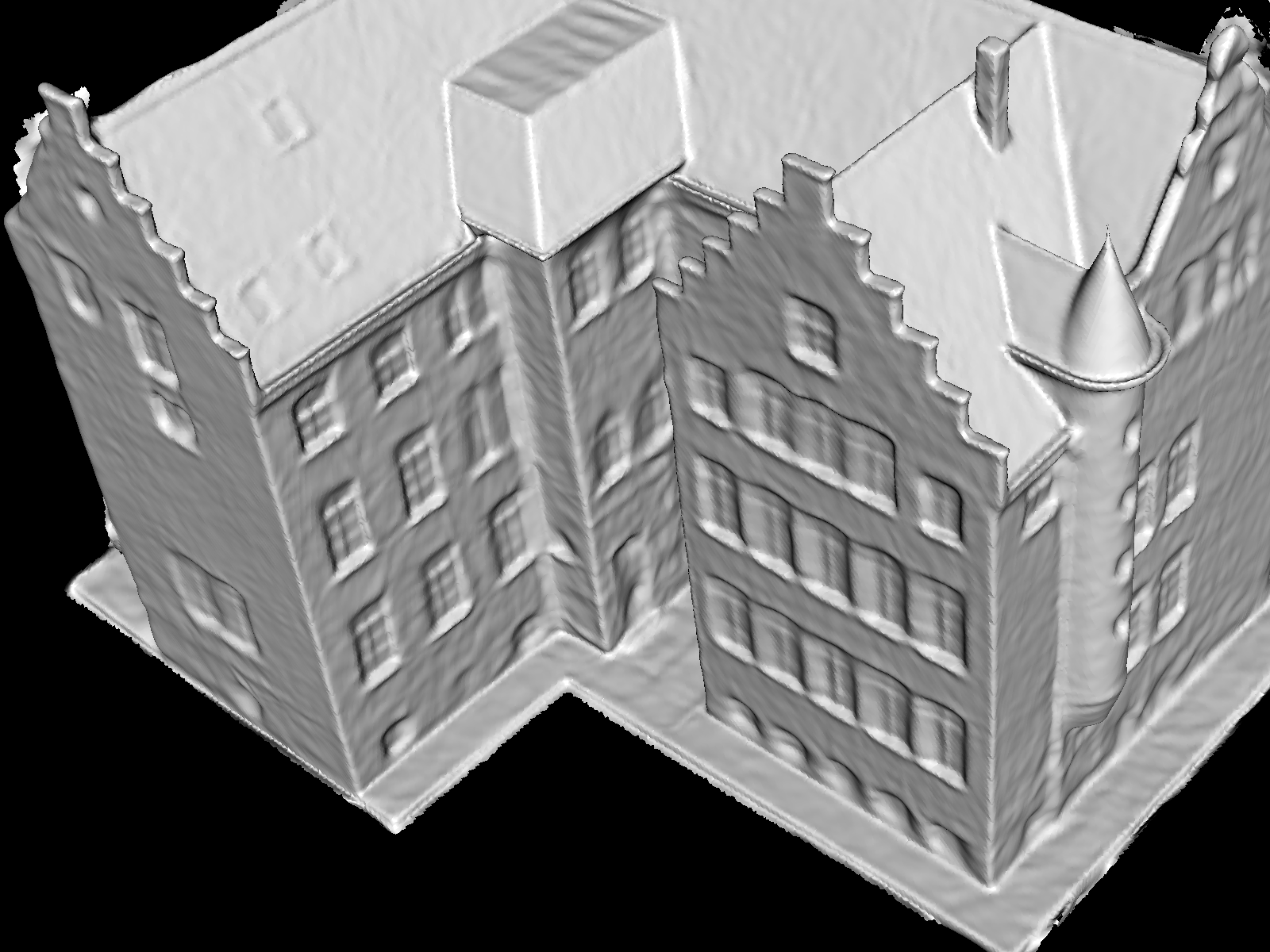} &
		\includegraphics[width=0.2\textwidth]{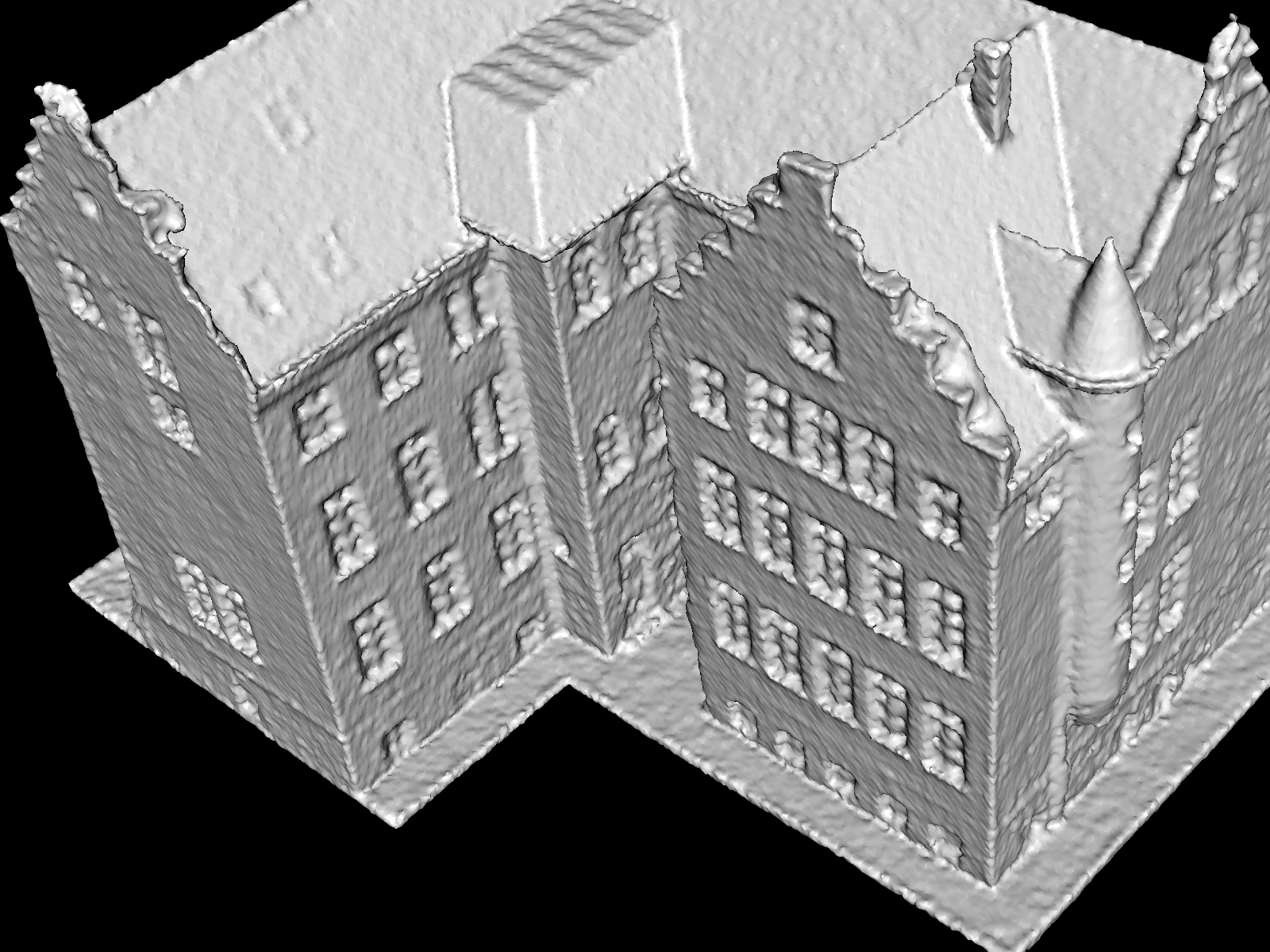} \\
		\includegraphics[width=0.2\textwidth]{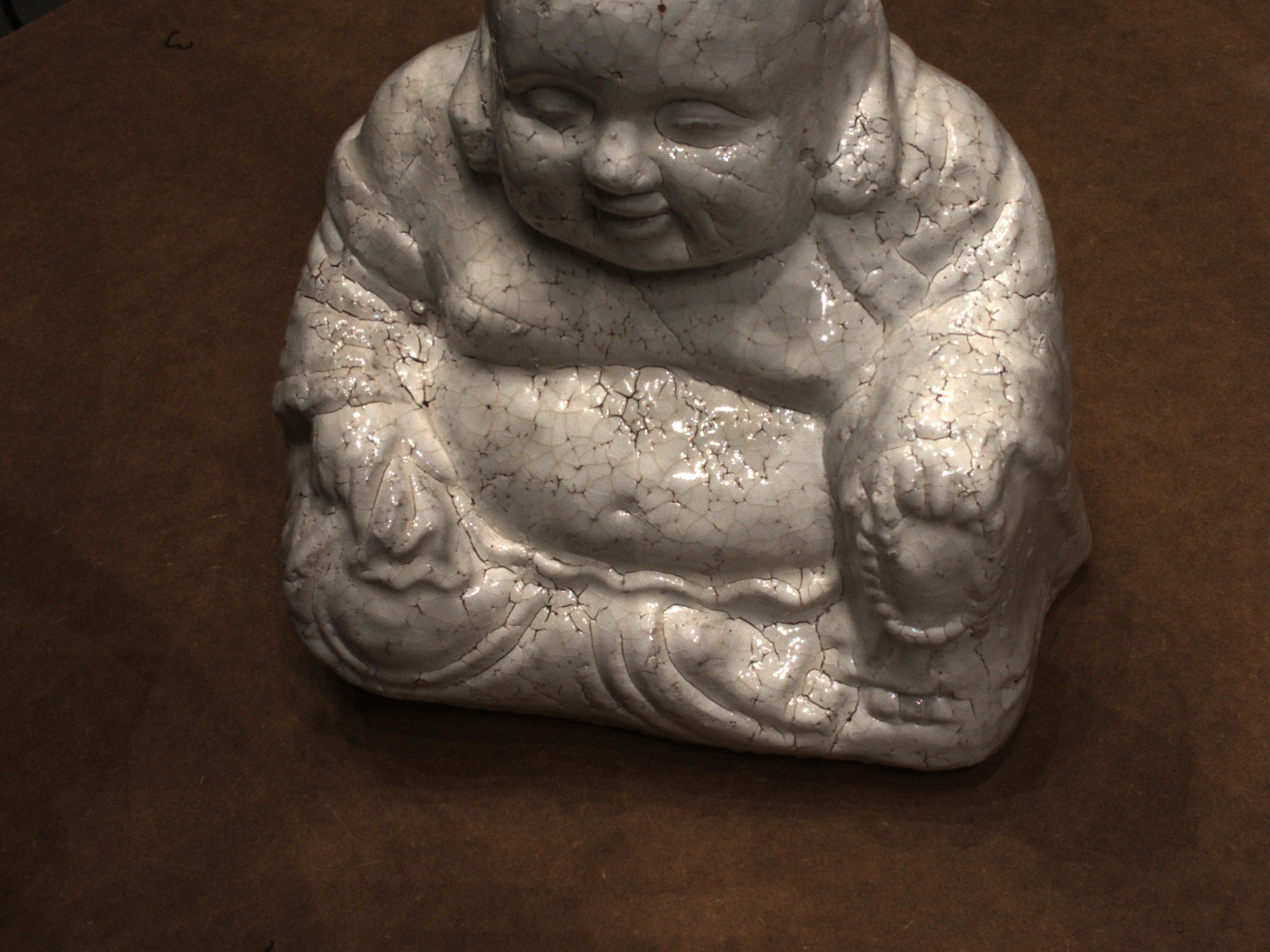} &
		\includegraphics[width=0.2\textwidth]{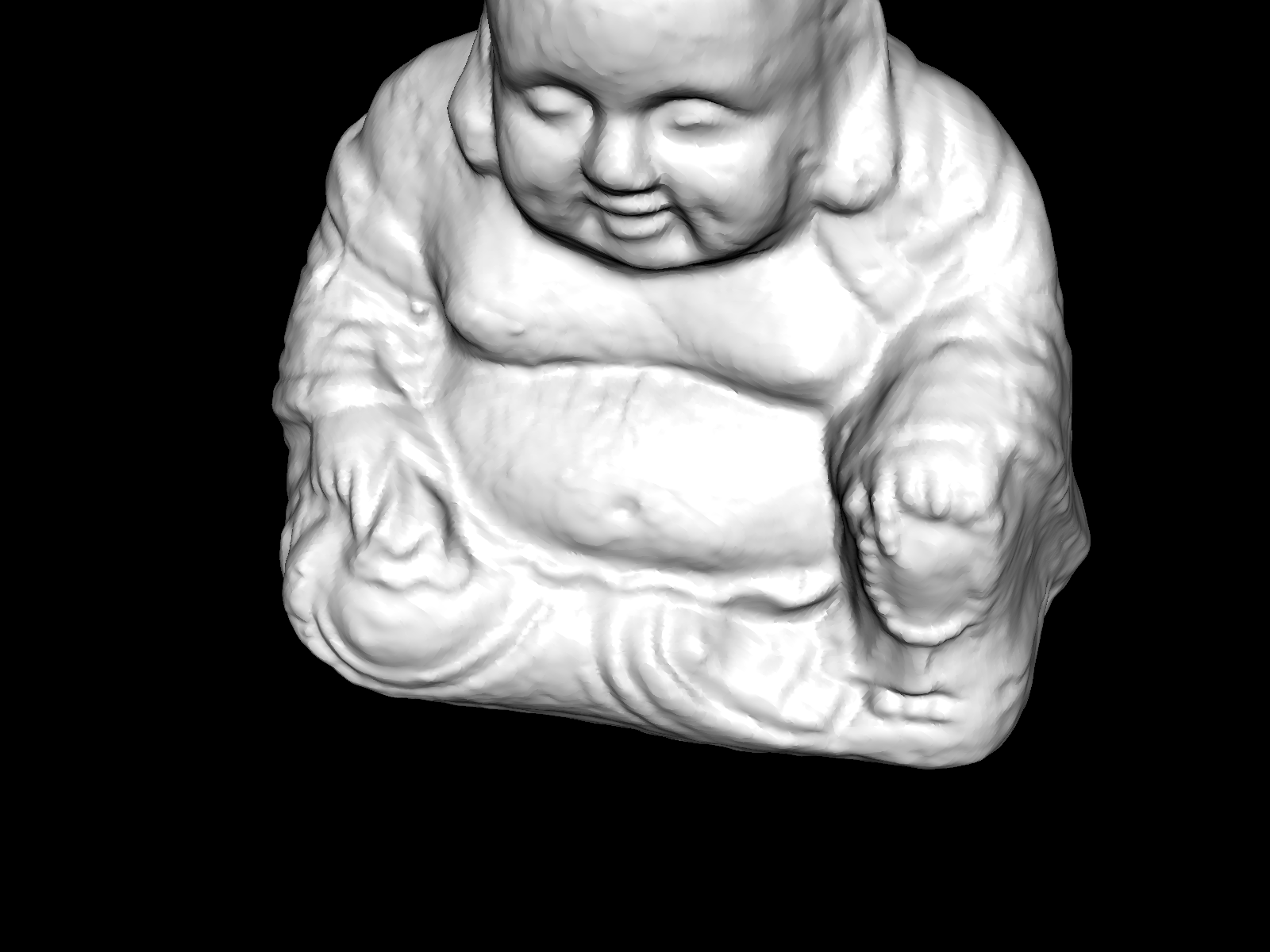} &
		\includegraphics[width=0.2\textwidth]{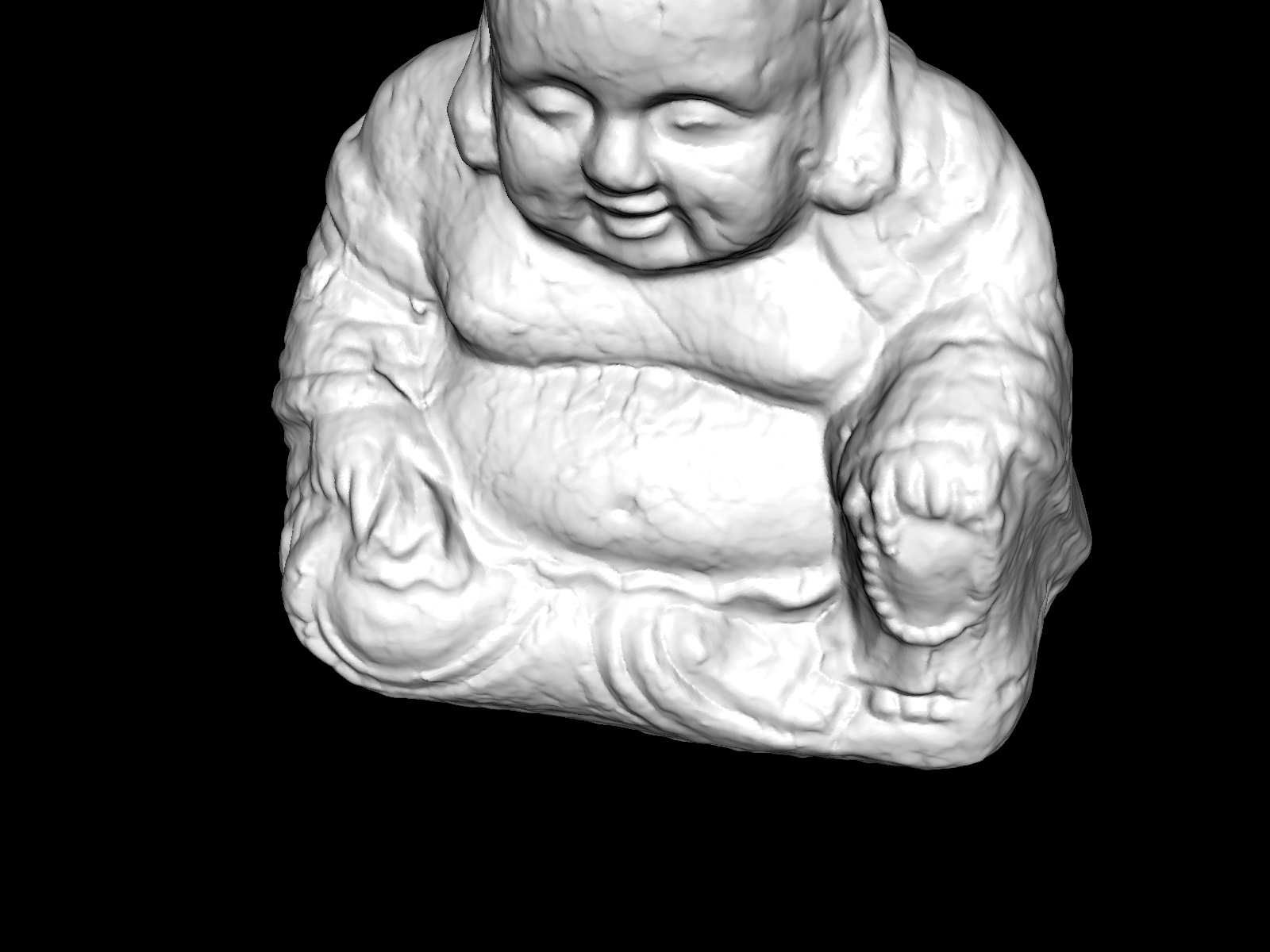} &
$  $		\includegraphics[width=0.2\textwidth]{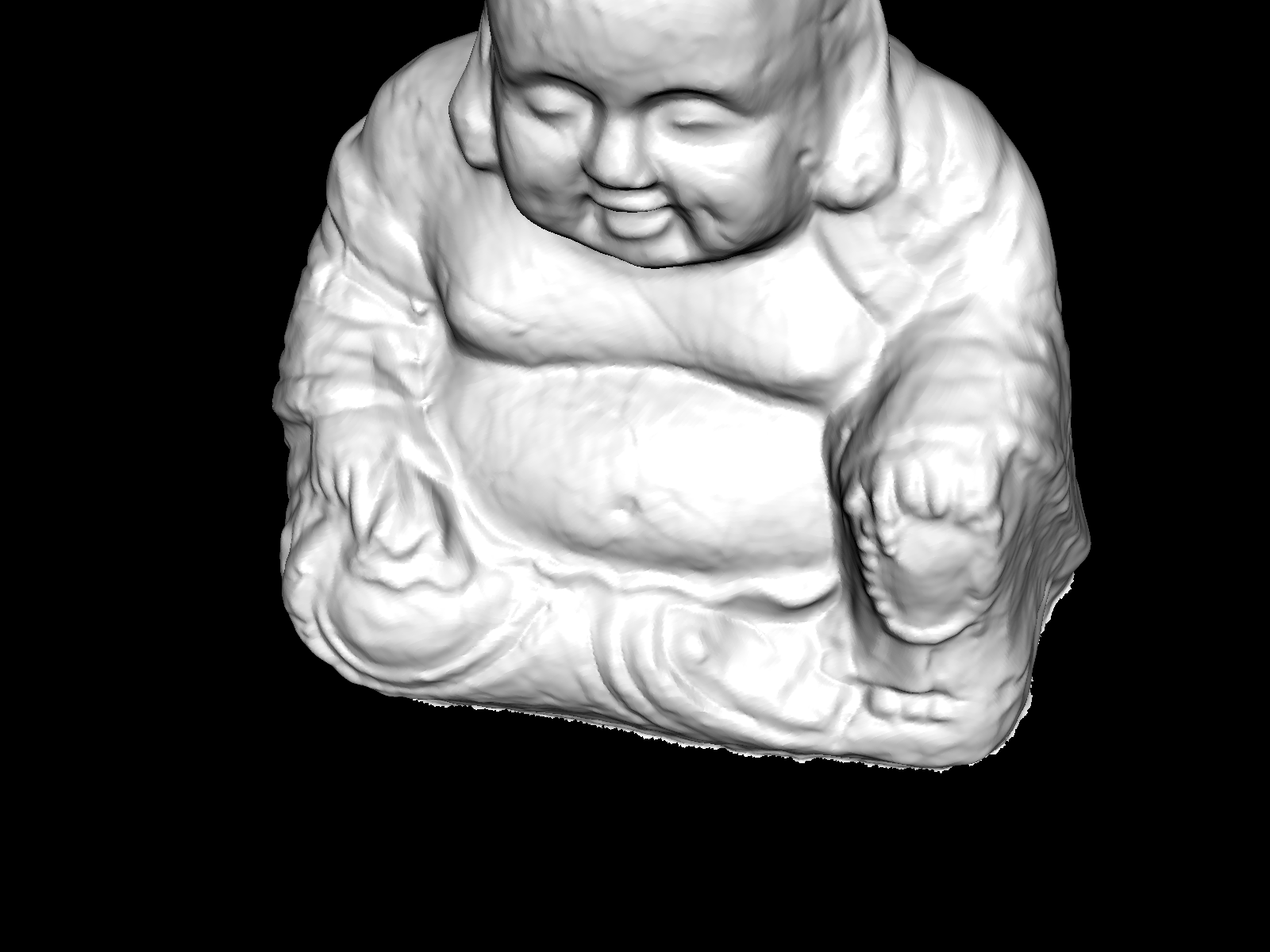} &
		\includegraphics[width=0.2\textwidth]{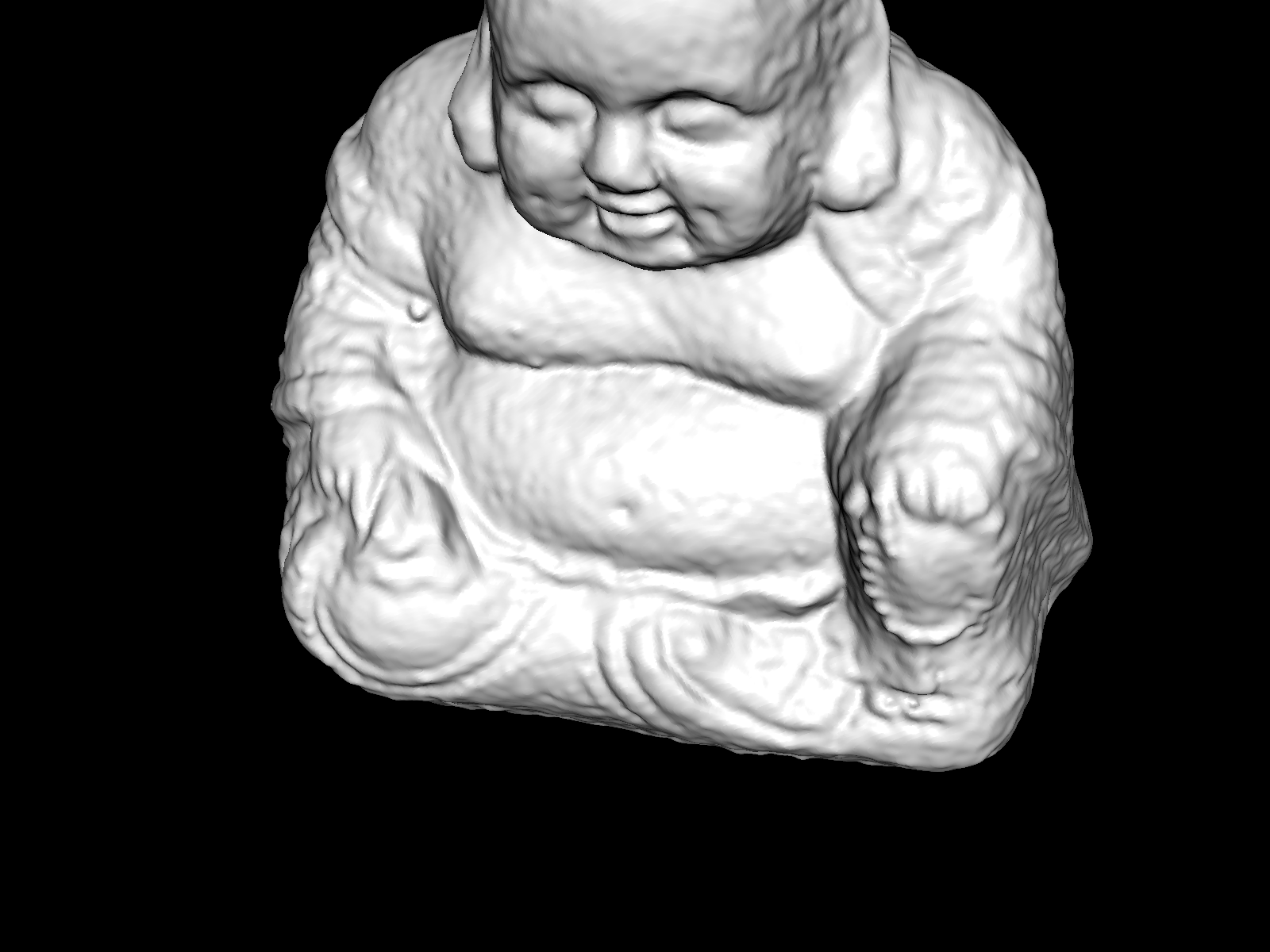} \\
		GT View & IDR~\cite{DBLP:conf/nips/idr20} & NeuS~\cite{DBLP:conf/nips/neus21} & NueralWarp~\cite{DBLP:conf/cvpr/neuralwarp22} & Ours
	\end{tabular}
	
	\caption{\textbf{Qualitative Comparison on DTU dataset~\cite{DBLP:conf/cvpr/DTU14}.} We compare our method with IDR~\cite{DBLP:conf/nips/idr20}, NeuS~\cite{DBLP:conf/nips/neus21}, and NeuralWarp~\cite{DBLP:conf/cvpr/neuralwarp22}. }
	\label{fig:dtu}
	\vspace{-0.15in}
\end{figure*}

\section{Results on DTU Dataset}

In this section, we give the reconstruction results of our proposed patch warping loss on general objects. We have conducted experiments on DTU dataset~\cite{DBLP:conf/cvpr/DTU14},  compared the reconstruction results against IDR~\cite{DBLP:conf/nips/idr20}, NeuS~\cite{DBLP:conf/nips/neus21}, and NeuralWarp~\cite{DBLP:conf/cvpr/neuralwarp22}. We employ colmap~\cite{DBLP:conf/eccv/colmap16} to obtain the sparse points as additional supervision. Fig.~\ref{fig:exp} shows the results of Visual Hull and sparse points generated by Colmap~\cite{DBLP:conf/eccv/colmap16}. The results of Visual Hull is not correct due to  the views of DTU dataset are only from one side. The sparse points generated by Colmap is too sparse to evaluate. For dense reconstruction results of Colmap on DTU dataset, you can refer to IDR~\cite{DBLP:conf/nips/idr20} for more information. As the objects in DTU dataset have specular reflections, we do not perform shape from shading refinement. Quantitative results are shown in Table~\ref{tab:dtu}. It can be seen that our proposed patch warping loss achieves the best Completeness distance and comparable Chamfer distance against NeuralWarp. Qualitative results are presented in Fig.~\ref{fig:dtu}. As mentioned in our paper, it takes 15 minutes for our proposed patch warping to reconstruct an object of DTU dataset while other methods requires several hours to obtain the results.

\end{document}